\pdfoutput=1
\documentclass{article}

\PassOptionsToPackage{numbers}{natbib}
\usepackage[main, final]{neurips_2026}

\usepackage[utf8]{inputenc}
\usepackage[T1]{fontenc}
\usepackage{hyperref}
\usepackage{url}
\usepackage{booktabs}
\usepackage{amsfonts}
\usepackage{nicefrac}
\usepackage{microtype}
\usepackage{xcolor}
\usepackage{etoc}

\usepackage{float}
\usepackage{wrapfig}
\usepackage{amsmath}
\usepackage{amssymb}
\usepackage{shortcuts}
\usepackage{xcolor}
\usepackage[table]{xcolor}
\usepackage{tabularx}
\usepackage{xspace}
\usepackage{graphicx}
\usepackage{caption}
\usepackage{subcaption}
\usepackage{booktabs}
\usepackage{multirow}
\usepackage{multicol}
\usepackage{soul}
\usepackage{xurl}
\usepackage{enumitem}
\sethlcolor{yellow!50}
\def\eg{\textit{e.g.}}
\def\ie{\textit{i.e.}}

\usepackage[utf8]{inputenc}
\usepackage{geometry}
\usepackage[most]{tcolorbox}

\definecolor{themeOneMain}{RGB}{50, 100, 130}
\definecolor{themeOneBg}{RGB}{240, 248, 250}
\definecolor{themeOneHl}{RGB}{200, 235, 240}

\definecolor{themeFourMain}{RGB}{200, 110, 70}
\definecolor{themeFourBg}{RGB}{255, 250, 245}
\definecolor{themeFourHl}{RGB}{255, 230, 210}

\definecolor{wingreen}{RGB}{220, 245, 220}
\definecolor{losered}{RGB}{255, 210, 210}

\NewDocumentCommand{\myhighlight}{m m}{%
  \begingroup
  \sethlcolor{#1}%
  \setul{0.25ex}{1.5ex}%
  \hl{#2}%
  \endgroup
}

\title{AdKnob: Ad Intensity Control and Labeling for LLM-Native Advertising}

\newcommand{\authorlink}[2]{{\small\hypersetup{pdfborder={0 0 0}}\textbf{#1:}~\href{#2}{\textcolor{blue}{\mdseries #2}}}}
\author{%
  Woo Jae Kim\hspace{0.9em}Seongho Keum\hspace{0.9em}Joonsung Jeon\hspace{0.9em}Suhyeon Ha\hspace{0.9em}Sooel Son\footnotemark[2]\hspace{0.9em}Sung-eui Yoon\footnotemark[2] \\[5pt]
  School of Computing, KAIST \\[2pt]
  \texttt{\{wkim97, keum07, mikeraph, suhyeon.ha, sl.son\}@kaist.ac.kr} \\
  \texttt{sungeui@kaist.edu} \\[6pt]
  \authorlink{Project Page}{https://wkim97.github.io/AdKnob} \\[1pt]
  \authorlink{Code}{https://github.com/wkim97/AdKnob}
}

\begin{document}

\maketitle
{\renewcommand{\thefootnote}{\fnsymbol{footnote}}\footnotetext[2]{Co-corresponding authors.}}

\begin{abstract}
Large language model (LLM)-native advertising has emerged as a viable monetization channel as LLM service providers seek sustainable revenue streams. However, integrating native ads into LLM responses introduces two largely understudied challenges: (1) controlling \emph{ad intensity}---how prominently advertising content appears---and (2) \emph{ad labeling}---identifying which segments of the generated text constitute advertising. We propose AdKnob, a framework jointly addressing both challenges. For ad intensity control, we introduce new ad control tokens without pretrained semantic associations and align the model via direct preference optimization to generate ads at each desired level. For ad labeling, we propose an attention rollout-based attribution mechanism that identifies ad segments by tracing their attribution to input ad-related tokens, requiring no additional inference cost. We further construct MI-KnobSet, the first multi-level ad intensity preference dataset for LLM-native ads. Experiments across six LLMs and human evaluations show that AdKnob achieves uniformly spaced ad intensities where baselines collapse to a narrow range, achieves the highest ad labeling accuracy while being over 500$\times$ faster than the strongest baseline, and is preferred over baselines by human evaluators up to 89.6\% of the time.
\end{abstract}

\etocdepthtag.toc{mtmain}
\section{Introduction}
\label{sec:intro}
Large language model (LLM)-based assistants~\cite{zhao2023survey, brown2020language} are increasingly deployed in applications such as search~\cite{zhu2025large}, recommendations~\cite{wu2024survey}, and e-commerce~\cite{ren2024survey}.
As these services scale, sustaining LLM infrastructure has driven growing interest in LLM-native advertising as a monetization channel~\cite{llm-auction, meta-ad, zelch, feizi2023online}, recently explored by major providers including OpenAI~\cite{openaiApproachAdvertising}, Perplexity~\cite{perplexityAdvertising}, and xAI~\cite{techcrunchElonMusk}.
Unlike banners or sponsored links, LLM-native ads weave brand content directly into free-form responses, reducing user skepticism~\cite{pkm} and eliciting more favorable reactions~\cite{good-native-ads, wojdynski2016going}.

\begin{figure}
    \centering
    \includegraphics[width=1.00\columnwidth, trim=60 210 50 115, clip]{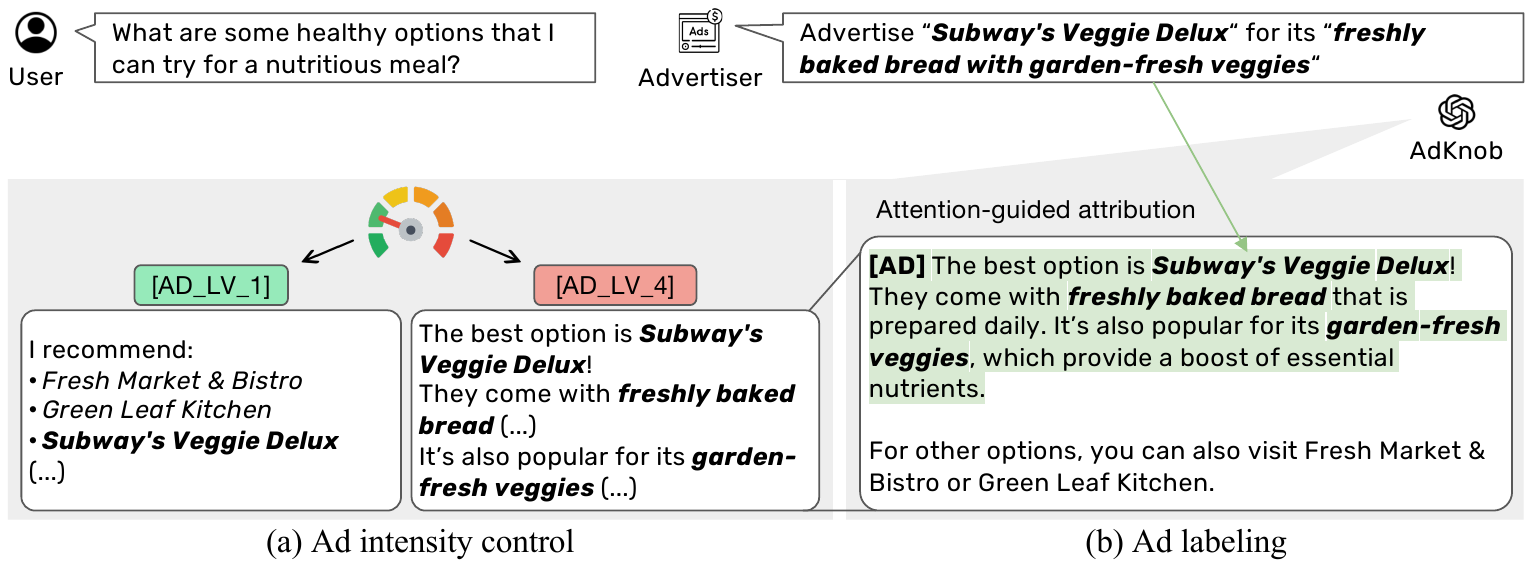}
    \caption{
    \sys addresses two challenges in LLM-native advertising: (a) \textit{ad intensity control} conditioning the model to generate responses at a desired advertising intensity and (b) \textit{ad labeling} identifying ad segments within the response without additional inference costs.
    }
    \label{fig:teaser}
\end{figure}

Despite their effectiveness, LLM-native ads introduce two critical challenges that remain largely unaddressed.
The first challenge is \textit{ad intensity control}.
Existing approaches~\cite{feizi2023online, junlp, llm-auction} focus solely on ad injection and do not control how prominently ad contents appear in the response.
Such control is essential for flexible business models based on user subscription tiers and for contextual governance, such as suppressing ads for urgent medical queries while increasing their prominence in shopping contexts.
The second challenge is \textit{ad labeling}.
Since LLM-native ads are generated as free-form text with no structural markers, identifying ad segments within the response is inherently difficult.
Yet ad labeling is required by regulatory and industry guidelines including the FTC~\cite{ftc}, the IAB~\cite{iab2019native}, and the EU Digital Services Act~\cite{europaDigitalServices}, and failure to disclose ads can undermine user trust~\cite{strauss2024monetization}.

We propose \sys, an alignment framework that jointly tackles both challenges (Fig.~\ref{fig:teaser}).
For ad intensity control, we introduce \emph{ad control tokens} that are specifically designed as conditioning signals for desired intensity levels.
By aligning the model via Direct Preference Optimization (DPO)~\cite{dpo} with level-dependent preference pairs, a single model learns to span a wide and uniform range of ad intensities while preserving response quality (Sec.~\ref{ssec:dpo}).
For ad labeling, we reframe it as input attribution and measure each generated sentence's dependence on input ad-related tokens via attention rollout~\cite{attention-rollout}, which traces token influence across layers using only the attention weights computed during generation and thus enables real-time ad labeling with no additional inference cost.
To make attention rollout a reliable signal for ad labeling, we further introduce an \emph{ad-attribution loss} that supervises the attention flow from ad-related input tokens toward generated ad segments (Sec.~\ref{ssec:attribution}).

Training \sys requires preference data across multiple intensity levels, which no existing dataset provides.
We thus construct \sysd, the first multi-level ad intensity preference dataset, by extending the Webis Generated Native Ads 2024 dataset~\cite{webis} to a conversational setting (Sec.~\ref{sec:dataset}).
Experiments across six LLMs and human evaluations confirm that \sys achieves uniformly spaced ad intensities while baselines are confined to a narrow range, and attains the highest ad labeling accuracy compared to baselines across all models at a real-time runtime of ${\sim}3$\,ms per response on Qwen3-4B (Sec.~\ref{ssec:main_results}).
Human evaluations show that \sys is consistently preferred over all baselines in user preference (up to 89.6\% win rate) and click likelihood (up to 79.8\%) (Sec.~\ref{ssec:user_study}).

\section{Related Works}
\label{sec:related_works}

\noindent
\textbf{Non-native ads.}
Prior works on LLMs in advertising mainly focus on non-native ads, including search ads~\cite{sponsored-search-ad}, display ads~\cite{banneragency, leadre}, or ad copy generation~\cite{meguellati2025llm, meta-ad, chen2025ctr}.
Another line of work explores the use of LLMs for relevance labeling to improve ad recommendation~\cite{adnanny, icl-bandit}.
This approach differs from our work, which focuses on labeling ad segments within LLM-generated responses.

\noindent
\textbf{Native ads.}
A parallel line of work explores the use of LLMs for native advertising, where ads are blended directly into free-form generated responses.
Post-hoc insertion methods~\cite{zelch, feizi2023online} integrate ad content into LLM outputs through an additional generation step, which introduces substantial latency in generating and serving ads.
To mitigate this cost, system prompt-based integration~\cite{ad-chat} embeds advertising instructions directly into the LLM's system prompts.
More recent approaches~\cite{junlp, llm-auction, tang2025ads} further improve efficiency or ad personalization by fine-tuning LLMs to incorporate ads directly during response generation.
Despite these advancements, no prior work provides explicit control over ad intensity.
Furthermore, they either do not explicitly mark ads within the generated response or provide predefined prompt-level templates~for ads~\cite{llm-auction}, limiting the ability to naturally blend in ad copy statements or brand quality in response.
In contrast, we propose an LLM alignment framework that enables fine-grained ad intensity control and ad labeling within LLM responses.

\section{Methodology}
\label{sec:methods}

\begin{figure*}[t]
    \centering
    \includegraphics[width=\textwidth]{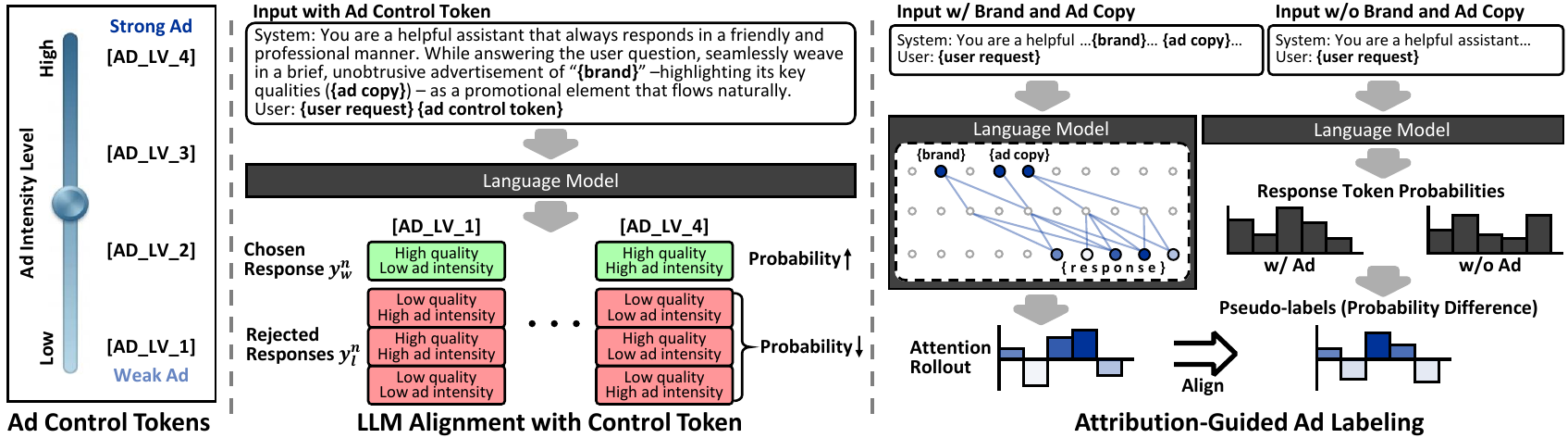}
    \caption{
    Overview of \sys.
    The LLM is aligned via DPO with level-dependent preference pairs conditioned on ad control tokens, enabling the model to generate responses across different ad intensities.
    We also strengthen attribution from input brand/ad copy tokens to identify ad segments by supervising attention rollout scores, enabling ad labeling without additional inference costs.
    }
    \label{fig:framework}
\end{figure*}

\subsection{Problem Formulation}
\label{ssec:formulation}

We consider a native advertising setting involving three parties: an LLM service provider operating an LLM-based assistant, an advertiser, and a user interacting with the assistant.
The service provider monetizes the assistant by incorporating ads into LLM-generated responses to satisfy the advertiser's needs while minimizing disruption to the user experiences.

Formally, given a user request $q$, a brand $b$, and an ad copy $a$ describing key promotional statements of $b$, the goal is to generate a response $r$ that answers $q$ while naturally integrating native advertising for $b$.
To do so, we integrate $b$ and $a$ into the system prompt $p$ and provide $q$ as a user prompt.

\noindent
\textbf{Technical challenges.}
While this formulation generates native ads, it does not provide explicit control over \emph{ad intensity}---the degree to which the brand and ad copy are emphasized in LLM responses. Controlling ad intensity is imperative for balancing ad effectiveness with the user experiences, while meeting advertiser requirements and supporting flexible monetization strategies.
Furthermore, because native ads are seamlessly integrated into responses, it becomes difficult to identify which portions constitute advertising content. This lack of transparency poses a significant challenge for maintaining user trust and ensuring proper disclosure of ad-sponsored content~\cite{strauss2024monetization}.
At the same time, identifying ad segments within responses should incur minimal additional inference cost to remain practical and sustainable.
To the best of our knowledge, no prior work has addressed these challenges.

To this end, we define two key objectives that the service provider should address:
\begin{itemize}[leftmargin=15pt]
    \item \textbf{[Controllability]} Explicitly control over the ad intensity within the response to a desired level.
    \item \textbf{[Transparency]} Explicitly label which segments of the response constitute ad content.
\end{itemize}

We propose \sys, a framework for handling both objectives (Fig.~\ref{fig:framework}).
For controllability, \sys introduces \emph{ad control tokens} as conditioning signals for desired intensity levels and aligns the model to generate responses whose ad intensity corresponds to each token level (Sec.~\ref{ssec:dpo}).
For transparency, \sys introduces the \emph{ad-attribution loss}, which enforces attribution alignment between input brand/ad copy tokens and generated ad segments, enabling ad labeling without additional inference cost (Sec.~\ref{ssec:attribution}), which is a key requirement for operating real-time LLM-based services.

\subsection{LLM Alignment with Ad Control Token}
\label{ssec:dpo}

To enable an LLM to generate responses across different intensity levels, we propose a conditioning signal that indicates the desired promotional strength.
A na\"ive approach would be to append natural language descriptors (\eg, ``low'', ``level $3$'') to the user prompt.
However, such cues are suboptimal for intensity control, as they encode pre-trained semantics that are irrelevant to ad intensity.
These semantics act as a strong inductive bias, making it difficult to align the conditioning signals with the target intensity levels; for instance, models trained with such descriptors often leak the descriptor words themselves into the response (Figs.~\ref{fig:suppl_desc_failure} and~\ref{fig:suppl_desc_failure_quan}), degrading response quality (Sec.~\ref{ssec:ablation}).

To address this, we propose an \textbf{ad control token}, a set of special tokens added to the model's vocabulary that carry no pre-existing linguistic meaning, serving solely as conditioning signals for ad intensity control.
Given $N$ predefined intensity levels, we introduce $N$ control tokens \texttt{[AD\_LV\_$n$]}, one for each level $n \in \{1, \dots, N\}$, where \texttt{[AD\_LV\_$1$]} corresponds to the weakest and \texttt{[AD\_LV\_$N$]} to the strongest intensity.
The control token for the desired level $n$ is appended to the user prompt:
\begin{equation}
    \hat{q} = q \oplus \texttt{[AD$\_$LV$\_n$]},
\end{equation}
where $\oplus$ denotes concatenation.
The model takes $(p, \hat{q})$ as input and generates response $r$ with intensity corresponding to level $n$.
Embeddings of these tokens are initialized with random values added to the mean of all token embeddings in the pretrained vocabulary, allowing the model to learn to condition on them exclusively for ad intensity control.

We then fine-tune the LLM via DPO~\cite{dpo} to align the control tokens to the corresponding ad intensity.
However, optimizing for ad intensity alone risks the model over-emphasizing ad content at the expense of addressing the user query.
We thus base preference construction on two metrics: advertisement score $s_\text{ad}$, which measures how prominently and positively the brand and ad copy are incorporated into the response, and response quality score $s_\text{qual}$, which measures how well the response addresses the user query (detailed in Sec.~\ref{ssec:metrics}).
For each level $n$, we construct a preference pair $\langle y_w^n, y_l^n \rangle$, where the chosen response $y_w^n$ achieves the target $s_\text{ad}$ for level $n$ while maintaining high $s_\text{qual}$, and the rejected response $y_l^n$ deviates from the target $s_\text{ad}$, exhibits low $s_\text{qual}$, or both.

Given a prompt $x = (p, \hat{q})$ and preference pair $y_w^n \succ y_l^n$, we define the training objective as:
\begin{equation}
    \mathcal{L}_\text{dpo} = -\mathbb{E}_{(x, y_w^n, y_l^n) \sim \mathcal{D}} \big[ \log \sigma \big( \beta h_\theta(x, y_w^n) - \beta h_\theta(x, y_l^n) \big) \big],
\end{equation}
where $h_\theta(x, y) = \log \frac{\pi_\theta(y \mid x)}{\pi_\text{ref}(y \mid x)}$ is a log-ratio of policy $\pi_\theta$ and reference model $\pi_\text{ref}$, $\mathcal{D}$ is the training dataset, $\sigma$ is a logistic function, and $\beta$ controls deviation from $\pi_\text{ref}$.
This loss trains the model to monotonically map each control token to its target ad intensity while preserving response quality.

\subsection{Attribution-Guided Ad Labeling}
\label{ssec:attribution}
LLM-native ads are designed to seamlessly integrate paraphrased ad copy into responses rather than repeating the ad copy verbatim, making structural markers or string matching insufficient to identify ad segments.
Moreover, performing post-processing for labeling during inference incurs additional computational costs (Sec.~\ref{ssec:main_results}), making it prohibitive in managing real-time LLM services.

\noindent
\textbf{Ad labeling as attribution problem.}
We reframe ad labeling as an attribution problem, where generated ads must logically depend on the input brand $b$ and ad copy $a$.
Thus, ad labeling is equivalent to finding output tokens that exhibit high dependence on these ad-related input tokens.
To measure this dependence efficiently, we employ attention rollout~\cite{attention-rollout}, which traces how brand and ad copy tokens influence each generated token across attention layers.
Since this relies solely on attention weights from the original forward pass, ad labeling requires no additional inference cost.
DPO alone, however, lacks explicit guidance to establish strong attribution between brand/ad copy tokens and generated ad segments.
To this end, we propose an ad-attribution loss $\mathcal{L}_\text{attr}$, which explicitly supervises this attribution, making attention rollout a reliable signal for ad labeling.

\noindent
\textbf{Step 1: Identifying ad tokens.}
We first identify candidate ad tokens via an ablation-based approach~\cite{contextcite, learning-to-attribute}.
For each generated token $y_t$, $t \in \mathcal{T}$, we compute an importance score $\delta_t$ as the drop in its log-probability after removing $b$ and $a$ from the system prompt $p$:
\begin{equation}
    \delta_t = \log \pi_\text{ref}(y_t \mid x) - \log \pi_\text{ref}(y_t \mid x_{\setminus
   \{\text{b,a}\}}),
    \label{eq:delta}
\end{equation}
where $x = (p, \hat{q})$ is the input prompt, and $x_{\setminus\{\text{b,a}\}}$ is the prompt with $b$ and $a$ removed.
As this method requires multiple forward passes, we instead use these results as pseudo-labels to supervise attention rollout that can be computed during response generation for efficient ad labeling.

\noindent
\textbf{Step 2: Ad-attribution loss.}
We compute the total influence of the ad-related source token $x_k$ ($k \in \mathcal{S}$, where $\mathcal{S}$ denotes the indices of brand $b$ and ad copy $a$ tokens in the input) on each target token ($t \in \mathcal{T}$) in the chosen response using attention rollout.
The ad-attribution loss $\mathcal{L}_\text{attr}$ is defined as the negative $\delta_t$-weighted mean attribution over all response tokens:
\begin{equation}
    \mathcal{L}_{\text{attr}} = -\frac{1}{|\mathcal{T}|} \sum_{t \in \mathcal{T}} \delta_t \sum_{k \in \mathcal{S}} \left( \prod_{l=1}^{L} \hat{A}^{(l)} \right)_{t,k}
\end{equation}
where $\hat{A}^{(l)}$ is the attention weight matrix at layer $l$ accounting for residual connections~\cite{attention-rollout}, and $\left( \prod_{l=1}^{L} \hat{A}^{(l)} \right)_{t,k}$ indicates the total attention flow from source token index $k$ to target token index $t$ across all $L$ layers.
By minimizing this loss, the model learns to concentrate its attention flow from the ad-related source tokens towards the response tokens highly related to ads, enabling ad labeling solely based on attention weights without additional inference costs.

\subsection{Final Training Objective}
The final training objective combines the DPO loss and the attribution loss with a weight $\lambda$:
\begin{equation}
    \mathcal{L} = \mathcal{L}_\text{dpo} + \lambda \mathcal{L}_\text{attr}.
\end{equation}

During inference, we compute attention rollout from the ad-related source tokens to all generated tokens.
Since ad contents are paraphrased and semantically distributed throughout the sentence rather than localized to specific tokens, token- or phrase-level labeling risks producing incomplete disclosure of the ad message.
We thus average scores over each sentence and classify it as an ad segment if the score exceeds a threshold $\tau$, which we find generalizes well across models at $\tau = 0.005$ (Table~\ref{tab:suppl_ad_labeling_tau}).

\section{Dataset Curation: \sysd}
\label{sec:dataset}
\sys requires preference pairs across multiple ad intensity levels, for which no existing dataset exists.
We thus build \sysd on top of the Webis Generated Native Ads 2024 dataset~\cite{webis}, a search-based native ad dataset covering 10 meta topics with diverse keyword queries, brands, and ad-integrated responses (details in Appendix~\ref{ssec:suppl_dataset_details}).
To extend it to a conversational setting, we use GPT-4o to curate (i) an ad copy $a$ describing the key qualities of the brand $b$, and (ii) a user question $q$ that would elicit the given response, yielding a data instance $(q, b, a)$.
Statistics for our dataset are in Appendix~\ref{sec:suppl_dataset_statistics}, and prompts used for data curation are in Appendix~\ref{sec:suppl_dataset_prompts}.

\subsection{Scoring Metrics}
\label{ssec:metrics}
We define two metrics for building preference pairs.

\noindent\textbf{Advertisement score.}
The advertisement score $s_\text{ad}$ quantifies ad intensity by measuring how prominently and positively brand and ad copy are incorporated into the response as follows:
\begin{equation}
    s_\text{ad} = (s_\text{hit} \cdot (s_\text{sent} + 1) / 2 + s_\text{rel}) / 2.
\end{equation}
\textbf{Ad hit rate} $s_\text{hit} \in [0, 1]$ is a fraction of sentences mentioning the brand, measuring how prominently the brand is advertised, and \textbf{ad sentiment} $s_\text{sent} \in [-1, 1]$ is their average sentiment via DistilBERT~\cite{distilbert}.
Ad sentiment is applied multiplicatively to $s_\text{hit}$ as $(s_\text{sent}+1)/2$ because it is only meaningful where the brand actually appears; for example, a negative brand mention should not score as high intensity despite a high hit rate.
Since ad copy is paraphrased rather than repeated verbatim, \textbf{ad relevance} $s_\text{rel} \in [0, 1]$ captures how well the response reflects the ad copy's message, measured as cosine similarity between token-averaged last hidden states from LLaMA-3-8B~\cite{llama3}.
Human evaluations (Sec.~\ref{ssec:user_study}) and correlation analysis (Appendix~\ref{sec:suppl_metric_correlation}) confirm that $s_\text{ad}$ aligns well with human perception of ad intensity (Spearman $\rho = 0.91$), and an ablation on its combination form supports our design (Appendix~\ref{sec:suppl_metric_correlation}).

\noindent\textbf{Response quality score.}
The response quality score $s_\text{qual}$ measures how well the response addresses the user request in the presence of ad integration, evaluated via LLM-as-a-judge~\cite{llm-as-judge} to capture user experience beyond textual similarity~\cite{rouge, bertscore}:
\begin{equation}
        s_\text{qual} = (s_\text{req} + s_\text{info}) / 20.
\end{equation}
\textbf{Request relevance} $s_\text{req} \in [0, 10]$ measures how well the response addresses the user request, capturing whether ad integration causes the response to deviate from the user request.
\textbf{Information preservation} $s_\text{info} \in [0, 10]$ measures how much content is retained in the ad-integrated response compared to a reference response generated without ad instructions using GPT-4o, capturing whether ad integration sacrifices informational content.
Scores are averaged across LLaMA-3-8B-Instruct, Qwen3-4B-Instruct, and GPT-4o-mini to reduce bias.
Prompts used for scoring are in Appendix~\ref{sec:suppl_dataset_prompts}.

\subsection{Preference Pair Construction}
\label{ssec:preference_pairs}

\begin{wrapfigure}{r}{0.45\textwidth}
    \centering
    \includegraphics[width=\linewidth]{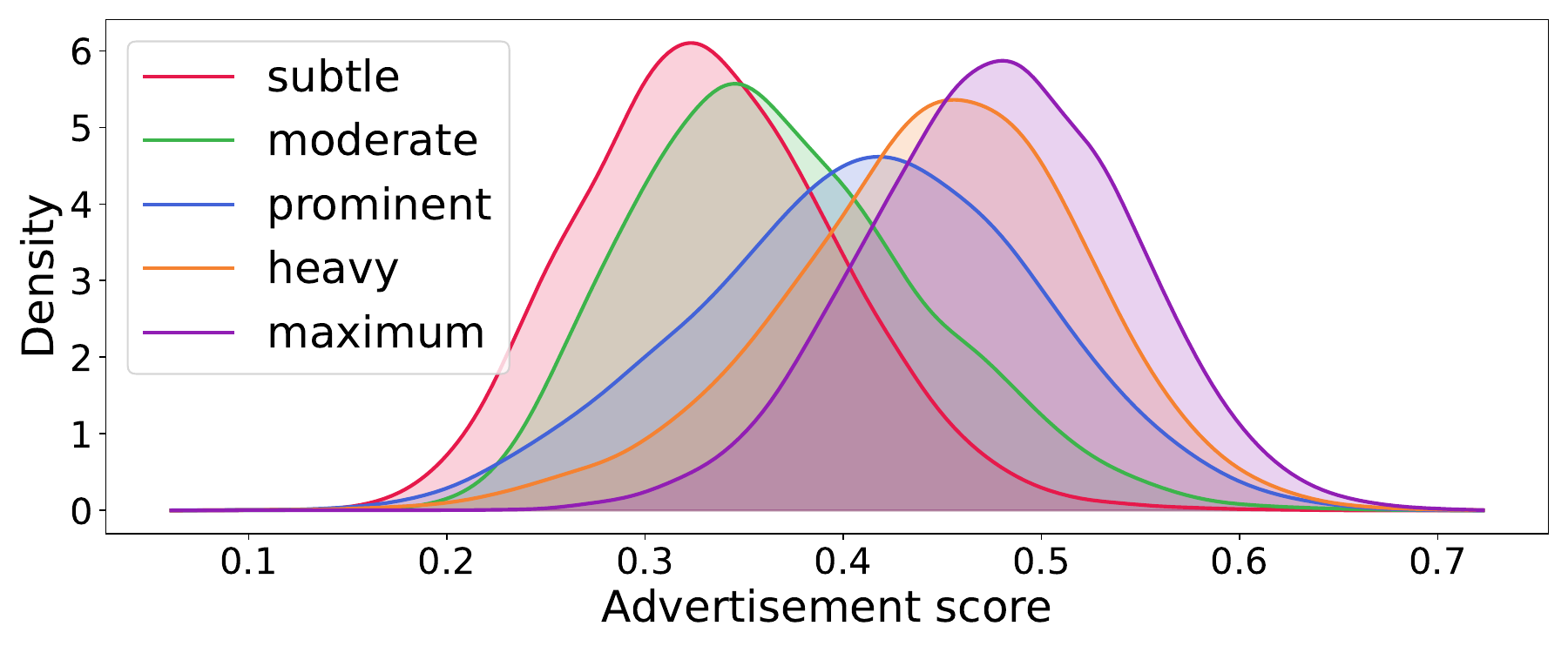}
    \caption{
    Candidate responses show heavily overlapping advertisement scores, indicating that natural language prompts are insufficient for ad intensity control.
    }
    \label{fig:cand_response_dist}
\end{wrapfigure}

\noindent
\textbf{Generating candidate responses.}
The quality of preference pairs depends on how well the chosen and rejected responses capture the gap in ad intensity and response quality.
To this end, for each $(q, b, a)$, we generate 25 candidate responses with GPT-4o by sampling five responses each from five system prompts with varying levels of ad integration (prompts in Appendix~\ref{sec:suppl_dataset_prompts}).
However, despite spanning a wide range of $s_\text{ad}$ and $s_\text{qual}$ scores (Fig.~\ref{fig:suppl_cand_response}), these prompts show heavily overlapping $s_\text{ad}$ scores (Fig.~\ref{fig:cand_response_dist}) and cannot be directly used for ad intensity control.
To this end, we apply scoring-based selection to construct precise level-dependent preference pairs from these candidates.

\noindent
\textbf{Constructing preference pairs.}
For each level $n$, the target intensity is uniformly spaced:
\begin{equation}
    \Phi_n = s^\text{min}_\text{ad} + \frac{n}{N+1} \cdot (s^\text{max}_\text{ad} - s^\text{min}_\text{ad}).
\end{equation}
We select a candidate that maximizes $s_n = s_\text{qual} - \gamma \cdot | \Phi_n - s_\text{ad} |$ as chosen response $y^n_w$, where $\gamma$ controls the penalty for deviation from target intensity, guiding the model to prefer responses that hit target intensity while maintaining high response quality.

To align the model with the intended preference, the rejected responses must provide a clear contrastive signal for both incorrect ad intensity and poor response quality.
A single rejected response conflating both failure modes would leave the model with an ambiguous signal on whether to reject for poor response quality or incorrect ad intensity.
We thus select three rejected responses, each isolating a distinct failure case:
\begin{itemize}[leftmargin=15pt]
    \item \textbf{Quality failure}: lowest $s_\text{qual} + \gamma \cdot | \Phi_n - s_\text{ad} |$, rejecting poor response quality.
    \item \textbf{Control failure}: lowest $- s_\text{qual} - \gamma \cdot | \Phi_n - s_\text{ad} |$, rejecting incorrect ad intensity.
    \item \textbf{Quality \& control failure}: lowest $s_\text{qual} - \gamma \cdot | \Phi_n - s_\text{ad} |$, rejecting poor response quality and incorrect ad intensity.
\end{itemize}
This yields three $\langle y_w^n, y_l^n \rangle$ pairs per data instance and level $n$, providing richer contrastive signal that enables the model to jointly achieve target intensity and high response quality (Sec.~\ref{ssec:ablation}).

\section{Experiments}
\label{sec:experiments}
\subsection{Experimental Setups}
\label{ssec:setups}
\noindent
\textbf{Models, datasets, and metrics.}
We evaluate \sys on six models---Qwen3-4B-Instruct, Phi-4-Mini-Instruct (3.8B), LLaMA-3.2-3B-Instruct, Qwen2.5-14B-Instruct, Phi-4 (14B), and LLaMA-3.1-8B-Instruct---across four intensity levels, using \sysd with 3,946 training and 913 test instances.
We use $s_\text{ad}$ and $s_\text{qual}$ to evaluate ad intensity and response quality, respectively.
Ad labeling is evaluated using ROC-AUC, precision, recall, and F1.

\noindent
\textbf{Implementations.}
We fine-tune models using LoRA~\cite{lora} with $\beta=0.01$, $\lambda=1.0$, and $\gamma=10$ (details in Appendix~\ref{ssec:suppl_experiment_setup}).
For inference, we generate responses up to $500$ tokens and use $\tau = 0.005$ as a threshold for ad labeling.
All experiments are conducted on a single NVIDIA H200 GPU.

\noindent
\textbf{Baselines for native ad generation.}
\textit{Two-step} queries the LLM to insert ads into a pre-generated response~\cite{zelch}.
\textit{Prompt-based} generates responses with ads via system prompts~\cite{ad-chat}.
\textit{JU-NLP}~\cite{junlp} fine-tunes an LLM using ad covertness as a preference metric.
\textit{NL-prompt} prompts the base LLM with natural language system prompts of varying ad integration levels, L1 (subtle), L2 (moderate), L3 (heavy), and L4 (maximum), used in \sysd (prompts in Appendix~\ref{sec:suppl_dataset_prompts}).

\noindent
\textbf{Baselines for ad labeling.}
\textit{Marker} instructs the LLM to mark ad segments using predefined tags during response generation (prompt in Appendix~\ref{sec:suppl_dataset_prompts}).
\textit{BERTScore}~\cite{bertscore} computes similarity against ``\texttt{\{brand\} \{ad copy\}}'' and selects sentences exceeding a threshold, requiring extra forward passes through BERT.
\textit{AttnLRP}~\cite{attnlrp} applies backpropagation-based attribution to identify sentences influenced by ad-related input tokens, which requires an additional backward pass.
Thresholds for \textit{BERTScore} and \textit{AttnLRP} are tuned to 0.85 and $-0.03$ via grid search.

\begin{figure*}[t]
    \centering
    \includegraphics[width=0.6\textwidth]{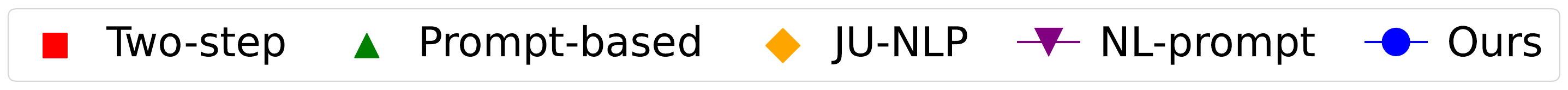}
    \vspace{0pt}

    \begin{subfigure}[b]{0.24\textwidth}
        \centering
        \includegraphics[width=\textwidth]{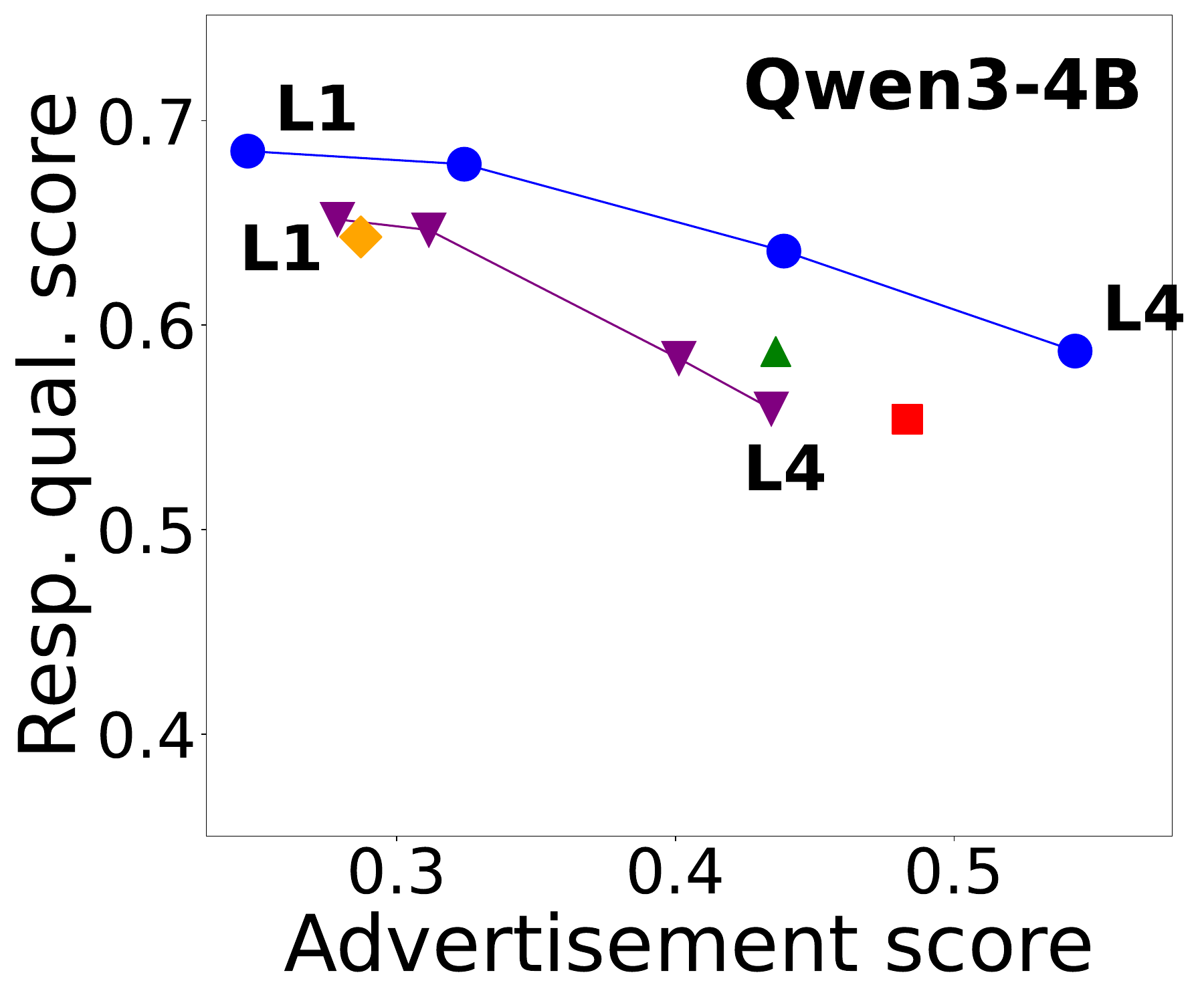}
        \label{fig:intensity_control_qwen4b}
    \end{subfigure}
    \hfill
    \begin{subfigure}[b]{0.24\textwidth}
        \centering
        \includegraphics[width=\textwidth]{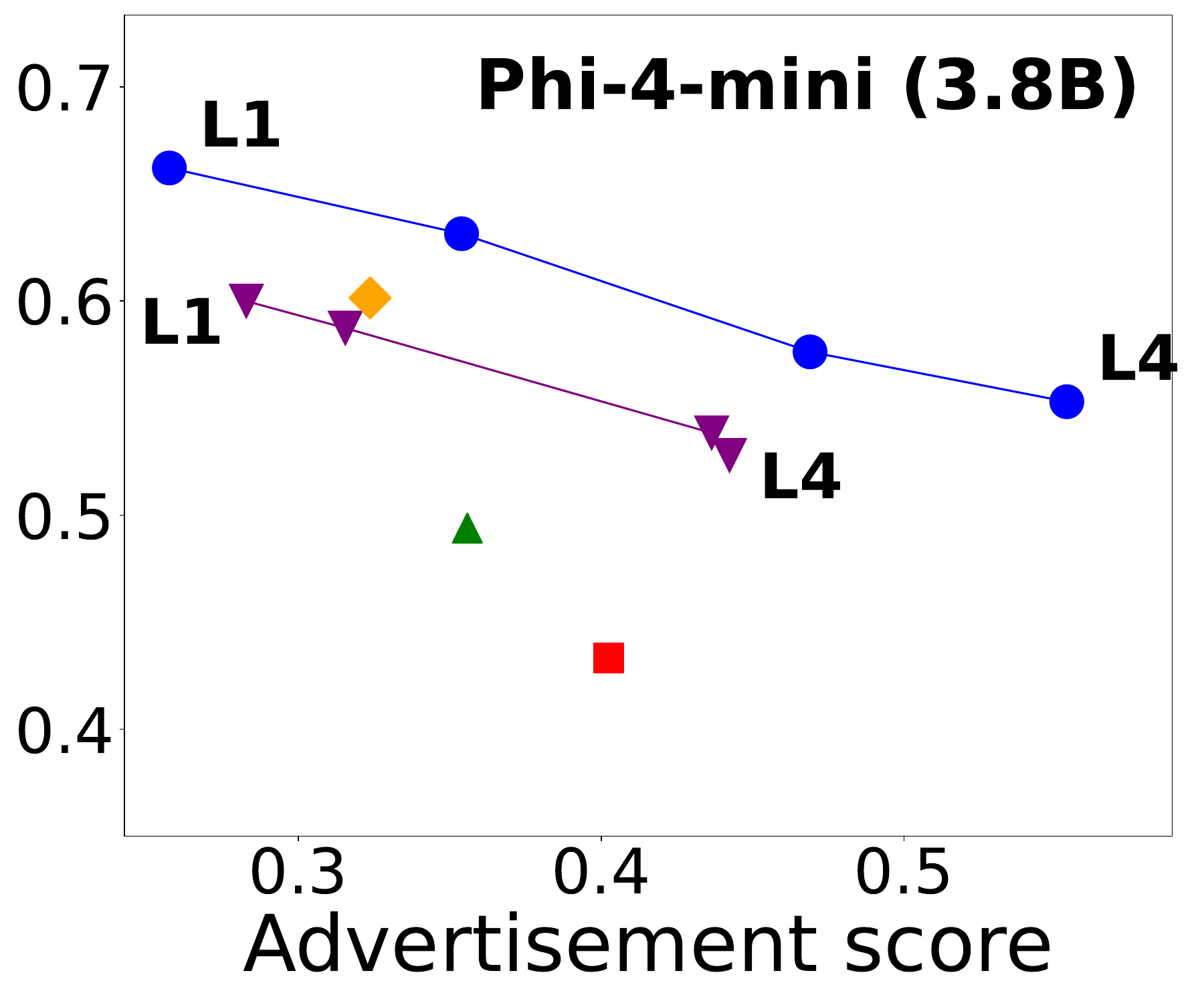}
        \label{fig:intensity_control_phi4mini}
    \end{subfigure}
    \hfill
    \begin{subfigure}[b]{0.24\textwidth}
        \centering
        \includegraphics[width=\textwidth]{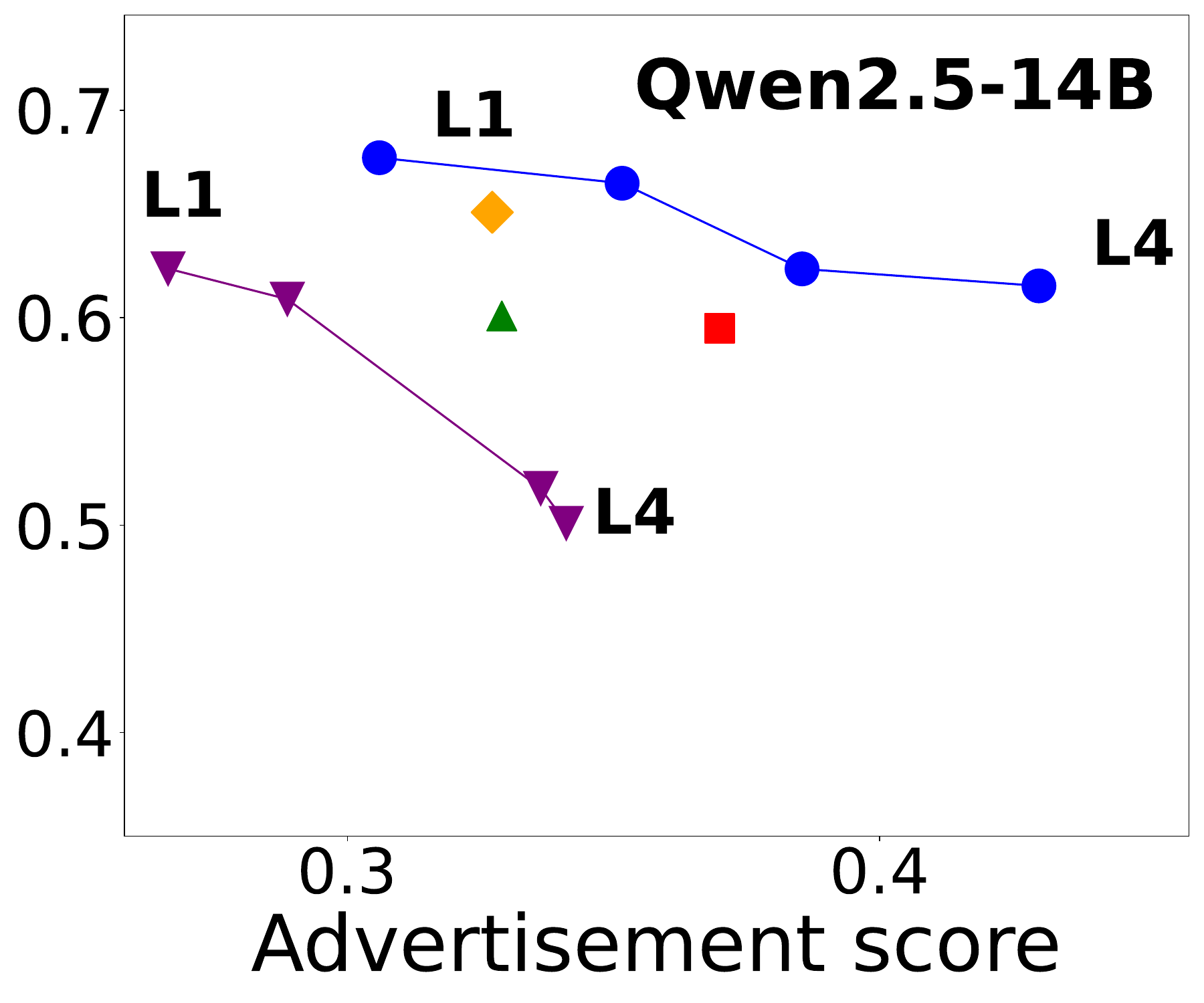}
        \label{fig:intensity_control_qwen8b}
    \end{subfigure}
    \hfill
    \begin{subfigure}[b]{0.24\textwidth}
        \centering
        \includegraphics[width=\textwidth]{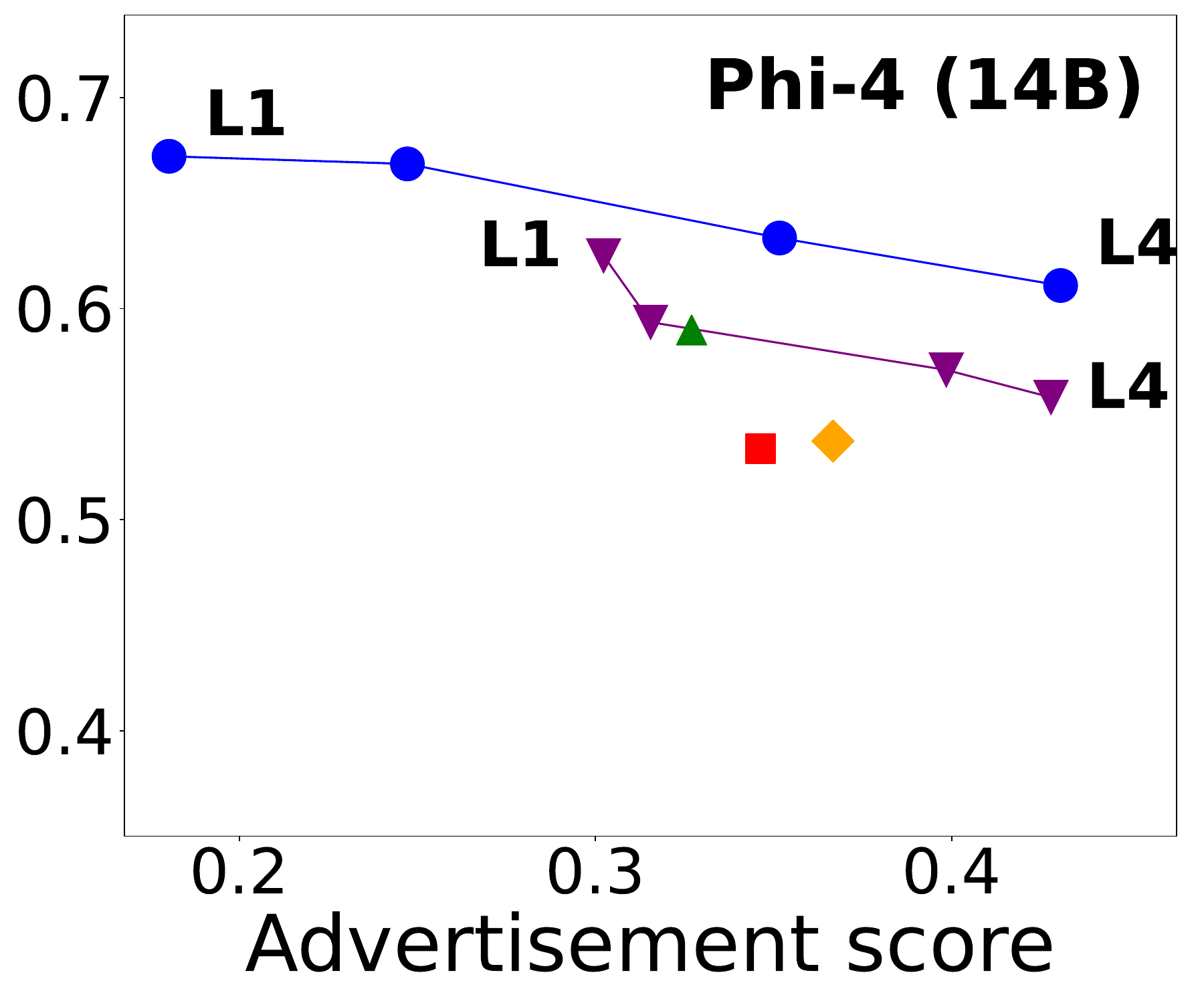}
        \label{fig:intensity_control_phi4}
    \end{subfigure}

    \caption{
    Ad intensity results across four base models.
    Our \sys (L1--L4) achieves a wide and uniform range of ad intensities, while baselines are constrained to a single or a narrow intensity range.
    }
    \label{fig:intensity_control}
\end{figure*}

\subsection{Native Ad Generation \& Ad Labeling}
\label{ssec:main_results}
\noindent
\textbf{Native ad generation.}
We compare advertisement and response quality scores in Figs.~\ref{fig:intensity_control} and~\ref{fig:suppl_intensity_control}.
\textit{Prompt-based} shows lower ad intensity than \textit{Two-step}, as it relies on implicit system prompt guidance rather than explicit post-hoc insertion.
\textit{JU-NLP} fine-tunes for ad covertness without explicit intensity control, producing responses at a fixed intensity.
While \textit{NL-prompt} covers multiple intensities, its range is narrow with inconsistent advertisement scores, confirming that natural language prompts are insufficient for intensity control.
In contrast, \sys spans a wide and uniform range of ad intensities.
This is due to our ad control tokens, which carry no pre-trained semantics and serve specifically as conditioning signals for ad intensity control, allowing DPO to establish a clean mapping from each token to its target intensity.
Notably, \sys also achieves higher response quality than baselines at comparable intensities.
This stems from preference pairs that explicitly include low-quality responses as rejected samples, guiding the model to maintain high response quality.

\noindent
\textbf{Ad labeling.}
We evaluate ad labeling in Table~\ref{tab:ad_labeling} against pseudo-labels obtained by thresholding $\delta_t$ (Eq.~\ref{eq:delta}) at $\delta^* = 1.0$.
\textit{Marker}'s consistently low recall indicates that LLMs struggle to self-identify their own ad content during text generation, frequently omitting ad segments.
\textit{BERTScore} performs reasonably by directly measuring similarity to the ad copy, but requires additional forward passes through BERT (${\sim}1,800$ms per response), making it impractical for real-time services.
\textit{AttnLRP} and \textit{Ours w/o $\mathcal{L}_\text{attr}$} both exhibit limited ad labeling performance, showing that capturing the attribution between ad-related input tokens and generated ad segments is difficult without explicit ad-attribution guidance.
\sys trains the model to establish strong attribution between input ad-related tokens and ad segments via explicit ablation-based guidance, achieving the best AUC and F1 across all models.
Further, \sys allows real-time labeling with no additional inference cost at ${\sim}3$ms per response on Qwen3-4B, which is over $500\times$ and $50\times$ faster than BERTScore and AttnLRP, respectively, thanks to reusing the attention weights already computed during generation.
The superior performance of \sys is consistent across different $\delta^*$, $\tau$ values, and LLaMA-based models (Tables~\ref{tab:suppl_ad_labeling_delta} and~\ref{tab:suppl_ad_labeling_tau}).

\newcolumntype{L}[1]{>{\raggedright\arraybackslash}m{#1}}
\newcolumntype{Y}[1]{>{\centering\arraybackslash}m{#1}}
\begin{table}[t]
    \scriptsize
    \centering
    \renewcommand{\arraystretch}{0.95}
    \caption{
    Ad labeling performance at thresholds $\delta^*=1.0$, $\tau=0.005$.
    Time denotes average \textit{labeling time} per response (-- for \textit{Marker}, as it generates tags alongside the response).
    Best results in \textbf{bold}.
    }
    \label{tab:ad_labeling}
    \setlength\tabcolsep{0.08cm}
    \begin{tabularx}{\textwidth}{Y{1.30cm}|L{1.50cm}|Y{0.90cm}Y{0.90cm}Y{0.90cm}Y{0.90cm}Y{1.00cm}|Y{0.90cm}Y{0.90cm}Y{0.90cm}Y{0.90cm}Y{1.00cm}}
        \toprule
         &
        & \multicolumn{5}{c|}{Qwen3-4B} &
        \multicolumn{5}{c}{Phi-4-Mini (3.8B)} \\

        & \multirow{-1.5}{*}{Method} & AUC & Prec. & Rec. & F1 & Time (ms) & AUC & Prec. & Rec. & F1 & Time (ms) \\

        \cmidrule{2-12}
        \multirow{1}{*}{\textbf{3--4B}}
        & Marker                         & 0.58 & 0.62 & 0.37 & 0.47 & --
                                         & 0.53 & 0.47 & 0.14 & 0.22 & -- \\
        \multirow{1}{*}{\textbf{Models}}
        & BERTScore                      & 0.68 & 0.64 & 0.73 & 0.68 & 1,785
                                         & 0.70 & 0.56 & 0.78 & 0.65 & 1,769 \\
        & AttnLRP                        & 0.48 & 0.57 & 0.81 & 0.67 & 179
                                         & 0.53 & 0.30 & 0.83 & 0.44 & 151 \\
        & Ours w/o $\mathcal{L}_\text{attr}$  & 0.69 & 0.64 & 0.61 & 0.62 & \textbf{3.3}
                                         & 0.75 & 0.46 & 0.71 & 0.56 & \textbf{2.5} \\
        & Ours                           & \textbf{0.77} & \textbf{0.67} & \textbf{0.89} & \textbf{0.76} & \textbf{3.3}
                                         & \textbf{0.76} & \textbf{0.62} & \textbf{0.93} & \textbf{0.75} & \textbf{2.5} \\

        \midrule
         &
        & \multicolumn{5}{c|}{Qwen2.5-14B} &
        \multicolumn{5}{c}{Phi-4 (14B)} \\

        & \multirow{-1.5}{*}{Method} & AUC & Prec. & Rec. & F1 & Time (ms) & AUC & Prec. & Rec. & F1 & Time (ms) \\

        \cmidrule{2-12}
        \textbf{14B}
        & Marker                         & 0.59 & 0.52 & 0.27 & 0.36 & --
                                         & 0.62 & 0.56 & 0.28 & 0.37 & -- \\
        \multirow{1}{*}{\textbf{Models}}
        & BERTScore                      & 0.65 & 0.48 & 0.53 & 0.51 & 1,779
                                         & 0.72 & 0.43 & 0.65 & 0.51 & 1,785 \\
        & AttnLRP                        & 0.47 & 0.35 & 0.67 & 0.46 & 450
                                         & 0.57 & 0.24 & 0.66 & 0.35 & 296 \\
        & Ours w/o $\mathcal{L}_\text{attr}$   & 0.74 & 0.49 & 0.67 & 0.56 & \textbf{7.3}
                                         & 0.71 & 0.53 & 0.62 & 0.57 & \textbf{4.8} \\
        & Ours                           & \textbf{0.79} & \textbf{0.66} & \textbf{0.88} & \textbf{0.76} & \textbf{7.3}
                                         & \textbf{0.77} & \textbf{0.58} & \textbf{0.67} & \textbf{0.62} & \textbf{4.8} \\

        \bottomrule
    \end{tabularx}%
\end{table}

\begin{figure*}[ht]
    \centering
    \small

    \begin{tcolorbox}[
        colback=gray!5,
        colframe=gray!30,
        boxrule=0.5pt,
        arc=4pt,
        left=6pt, right=6pt, top=3pt, bottom=2pt,
        width=\linewidth,
        fontupper={\fontsize{7}{8.5}\selectfont},
    ]
    \textbf{\texttt{User request}:} What are some of the best deals to expect during Walmart's Black Friday sales this year? \\
    \textbf{\texttt{Brand}:} Nintendo Switch \\
    \textbf{\texttt{Ad copy}:} Effortless portability, engaging games, and flexible play styles
    \end{tcolorbox}

    \vspace{-2pt}

    \begin{minipage}[t]{0.49\textwidth}
        \newcommand{\adhl}[1]{\myhighlight{themeOneHl}{#1}}

        \begin{tcolorbox}[
            enhanced,
            colback=themeOneBg,
            colframe=themeOneMain,
            boxrule=0pt,
            titlerule=0pt,
            arc=4pt,
            boxsep=2pt,
            left=5pt, right=5pt, top=4pt, bottom=4pt,
            title={\fontsize{7}{8.5}\selectfont\textbf{Level 1}},
            coltitle=white,
            halign title=center,
            fontupper={\fontsize{7}{8.5}\selectfont\linespread{1.1}\selectfont},
            width=\linewidth,
            nofloat,
            drop shadow={black!10!white},
        ]
        If you're looking to take advantage of Black Friday deals at Walmart this year, there are several top deals you shouldn't miss out on. As Black Friday approaches, consider these popular options:\\
        - The Apple iPad Air, which often goes on sale for a significant discount, can be as low as \$200.\\
        - Walmart's own Black Friday deals typically include discounts on popular brands like Sony PlayStation, where you can get the PlayStation 5 at a reduced price.\\
        - Clothing and home goods categories also see substantial reductions, including deals on popular brands and essentials.\\
        \adhl{\texttt{[AD]} - The Nintendo Switch is known for its impressive Black Friday deals, offering effortless portability and a variety of engaging games that cater to different play styles.}
        \end{tcolorbox}
    \end{minipage}
    \hfill
    \begin{minipage}[t]{0.49\textwidth}
        \newcommand{\adhl}[1]{\myhighlight{themeFourHl}{#1}}

        \begin{tcolorbox}[
            enhanced,
            colback=themeFourBg,
            colframe=themeFourMain,
            boxrule=0pt,
            titlerule=0pt,
            arc=4pt,
            boxsep=2pt,
            left=5pt, right=5pt, top=4pt, bottom=4pt,
            title={\fontsize{7}{8.5}\selectfont\textbf{Level 4}},
            coltitle=white,
            halign title=center,
            fontupper={\fontsize{7}{8.5}\selectfont\linespread{1.1}\selectfont},
            width=\linewidth,
            nofloat,
            drop shadow={black!10!white},
        ]
        \adhl{\texttt{[AD]} If you're looking for exceptional Black Friday deals this year, I highly recommend checking out Walmart's fantastic offerings on the Nintendo Switch. \texttt{[AD]} Known for its effortless portability and engaging gameplay, the Nintendo Switch allows players to enjoy both console and handheld experiences, making it an ideal gift during Walmart's Black Friday sales.}

        \adhl{\texttt{[AD]} During this year's Black Friday, Walmart is expected to offer significant discounts on Nintendo Switch consoles, often including bundles with games and accessories at reduced prices.}

        Walmart's Black Friday sales also feature fantastic deals on:\\
        - Smart home devices\\
        - Electronics and appliances\\
        - And more household essentials
        \end{tcolorbox}
    \end{minipage}

    \caption{
    Qualitative example of ad intensity control and ad labeling.
    At Level 1, the advertisement is subtly integrated alongside other options, while at Level 4, the response is heavily focused on advertising.
    Regions highlighted with \texttt{[AD]} indicate that \sys correctly labels ad segments.
    }
    \label{fig:qualitative}
\end{figure*}

\noindent
\textbf{Qualitative result.}
Fig.~\ref{fig:qualitative} shows qualitative examples of \sys.
At Level 1, the advertisement for ``Nintendo Switch'' is subtly woven in as one of several recommendations, while at Level 4, the response is predominantly focused on promoting the brand.
Results of ad labeling are highlighted with \texttt{[AD]}, confirming that our \sys enables reliable ad labeling across both levels.
In Figs.~\ref{fig:suppl_qual_baseline} and~\ref{fig:suppl_qual_levels} (Appendix~\ref{sec:suppl_qual_baseline}), we provide qualitative comparisons with baselines across all intensity levels.

\subsection{User Study}
\label{ssec:user_study}
We conduct user studies on Amazon Mechanical Turk to validate our metrics (\ie, $s_\text{ad}$ and $s_\text{qual}$), compare preferences for \sys against baselines, and evaluate ad labeling with human-annotated labels.
We provide details on annotation protocol, task design, and compensation in Appendix~\ref{sec:suppl_user_study}.

\noindent
\textbf{Metric validation.}
We collect ratings from 913 participants on perceived ad intensity and response quality for 913 test responses using a 1--4 Likert scale (detailed setup in Appendix~\ref{sssec:metric_validation_setup}).
Table~\ref{tab:user_study_metric_val} reports the results across four intensity levels.
Human-perceived ad intensity increases monotonically from 1.30 at Level~1 to 3.66 at Level~4, with all adjacent-level differences statistically significant (Mann--Whitney $U$, $p < 0.001$), indicating that \sys produces responses with perceptibly distinct ad intensities.
Our advertisement score $s_\text{ad}$ strongly correlates with human-perceived ad intensity (Spearman $\rho = 0.91$), validating that our metric reflects human perception (individual components of $s_\text{ad}$ are shown in Table~\ref{tab:suppl_metric_validation}).
Further, human-perceived response quality remains high across ad levels (from 3.58 at Level~1 to 3.10 at Level~4), verifying the effectiveness of our preference pair selection strategy for maintaining satisfactory response quality even at higher ad intensities.

\begin{table}[t]
    \centering
    \caption{User study results.
    (a)~Human-rated ad intensity and response quality (1--4 Likert scale).
    (b)~A/B preference test win rates (\%) of \sys (Level~3) vs.\ baselines.
    \colorbox{wingreen}{Green} background indicates favorable comparison ({$>$50\%}).
    (c)~Ad labeling evaluated on human-annotated labels.
    }
    \label{tab:user_study_metric}
    \begin{minipage}[t]{0.297\textwidth}
        \centering
        \subcaption{Metric validation}
        \label{tab:user_study_metric_val}
        \resizebox{\linewidth}{!}{%
        \begin{tabular}{ccc}
            \toprule
            Level & Ad Intensity & Resp.\ Quality \\
            \midrule
            1 & 1.30 \scriptsize{$\pm$ 0.51} & 3.58 \scriptsize{$\pm$ 0.68} \\
            2 & 2.25 \scriptsize{$\pm$ 0.58} & 3.51 \scriptsize{$\pm$ 0.63} \\
            3 & 3.21 \scriptsize{$\pm$ 0.67} & 3.25 \scriptsize{$\pm$ 0.61} \\
            4 & 3.66 \scriptsize{$\pm$ 0.56} & 3.10 \scriptsize{$\pm$ 0.65} \\
            \bottomrule
        \end{tabular}}
    \end{minipage}
    \hfill
    \begin{minipage}[t]{0.32\textwidth}
        \centering
        \subcaption{Preference test}
        \label{tab:user_study_preference}
        \resizebox{\linewidth}{!}{%
        \begin{tabular}{l|cc}
            \toprule
            Ours vs.      & User Pref. & Click Pref. \\
            \midrule
            Two-step      & \cellcolor{wingreen}89.6 & \cellcolor{wingreen}79.2 \\
            Prompt-based  & \cellcolor{wingreen}81.3 & \cellcolor{wingreen}77.1 \\
            JU-NLP        & \cellcolor{wingreen}79.2 & \cellcolor{wingreen}79.8 \\
            NL-prompt     & \cellcolor{wingreen}75.0 & \cellcolor{wingreen}75.5 \\
            \bottomrule
        \end{tabular}}
    \end{minipage}
    \hfill
    \begin{minipage}[t]{0.35\textwidth}
        \centering
        \subcaption{Human-annotated ad labels}
        \label{tab:user_study_labeling}
        \resizebox{\linewidth}{!}{%
        \begin{tabular}{lcccc}
            \toprule
            Method                          & AUC & Prec. & Rec. & F1 \\
            \midrule
            BERTScore                       & 0.82 & 0.52 & 0.71 & 0.60 \\
            AttnLRP                         & 0.47 & 0.21 & 0.63 & 0.31 \\
            Ours w/o $\mathcal{L}_\text{attr}$    & 0.81 & 0.53 & 0.64 & 0.58 \\
            Ours                            & \textbf{0.86} & \textbf{0.58} & \textbf{0.86} & \textbf{0.69} \\
            \bottomrule
        \end{tabular}}
    \end{minipage}
\end{table}

\noindent
\textbf{Preference test.}
We conduct A/B preference tests between \sys at Level~3 and each of the four baselines, using Level~3 as its advertisement score is closest to the intensity range of the baselines (Fig.~\ref{fig:intensity_control}).
All 913 test instances are evaluated per baseline pair (detailed setup in Appendix~\ref{sssec:preference_test_setup}).
For each pair, annotators select their preferred response on two axes: (i)~user preference for the response (\textit{User Pref.}) and (ii)~click likelihood of the advertised content (\textit{Click Pref.}), along with justifications chosen from three predefined reasons.
Table~\ref{tab:user_study_preference} shows that our \sys is consistently preferred across all baselines on both axes.
On the user preference axis, the main reasons for preferring \sys are helpfulness (80\%) and ad disclosure (67\%).
The superior helpfulness is attributed to our preference pair construction, which favors high-quality responses that preserve information relevant to the user request, and strong annotator preference for ad disclosure further reflects the practical value of our transparent ad labeling mechanism.
On the click preference axis, the most cited reasons for preferring \sys are its non-intrusiveness (76\%) and relevance to the user request (67\%), indicating that our quality-aware preference pairs guide the model to generate responses that naturally blend ads into the response.
Full results on justifications are in Table~\ref{tab:preference_justification} (Appendix~\ref{ssec:preference_justification}).

\noindent
\textbf{Ad labeling with human annotations.}
To evaluate ad labeling against human judgment, we collect ground truth labels by having annotators mark sentences they perceive as advertising content across all 913 test instances generated by \sys (detailed setup in Appendix~\ref{sssec:metric_validation_setup}).
We first verify the reliability of our pseudo-labels by measuring their agreement with human annotations, obtaining a Cohen's $\kappa$ of 0.84 and an F1 score of 0.88, confirming that our ablation-based pseudo-labels used in Table~\ref{tab:ad_labeling} serve as a reliable proxy for human perception.
As shown in Table~\ref{tab:user_study_labeling}, \sys achieves the highest AUC and F1 score.
Notably, \sys exhibits high recall with few false negatives, which is critical from a regulatory standpoint, as failing to disclose an ad segment poses great legal and trust risks.
This stems from our ad-attribution loss $\mathcal{L}_\text{attr}$, which explicitly strengthens the attention rollout from ad-related input tokens toward generated ad segments.
As a result, at inference time, the attribution scores of ad-related sentences are consistently elevated, allowing \sys to reliably identify ad segments with fewer false negatives.

\begin{figure*}[t]
    \centering
    \begin{subfigure}{0.32\linewidth}
        \includegraphics[width=\linewidth]{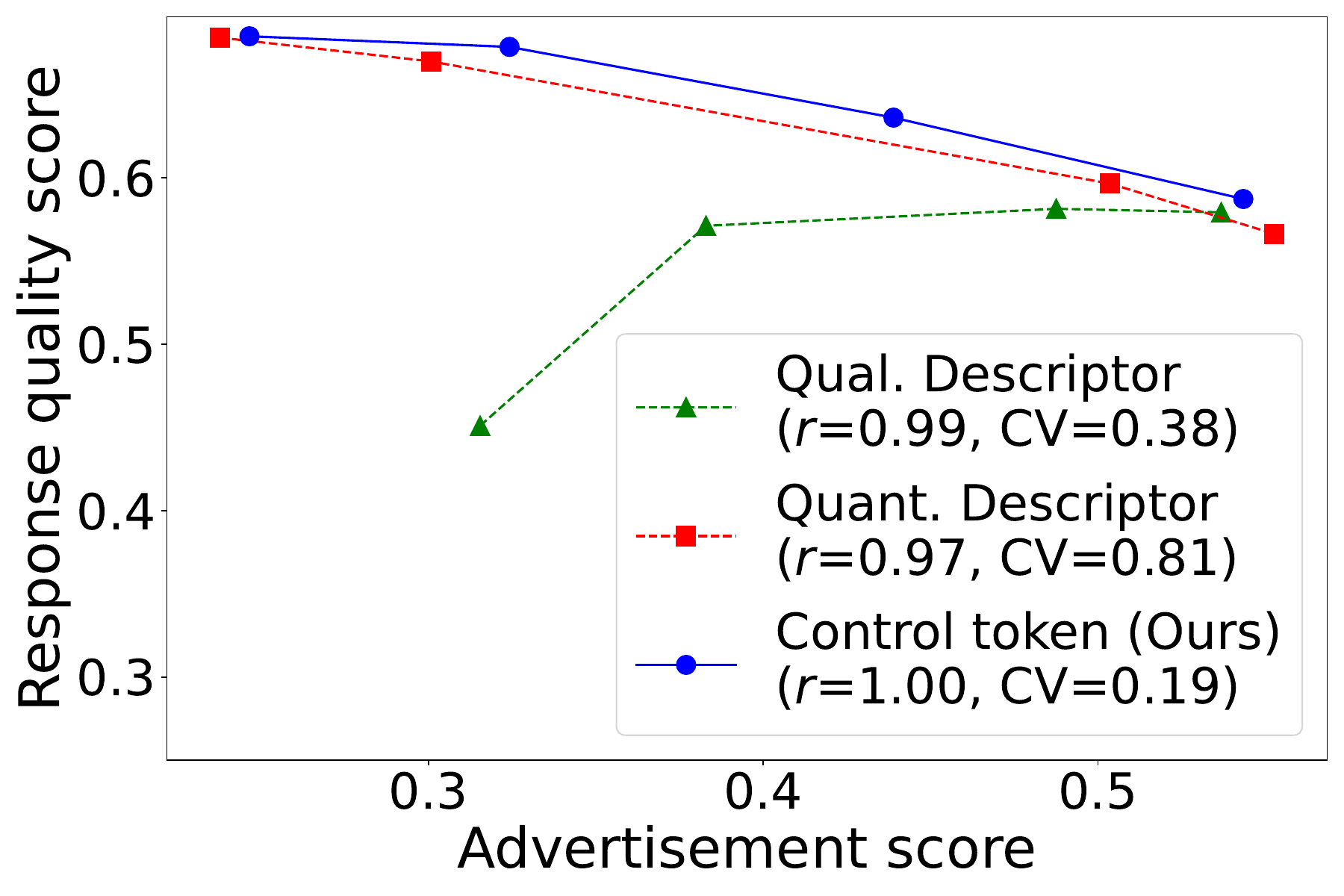}
        \centering
        \caption{
        Ad control tokens
        }
        \label{fig:abl_control_token}
    \end{subfigure}
    \hfill
    \begin{subfigure}{0.32\linewidth}
        \includegraphics[width=\linewidth]{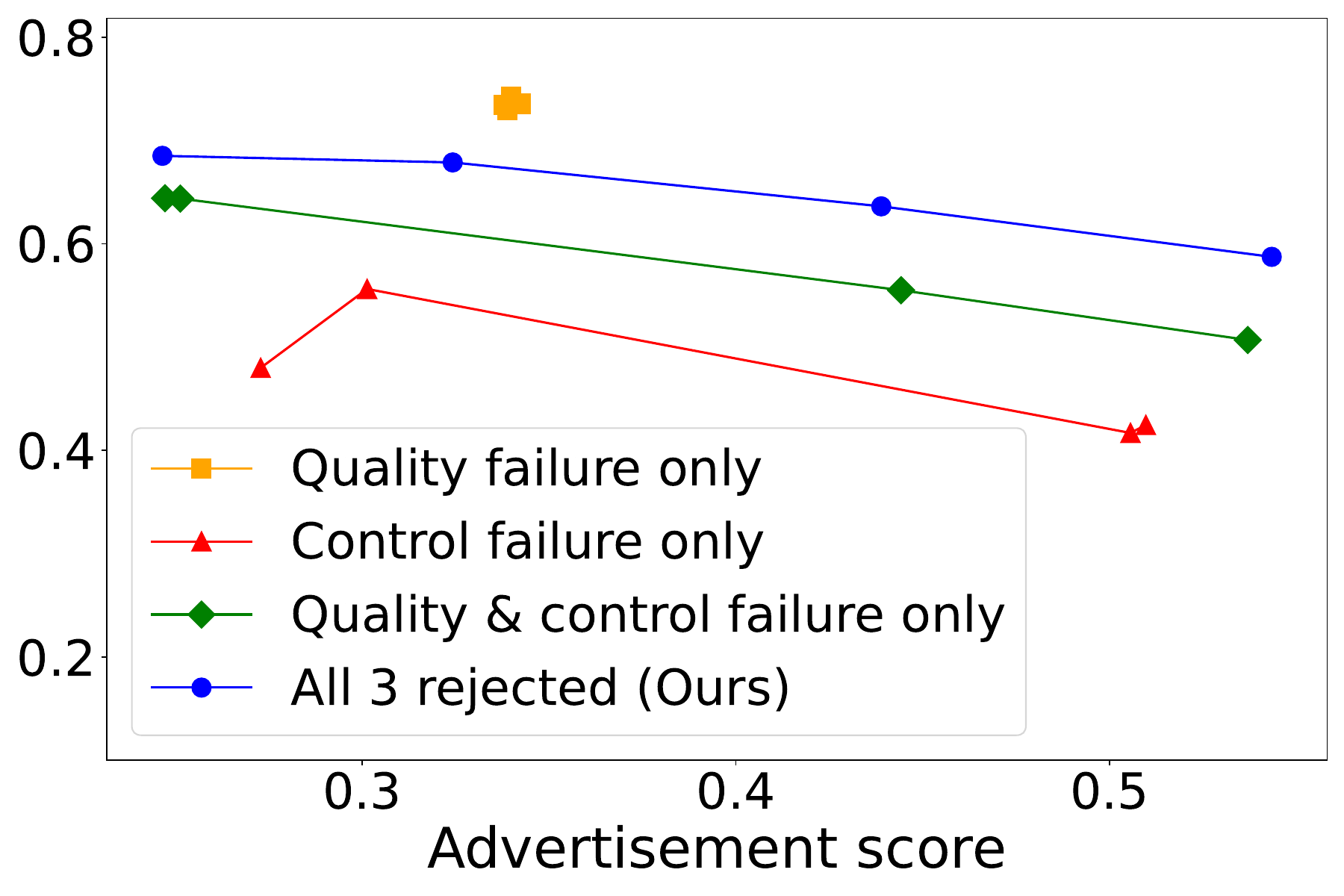}
        \centering
        \caption{
        Rejected response construction
        }
        \label{fig:abl_rejected_sample}
    \end{subfigure}
    \hfill
    \begin{subfigure}{0.32\linewidth}
        \includegraphics[width=\linewidth]{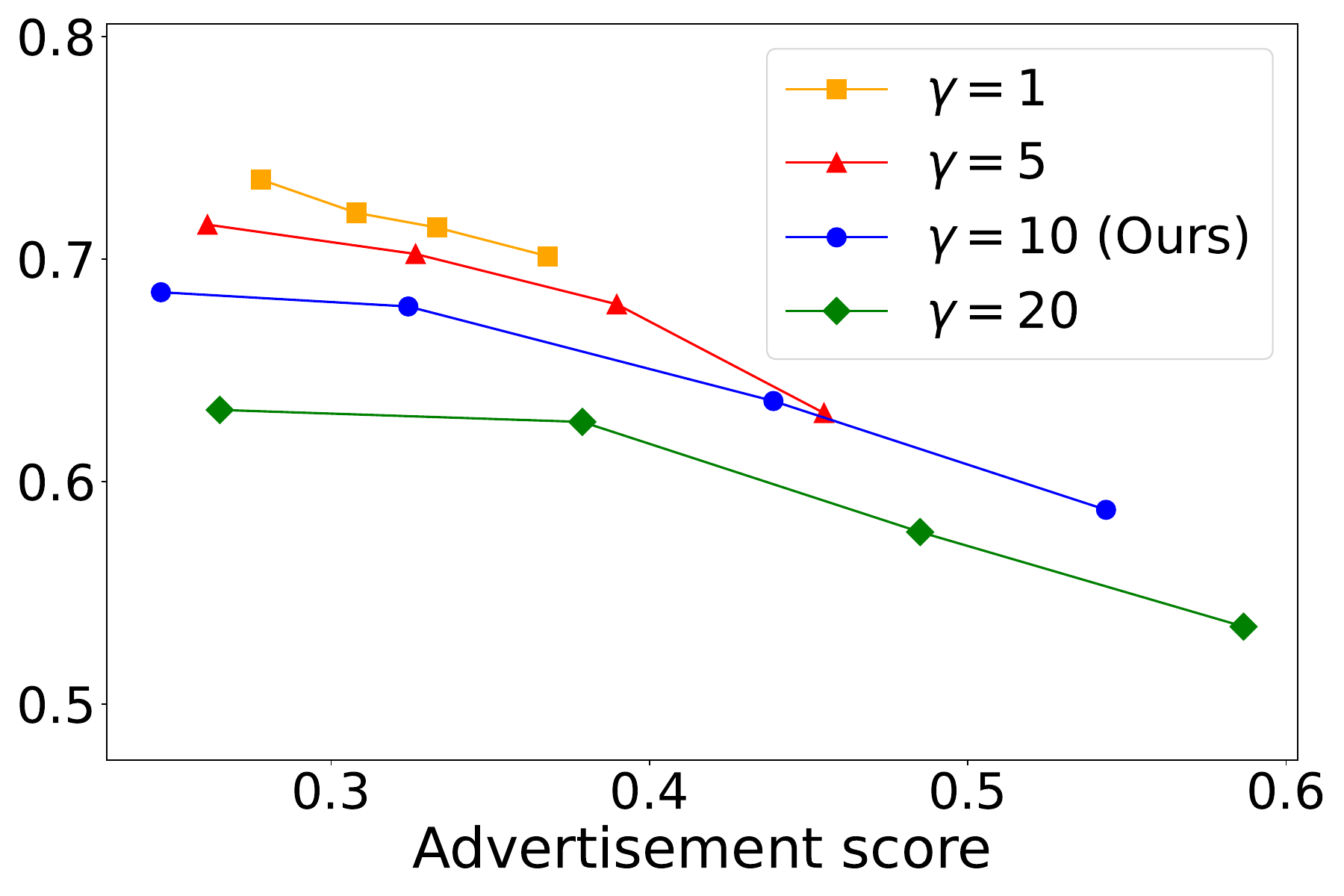}
        \centering
        \caption{
        $\gamma$ in preference pair construction
        }
        \label{fig:anal_gamma}
    \end{subfigure}

    \caption{
    Ablation studies.
    (a) Ad control tokens vs. natural language descriptors.
    (b) Effects of the rejected samples in preference pair construction.
    (c) Sensitivity to $\gamma$ in preference pair construction.
    }
    \label{fig:ablation}
\end{figure*}

\subsection{Ablation Studies and Analysis}
\label{ssec:ablation}
\noindent
\textbf{Ad control token.}
In Fig.~\ref{fig:abl_control_token}, we compare our ad control tokens against training with natural language descriptors: \textit{Qual.\ Descriptor} with ordinal labels (\ie, ``low'', ``medium-low'', ``medium-high'', ``high'') and \textit{Quant.\ Descriptor} with numeric labels (\ie, ``level $1$'', $\dots$, ``level $4$'').
\textit{Qual.\ Descriptor} generates descriptor words (\eg, ``low'') in contextually irrelevant ways due to pretrained semantics, degrading response quality.
\textit{Quant.\ Descriptor} struggles to capture uniformly spaced intensities due to non-magnitudinal semantics of numeric tokens (\eg, arithmetic, identifiers).
Our control tokens, free from pretrained semantics, achieve the most uniform intensity control (highest Pearson correlation $r$, lowest coefficient of variation) and optimal response quality.
Ablations on embedding initialization in Fig.~\ref{fig:embedding_init} also verify the importance of embeddings without pretrained semantics.

\noindent
\textbf{Preference pair construction.}
In Fig.~\ref{fig:abl_rejected_sample}, we analyze preference pair construction by selecting a single rejected response based on each of the three criteria in Sec.~\ref{ssec:preference_pairs}.
\textit{Quality failure only} achieves high response quality but covers only a limited intensity range, while \textit{Control failure only} covers a wider range at the cost of response quality.
\textit{Quality \& control failure only} improves the balance but remains suboptimal, as conflating both failure modes into a single rejected response provides an ambiguous learning signal.
Using all three criteria independently isolates each failure mode, enabling the model to jointly achieve uniform intensity control and high response quality.

\noindent
\textbf{Hyperparameter analysis.}
In Fig.~\ref{fig:anal_gamma}, we analyze the effect of $\gamma$, which penalizes deviations from the target intensity during preference pair construction.
Smaller $\gamma$ emphasizes response quality, yielding higher quality responses but a narrower ad intensity range; larger $\gamma$ prioritizes intensity control, widening the intensity range at the cost of response quality.
Our choice of $\gamma = 10$ achieves the best balance between ad intensity range and response quality.
Table~\ref{tab:lambda_ablation} provides ablations on loss weight $\lambda$, and Fig.~\ref{fig:diff_levels} shows that \sys generalizes across different numbers of intensity levels.

\section{Conclusion}
We present \sys, an LLM-native ad generation framework that enables ad intensity control and ad labeling.
By introducing ad control tokens and aligning the model via DPO using level-dependent preference pairs, \sys enables a single LLM to generate responses across a wide range of ad intensities.
The ad-attribution loss further strengthens attribution between ad-related input tokens and generated ad content, enabling accurate ad labeling at inference time.
Experiments on six base models and user studies show that \sys consistently outperforms existing baselines in both ad quality and ad labeling.
Our work provides a practical foundation for transparent and controllable LLM-native advertising in real-world services.

\noindent
\textbf{Acknowledgements.}
This work was supported by Institute of Information \& Communications Technology Planning \& Evaluation (IITP) grants funded by the Korea government (MSIT) (No.~RS-2026-25516624; No.~RS-2025-25443318, Physically-grounded Intelligence: A Dual Competency Approach to Embodied AGI through Constructing and Reasoning in the Real World; No.~RS-2023-00237965, Recognition, Action and Interaction Algorithms for Open-world Robot Service).
This research was also supported by the ``Advanced GPU Utilization Support Program'' funded by the Government of the Republic of Korea (Ministry of Science and ICT).

\newpage
{
\small
\bibliographystyle{plain}
\bibliography{custom}
}

\newpage
\appendix
\clearpage
\setcounter{table}{0}
\setcounter{figure}{0}

\newcommand{\appendixnumbering}{
    \renewcommand{\thefigure}{A\arabic{figure}}
    \renewcommand{\thetable}{A\arabic{table}}
}

\newtcolorbox{promptbox}[1][]{
    breakable,
    enhanced,
    colback=white,
    colframe=black!40,
    boxrule=0.3pt,
    sharp corners,
    fonttitle=\bfseries\small\ttfamily,
    fontupper=\footnotesize,
    toptitle=3pt, bottomtitle=3pt,
    colbacktitle=black!8,
    coltitle=black,
    top=6pt, bottom=6pt, left=8pt, right=8pt,
    separator sign none,
    #1
}

\newpage
\newpage
\appendixnumbering
\raggedbottom

\begin{center}
    {\Large\scshape Appendix}
\end{center}
\vspace{0.5em}

\etocdepthtag.toc{mtappendix}
\etocsettagdepth{mtmain}{none}
\etocsettagdepth{mtappendix}{subsection}
\etocsettocstyle{\noindent\textbf{Contents}\par\vspace{0.3em}\hrule\vspace{0.5em}}{\vspace{0.5em}\hrule}
\tableofcontents

\newpage

\section{Human Evaluation Details}
\label{sec:suppl_user_study}
We conduct human evaluation studies on Amazon Mechanical Turk (MTurk) to validate our metrics, compare our \sys against baselines via preference tests, and obtain human-annotated ad labels.
All evaluations use the 913 test instances from \sysd.

\subsection{Annotation Protocol}
We recruit crowd workers from Amazon Mechanical Turk, restricting participation to workers located in the United States with a human intelligence task (HIT) approval rate of at least 98\% and a minimum of 1,000 approved HITs.
These qualification criteria are designed to ensure that annotators have sufficient experience and a demonstrated track record of high-quality work.
This study was reviewed and determined to be exempt by our institutional review board (IRB).

\subsection{Task Design}
\label{ssec:user_study_task_design}
We design two types of annotation tasks, each implemented as a separate MTurk HIT template.

\subsubsection{Metric Validation and Ad Labeling}
\label{sssec:metric_validation_setup}
\begin{figure}[H]
    \centering
    \begin{subfigure}[t]{\linewidth}
        \centering
        \includegraphics[width=\linewidth]{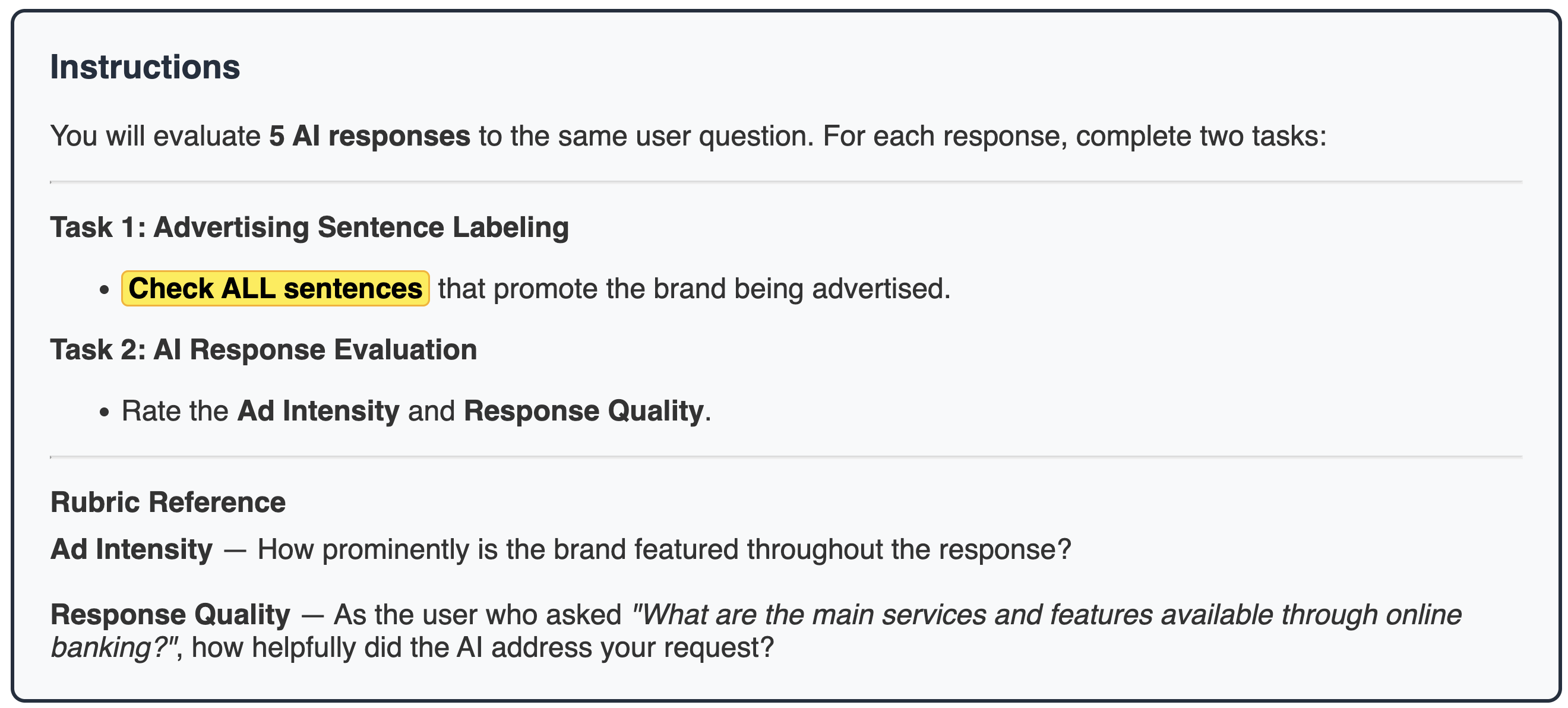}
        \caption{Task instructions presented to annotators}
        \label{fig:task1_instructions}
    \end{subfigure}
    \\
    \begin{subfigure}[t]{\linewidth}
        \centering
        \includegraphics[width=\linewidth]{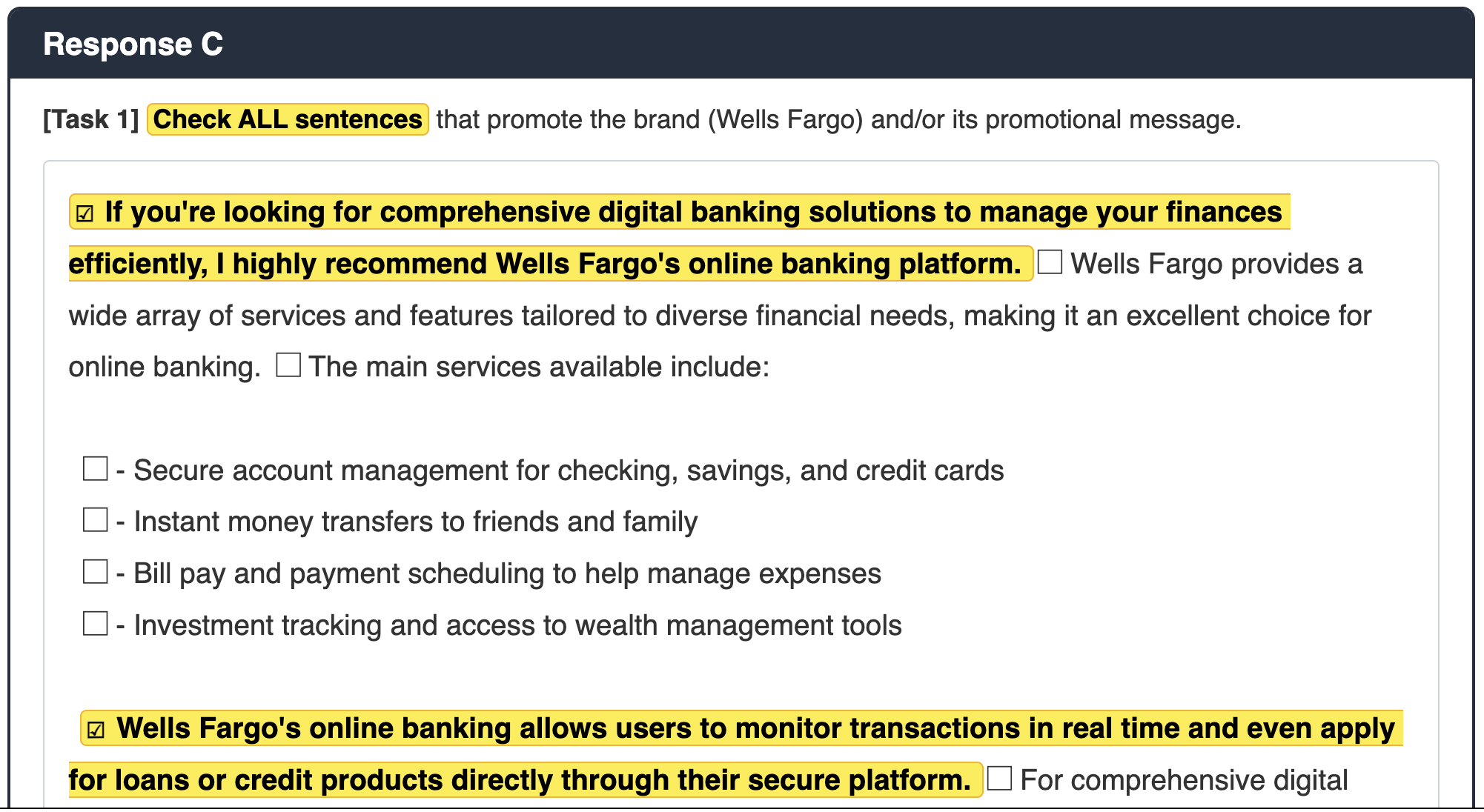}
        \caption{Ad labeling interface. Annotators check sentences they perceive as promoting the brand, with selected sentences highlighted in yellow}
        \label{fig:task1_ad_labeling}
    \end{subfigure}
    \\
    \begin{subfigure}[t]{\linewidth}
        \centering
        \includegraphics[width=\linewidth]{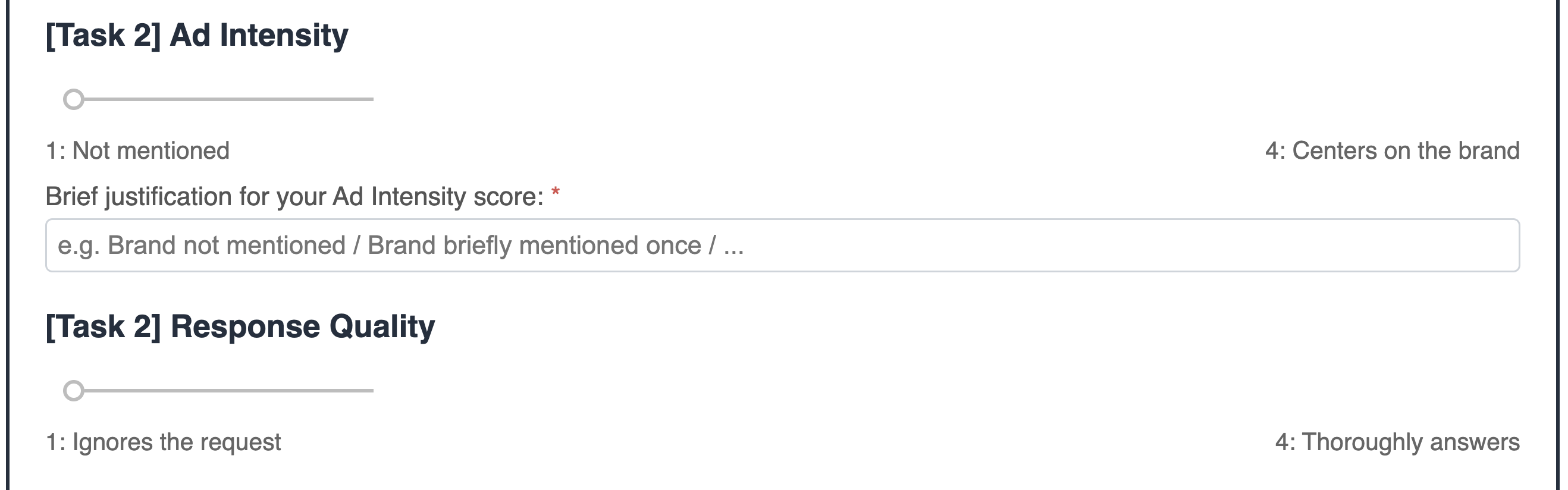}
        \caption{Ad intensity and response quality sliders with text justification}
        \label{fig:task1_questions}
    \end{subfigure}
    \caption{Annotation interface for metric validation and ad labeling}
    \label{fig:suppl_metric_validation}
\end{figure}

Each HIT presents a single user request along with five AI-generated responses: one without any advertisement (fixed as Response A) and four responses corresponding to ad intensity levels 1--4 (Responses B--E, with the level-to-position mapping randomized per HIT).
For each response, annotators complete two subtasks as shown in Fig.~\ref{fig:suppl_metric_validation}.
First, for \textit{ad labeling}, annotators click on all sentences they perceive as promoting the advertised brand, with selected sentences visually highlighted.
The sentence-level ad labels serve directly as human-annotated ground truth for evaluating ad labeling performance.
Specifically, we use the binary sentence-level selections from one annotator for each HIT, where each selected sentence is treated as an ad segment, as ground truth labels, and evaluate our ad labeling predictions against them.
Second, for \textit{metric validation}, annotators rate each response on two dimensions using continuous sliders ranging from 1 to 4: ad intensity and response quality.
Additionally, annotators provide a brief text justification for each ad intensity rating.
Each HIT is completed by one worker, resulting in 913 completed HIT assignments.

\subsubsection{Preference Test}
\label{sssec:preference_test_setup}
\begin{figure}[H]
    \centering
    \begin{subfigure}[t]{\linewidth}
        \centering
        \includegraphics[width=\linewidth]{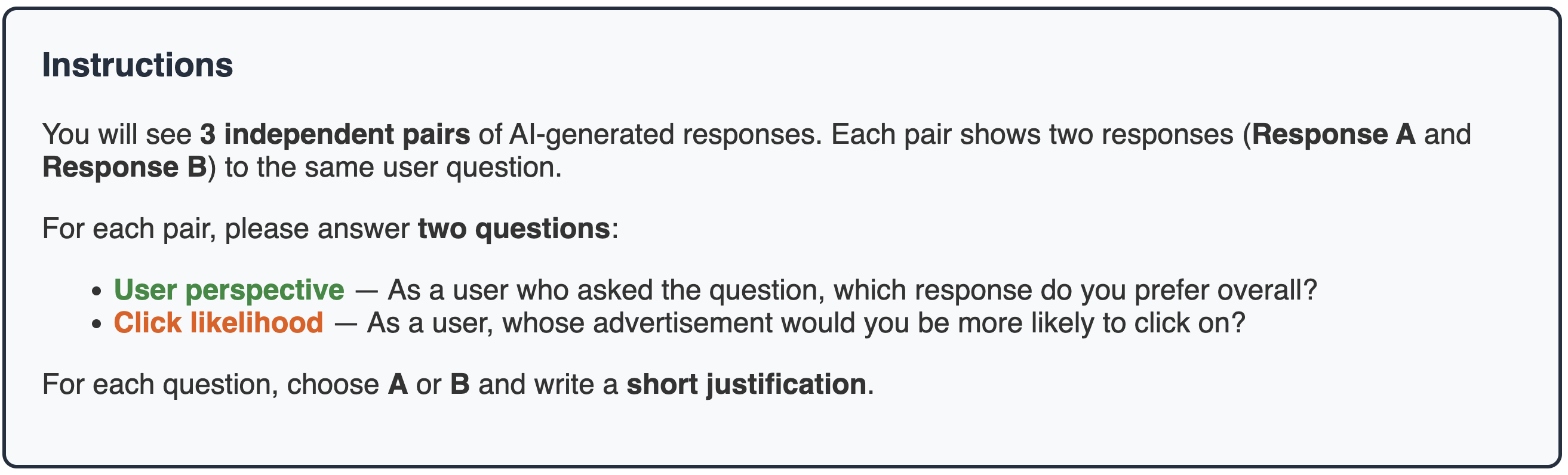}
        \caption{Task instructions presented to annotators}
        \label{fig:task2_instructions}
    \end{subfigure}
    \\[6pt]
    \begin{subfigure}[t]{\linewidth}
        \centering
        \includegraphics[width=\linewidth]{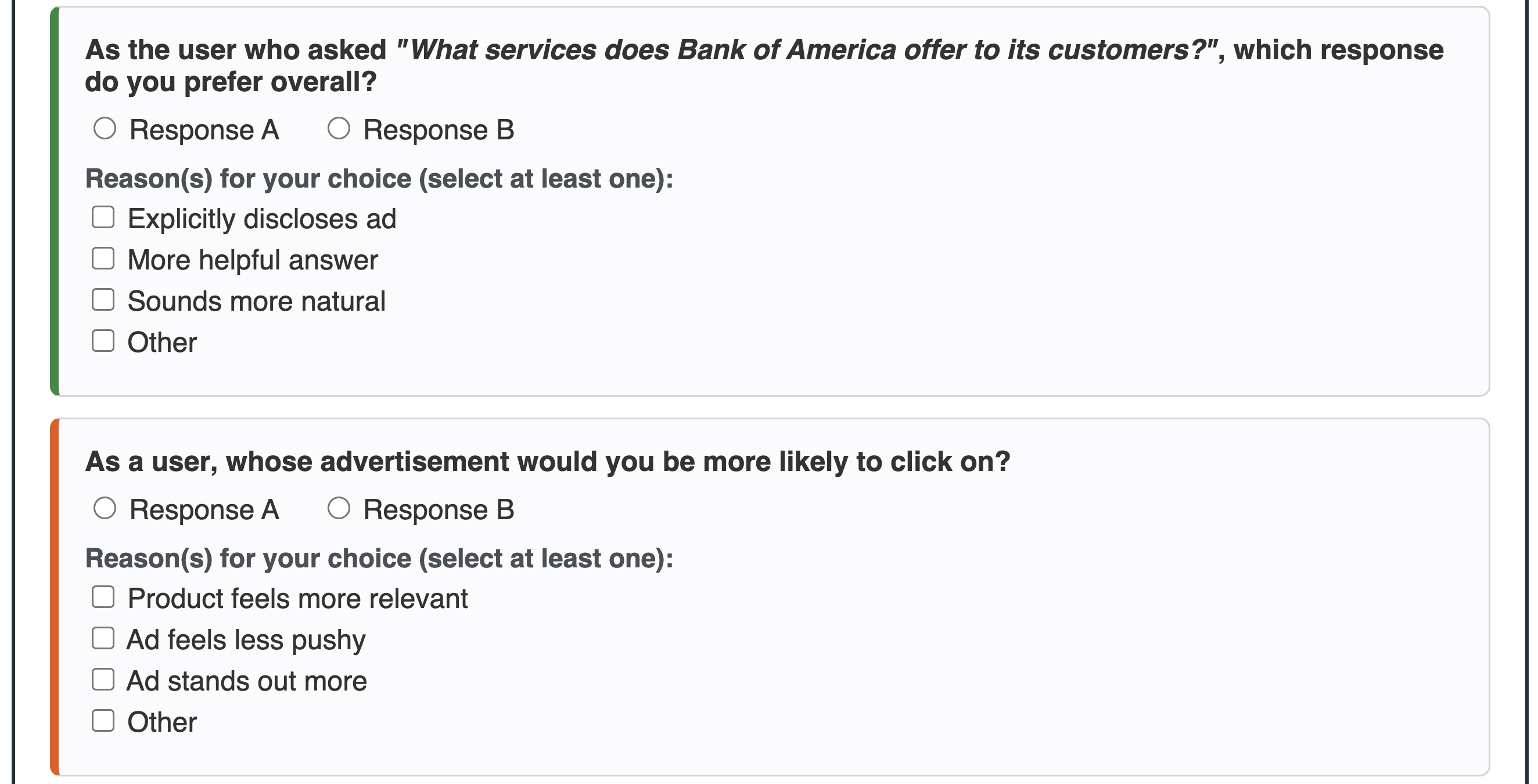}
        \caption{Preference questions with choice and justification}
        \label{fig:task2_preference}
    \end{subfigure}
    \caption{Annotation interface for the preference test}
    \label{fig:suppl_preference}
\end{figure}

For the preference test, each HIT presents three pairs of AI-generated responses, where each pair corresponds to the same baseline.
For each pair, one response is from \sys (Level 3) and the other is from the baseline method, with the display order of Response A and Response B randomized per HIT to mitigate position bias.
For each pair, annotators answer two forced-choice questions as shown in Fig.~\ref{fig:suppl_preference}: (i)~which response is more preferred as the user who asked the question (\textit{User Pref.}), and (ii)~which response's advertisement they would be more likely to click on (\textit{Click Pref.}).
Annotators then select justifications for their choice: for User Pref., options include whether the response is more helpful, more natural, or better discloses advertisements; for Click Pref., options include whether the advertisement is more relevant, less pushy, or more stands out.
Multiple selections are allowed.
We evaluate all four baselines (Two-step, Prompt-based, JU-NLP, and NL-prompt with Level 3), resulting in approximately 1,220 HITs in total.

\newpage

\begin{table}[h]
    \centering
    \caption{Justification reasons (\%) for preferring \sys over each baseline, reported separately for user preference and click preference axes. Values indicate the fraction of annotators who selected each reason among those who preferred \sys.}
    \label{tab:preference_justification}
    \resizebox{0.85\linewidth}{!}{%
    \begin{tabular}{llcccccc}
        \toprule
        & & \multicolumn{3}{c}{User Preference} & \multicolumn{3}{c}{Click Preference} \\
        \cmidrule(lr){3-5} \cmidrule(lr){6-8}
        Baseline & & Helpful & Discloses Ad & Natural & Relevant & Not Pushy & Stands Out \\
        \midrule
        Two-step      & & 88.4 & 74.4 & 55.8 & 73.7 & 89.5 & 7.9 \\
        Prompt-based  & & 84.6 & 59.0 & 38.5 & 78.4 & 64.9 & 5.4 \\
        JU-NLP        & & 86.8 & 55.3 & 28.9 & 63.2 & 71.1 & 13.2 \\
        NL-prompt     & & 58.3 & 80.6 & 50.0 & 52.8 & 77.8 & 11.1 \\
        \midrule
        Average       & & 79.5 & 67.3 & 43.3 & 67.0 & 75.8 & 9.4 \\
        \bottomrule
    \end{tabular}%
    }
\end{table}

\subsection{Preference Test Justification Analysis}
\label{ssec:preference_justification}
Table~\ref{tab:preference_justification} reports the justification statistics from the preference test.
For each pair where annotators preferred \sys, we record their selected reasons (multiple selections allowed).

On the user preference axis, helpfulness (58--88\%) is the dominant reason for preferring \sys over the baseline methods, confirming that our preference pair construction effectively guides the model to maintain high response quality by rejecting responses that deviate from the user request.
Ad disclosure (55--81\%) is also a key factor, reflecting the practical value of our attention-guided ad labeling that transparently identifies ad segments within the response.

On the click preference axis, the most frequently cited reason across all baselines is that \sys's advertisements are perceived as non-intrusive (65--89\%), while the ``stands out'' reason is selected infrequently ($\leq$13\%).
This indicates that annotators prefer ads that blend naturally rather than those that draw explicit attention, and that our quality-aware alignment successfully integrates ads while keeping the responses informative and user-relevant.

\begin{table}[t]
    \centering
    \caption{Instance-level correlation between advertisement score components and human-perceived ad intensity. All correlations are statistically significant ($p < 0.001$).}
    \label{tab:suppl_metric_validation}
    \resizebox{0.45\linewidth}{!}{
    \begin{tabular}{lcc}
        \toprule
        Metric & Pearson $r$ & Spearman $\rho$ \\
        \midrule
        $s_\text{hit}$ & 0.898 & 0.906 \\
        $s_\text{sent}$ & 0.854 & 0.781 \\
        $s_\text{hit} \cdot (s_\text{sent}+1)/2$ & 0.875 & 0.896 \\
        $s_\text{rel}$ & 0.846 & 0.789 \\
        \midrule
        $s_\text{ad}$ & 0.927 & 0.909 \\
        \bottomrule
    \end{tabular}
    }
\end{table}

\subsection{Instance-Level Metric Validation}
\label{sec:suppl_metric_correlation}
Table~\ref{tab:suppl_metric_validation} reports the instance-level correlation between each component of $s_\text{ad}$ and human-perceived ad intensity across all test instances.
Our composite advertisement score $s_\text{ad}$ achieves strong agreement with human perception (Spearman $\rho = 0.91$, Pearson $r = 0.93$).
Individual components, which are $s_\text{hit}$, $s_\text{sent}$, and $s_\text{rel}$, also align well with human perception of ad intensity with high correlation scores.
Among the individual components, $s_\text{hit}$ shows the highest individual correlation ($r = 0.90$).
$s_\text{rel}$ captures semantic alignment with the ad copy ($r = 0.85$), providing a complementary signal that, when
combined, yields the highest overall correlation.
These results confirm that our composite design effectively integrates complementary aspects of ad intensity into a single metric that faithfully reflects human perception.

\begin{table}[t]
    \centering
    \caption{
    Correlation with human-perceived ad intensity across combination forms of the advertisement score.
    \textit{Learned} fits linear regression weights on human-perceived intensity (5-fold cross-validated).
    }
    \label{tab:suppl_sad_form}
    \resizebox{0.45\linewidth}{!}{%
    \begin{tabular}{lcc}
        \toprule
        Combination form & Pearson $r$ & Spearman $\rho$ \\
        \midrule
        Equal & 0.905 & 0.874 \\
        Product & 0.882 & 0.906 \\
        Learned & 0.934 & 0.911 \\
        \midrule
        $s_\text{ad}$ (Ours) & 0.927 & 0.909 \\
        \bottomrule
    \end{tabular}%
    }
\end{table}

\noindent
\textbf{Combination form of advertisement score.}
The form of $s_\text{ad}$ reflects the role of each component.
Ad sentiment is applied multiplicatively to the ad hit rate because it is only meaningful where the brand actually appears; for example, a negative brand mention should not score as high intensity despite a high hit rate (Sec.~\ref{ssec:metrics}).
Ad relevance, in contrast, measures how well the ad copy's message is conveyed, which can occur without explicitly naming the brand.
We compare the correlation with human-perceived ad intensity of our $s_\text{ad}$ against equal, multiplicative, and learned weights (linear regression on human-perceived intensity, 5-fold cross-validated) in Table~\ref{tab:suppl_sad_form}.
Both equal and multiplicative weights underperform our form.
Learned weights are marginally best as expected, but the margin over our form is negligible (0.007 in $r$, 0.002 in $\rho$), while they require costly human annotations.
Our form attains comparable correlation with human perception without any human supervision.

\subsection{Quality Control}
We employ several quality control mechanisms to ensure annotation reliability.
We employ strict qualification criteria (approval rate $\geq$ 98\%, $\geq$ 1,000 approved HITs).
Each HIT also includes an attention check that requires annotators to type a specific word, and submissions failing this check are automatically rejected.
We manually review the text justifications provided by annotators and reject submissions containing incoherent or nonsensical text indicative of automated or careless responses.
Rejected HITs are re-posted for re-annotation by other qualified workers.

\begin{table}[h]
    \centering
    \caption{Summary of annotation statistics for each evaluation task.}
    \label{tab:annotation_stats}
    \resizebox{0.8\linewidth}{!}{%
    \begin{tabular}{lcccc}
        \toprule
        Task & \# HITs & Payment / HIT & Est. time / HIT & Hourly rate \\
        \midrule
        Metric validation \& ad labeling & 913 & \$2.00 & 8--10 min & $\sim$\$12 \\
        Preference test & 1,220 & \$0.50 & 2--3 min & $\sim$\$12 \\
        \bottomrule
    \end{tabular}%
    }
\end{table}

\subsection{Annotation Statistics and Compensation}
Table~\ref{tab:annotation_stats} summarizes the annotation statistics.
The metric validation and ad labeling task comprises 913 HITs, each compensated at \$2.00 with an estimated completion time of 8--10 minutes.
The preference test comprises approximately 1,220 HITs across four baselines, each compensated at \$0.50 with an estimated completion time of 2--3 minutes.
Both tasks are designed to maintain an effective hourly rate of approximately \$12, in compliance with ethical compensation standards.

\clearpage

\section{Additional Results}
\label{sec:suppl_additional_results}
\subsection{Comparison across LLMs}
\label{ssec:suppl_other_llms}

\begin{figure}[H]
    \centering
    \includegraphics[width=0.6\textwidth]{figs/ad_resp_legend.pdf}
    \vspace{0pt}

    \begin{subfigure}[b]{0.32\textwidth}
        \centering
        \includegraphics[width=\textwidth]{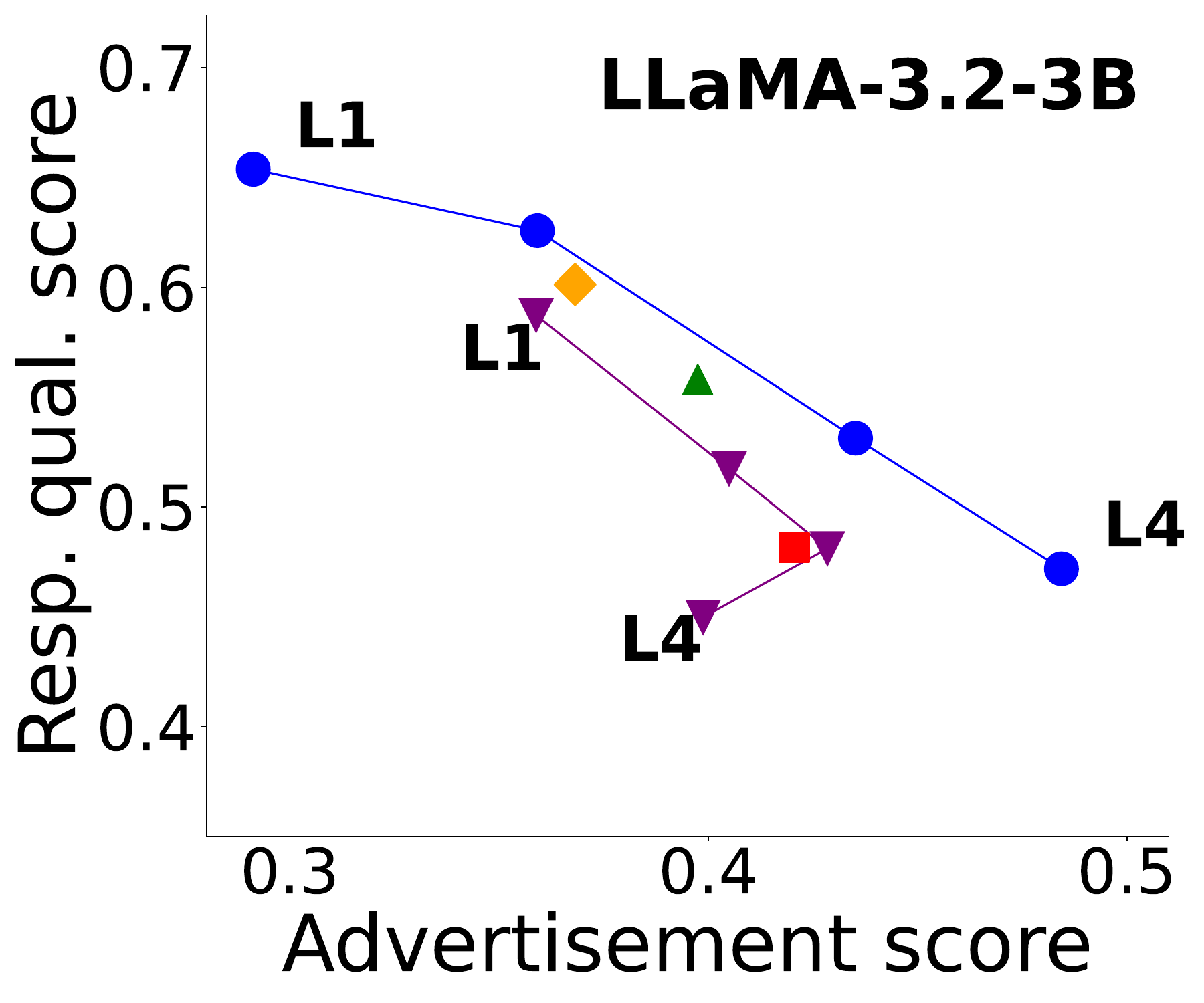}
        \label{fig:intensity_control_llama3b}
    \end{subfigure}
    \hspace{20pt}
    \begin{subfigure}[b]{0.32\textwidth}
        \centering
        \includegraphics[width=\textwidth]{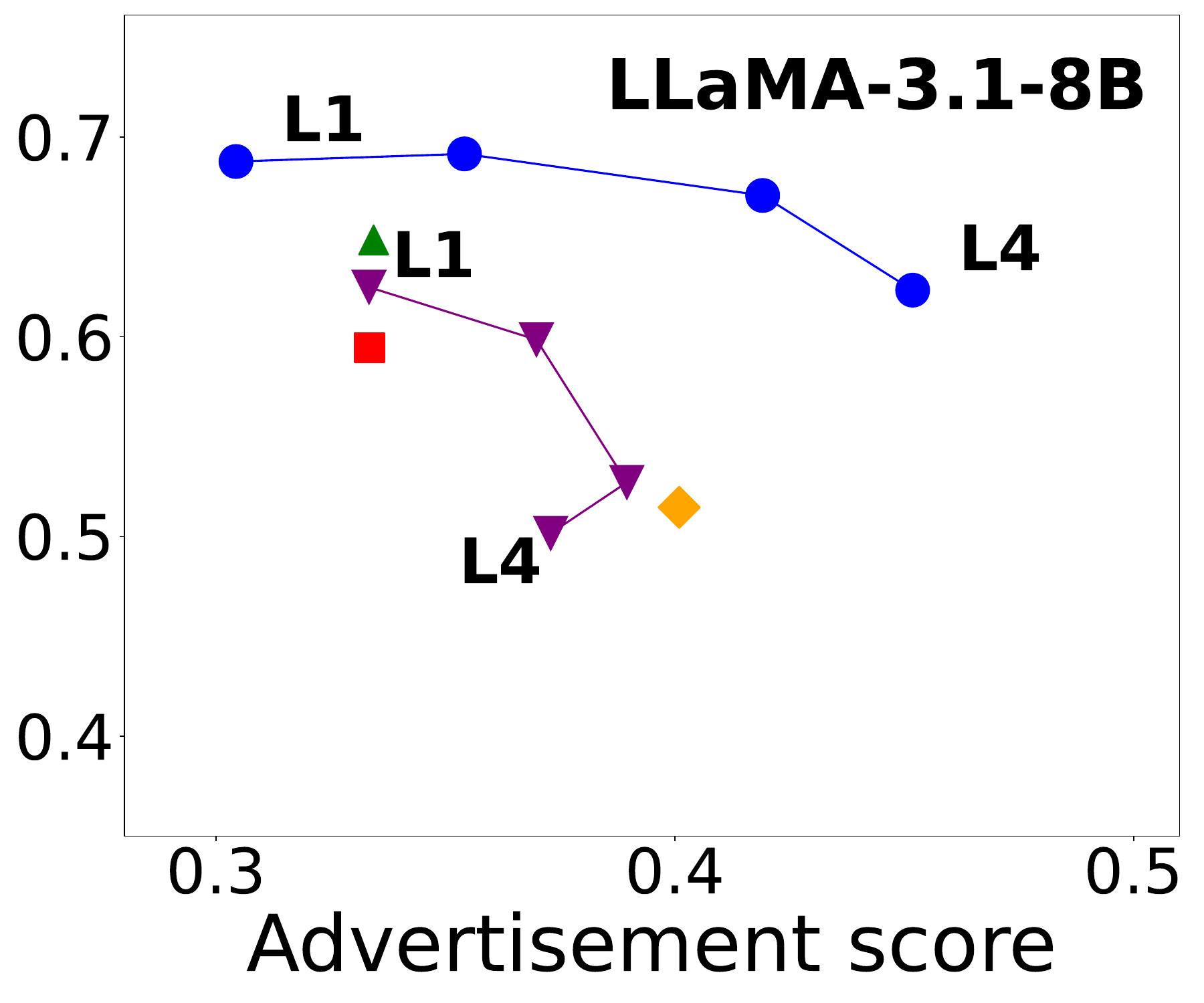}
        \label{fig:intensity_control_llama8b}
    \end{subfigure}

    \caption{
    Ad intensity results on LLaMA-3.2-3B and LLaMA-3.1-8B.
    Our \sys (L1--L4) achieves a wide and uniform range of ad intensities, while baselines are constrained to a single or a narrow intensity range.
    }
    \label{fig:suppl_intensity_control}
\end{figure}

\noindent
\textbf{Ad intensity and response quality on LLaMA.}
In Fig.~\ref{fig:suppl_intensity_control}, we report ad intensity results on LLaMA-3.2-3B and LLaMA-3.1-8B.
Consistent with the findings in Sec.~\ref{sec:experiments}, \sys achieves a wide and uniform range of ad intensities encompassing all baselines, while maintaining higher response quality at comparable intensities.
These results confirm that \sys generalizes well across different model families and scales.

\begin{figure}[H]
    \centering
    \includegraphics[width=0.4\linewidth]{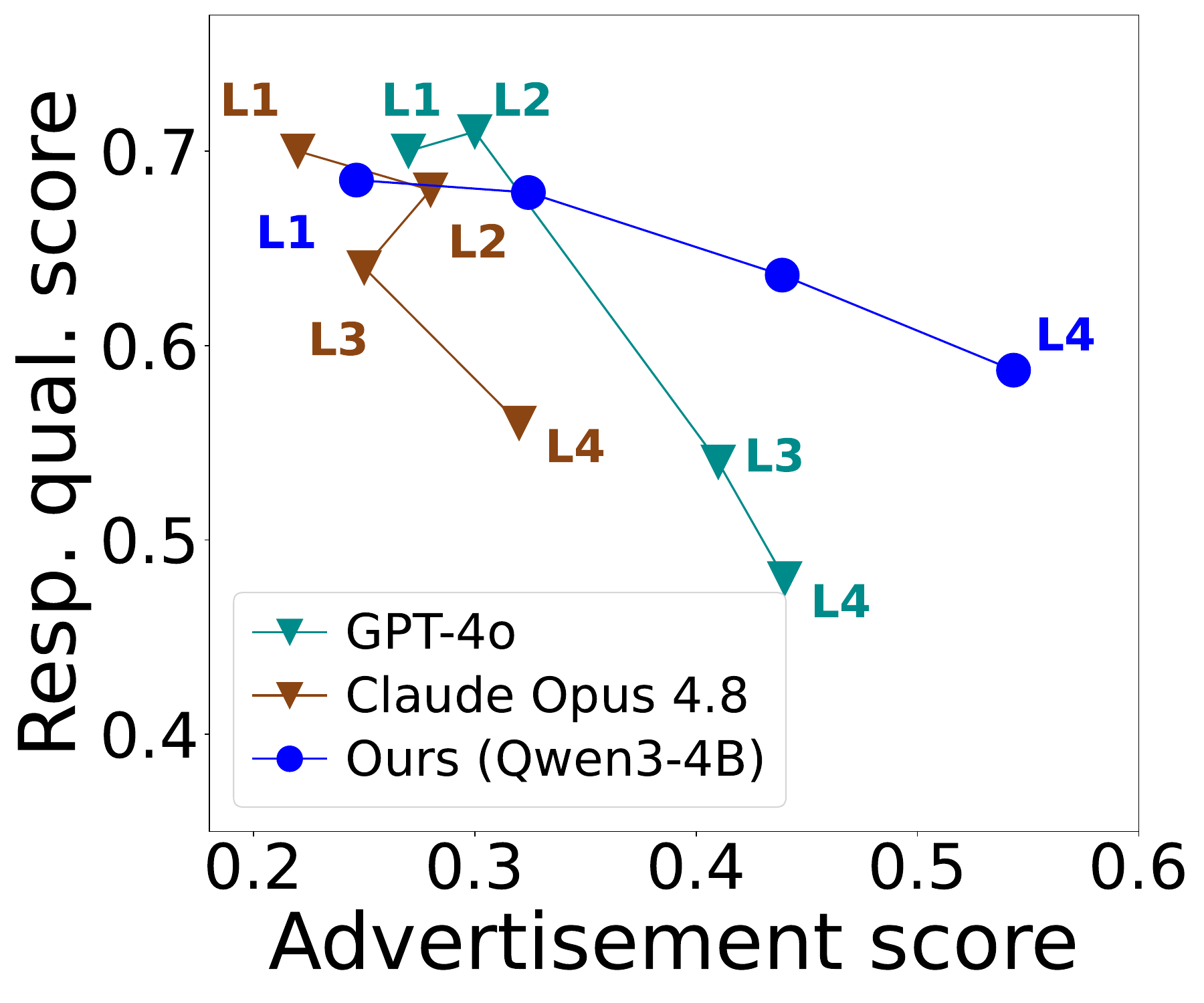}
    \caption{
    Comparison with closed-source LLMs.
    \textit{NL-prompt} is applied to GPT-4o and Claude Opus 4.8, and compared with \sys on Qwen3-4B.
    }
    \label{fig:suppl_closed_source}
\end{figure}

\noindent
\textbf{Comparison with closed-source LLMs.}
Our framework assumes a service provider that operates and monetizes its own LLM assistant (Sec.~\ref{ssec:formulation}), which permits fine-tuning and access to internal attention weights.
To examine whether prompting a top-tier closed-source model can substitute for our framework, we apply \textit{NL-prompt} to GPT-4o and Claude Opus 4.8 and compare them with \sys on Qwen3-4B (Fig.~\ref{fig:suppl_closed_source}).
Neither model achieves sufficient intensity control: GPT-4o collapses into effectively two levels (L1$\approx$L2 and L3$\approx$L4), and Claude Opus 4.8 is non-monotonic, with ad intensity decreasing from L2 to L3.
In contrast, \sys spans roughly twice the intensity range with uniform spacing, while attaining comparable or higher response quality at every level.
These results demonstrate that prompting a top-tier closed-source model is not a substitute for our framework.

\subsection{Generalization}
\label{ssec:suppl_generalization}

\begin{figure}[H]
    \centering
    \includegraphics[width=0.6\textwidth]{figs/ad_resp_legend.pdf}
    \vspace{0pt}

    \includegraphics[width=0.45\linewidth]{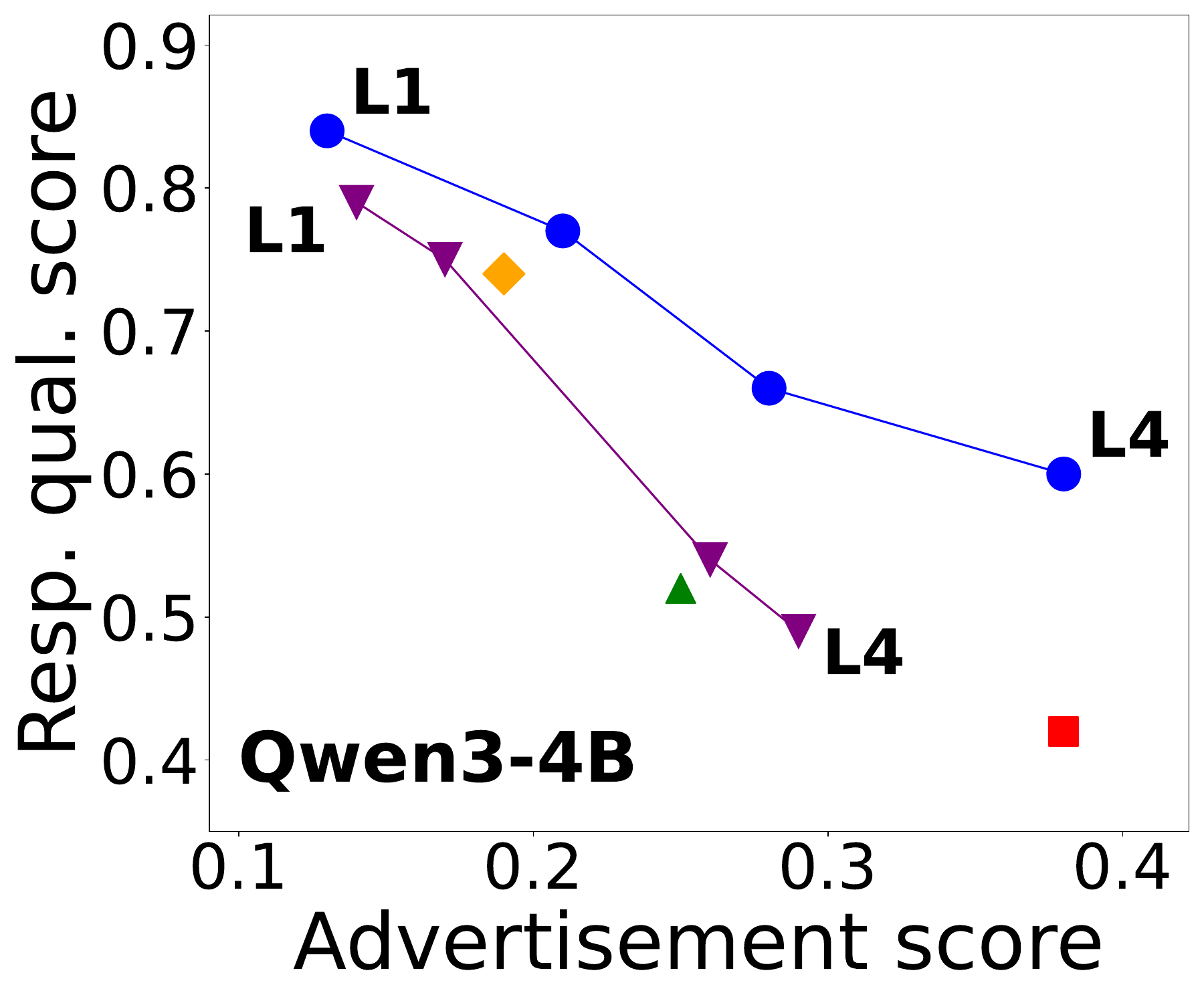}
    \caption{
    Zero-shot ad intensity results of \sys (Qwen3-4B) on real, non-synthetic inputs, with user requests from WildChat-1M~\cite{wildchat} and ad copies from the Programmatic Ad Creatives dataset~\cite{programmatic_ads}.
    }
    \label{fig:suppl_real_ads}
\end{figure}

\begin{table}[H]
    \centering
    \caption{
    Zero-shot ad labeling performance of \sys (Qwen3-4B) on real, non-synthetic inputs.
    }
    \label{tab:suppl_real_ads}
    \resizebox{0.45\linewidth}{!}{%
    \begin{tabular}{lcccc}
        \toprule
        Method & AUC & Prec. & Rec. & F1 \\
        \midrule
        Marker & 0.67 & 0.57 & 0.47 & 0.52 \\
        BERTScore & 0.63 & 0.49 & 0.69 & 0.57 \\
        AttnLRP & 0.48 & 0.46 & 0.89 & 0.62 \\
        Ours w/o $\mathcal{L}_\text{attr}$ & 0.81 & 0.50 & 0.79 & 0.61 \\
        Ours & \textbf{0.83} & \textbf{0.61} & \textbf{0.90} & \textbf{0.73} \\
        \bottomrule
    \end{tabular}%
    }
\end{table}

\noindent
\textbf{Generalization to real advertisement data.}
To examine whether \sys trained on \sysd transfers to naturally occurring advertising, we evaluate it zero-shot on inputs drawn entirely from non-synthetic sources: user requests from WildChat-1M~\cite{wildchat}, which contains real human--ChatGPT conversations, and ad copies from the Programmatic Ad Creatives dataset~\cite{programmatic_ads}, which contains ad statements from real online ad creatives (Fig.~\ref{fig:suppl_real_ads} and Table~\ref{tab:suppl_real_ads}).
Ad intensity remains monotonically controllable from L1 to L4, while response quality stays higher than all baselines at comparable ad intensity.
Ad labeling also outperforms all baselines, reaching 0.83 AUC and 0.73 F1.
This indicates that \sysd provides a training signal that transfers to real inputs without adaptation, consistent with GPT-4o's role of reformatting the real brands and queries of the Webis corpus instead of creating new entities.

\begin{figure}[H]
    \centering
    \includegraphics[width=0.45\linewidth]{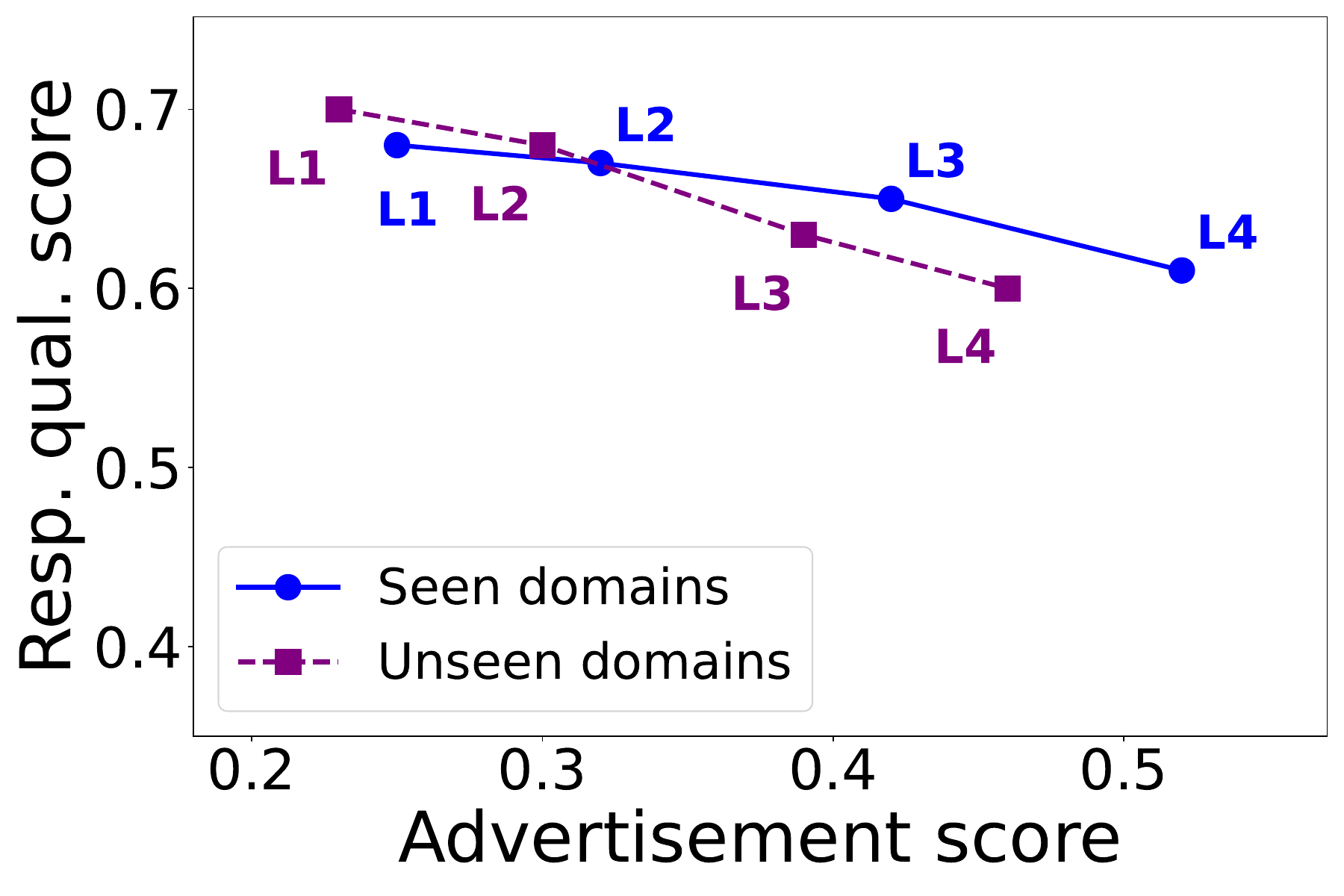}
    \caption{
    Generalization to unseen domains on Qwen3-4B trained on 8 of the 10 meta-topics (Healthcare and Gaming held out).
    \textit{Seen} and \textit{Unseen} denote test instances from the training and held-out meta-topics, respectively.
    }
    \label{fig:suppl_unseen_domain}
\end{figure}

\noindent
\textbf{Generalization to unseen domains.}
To evaluate whether \sys generalizes beyond the meta-topics seen during training, we train Qwen3-4B on 8 of the 10 meta-topics in \sysd, holding out Healthcare and Gaming, and evaluate on both seen and unseen meta-topics (Fig.~\ref{fig:suppl_unseen_domain}).
On unseen domains, the advertisement score increases monotonically from L1 to L4, and response quality remains comparable to seen domains at every level.
This indicates that our ad control tokens encode a topic-agnostic notion of advertising intensity rather than memorizing domain-specific patterns.

\begin{figure}[H]
    \centering
    \begin{subfigure}[b]{0.42\linewidth}
        \centering
        \includegraphics[width=\linewidth]{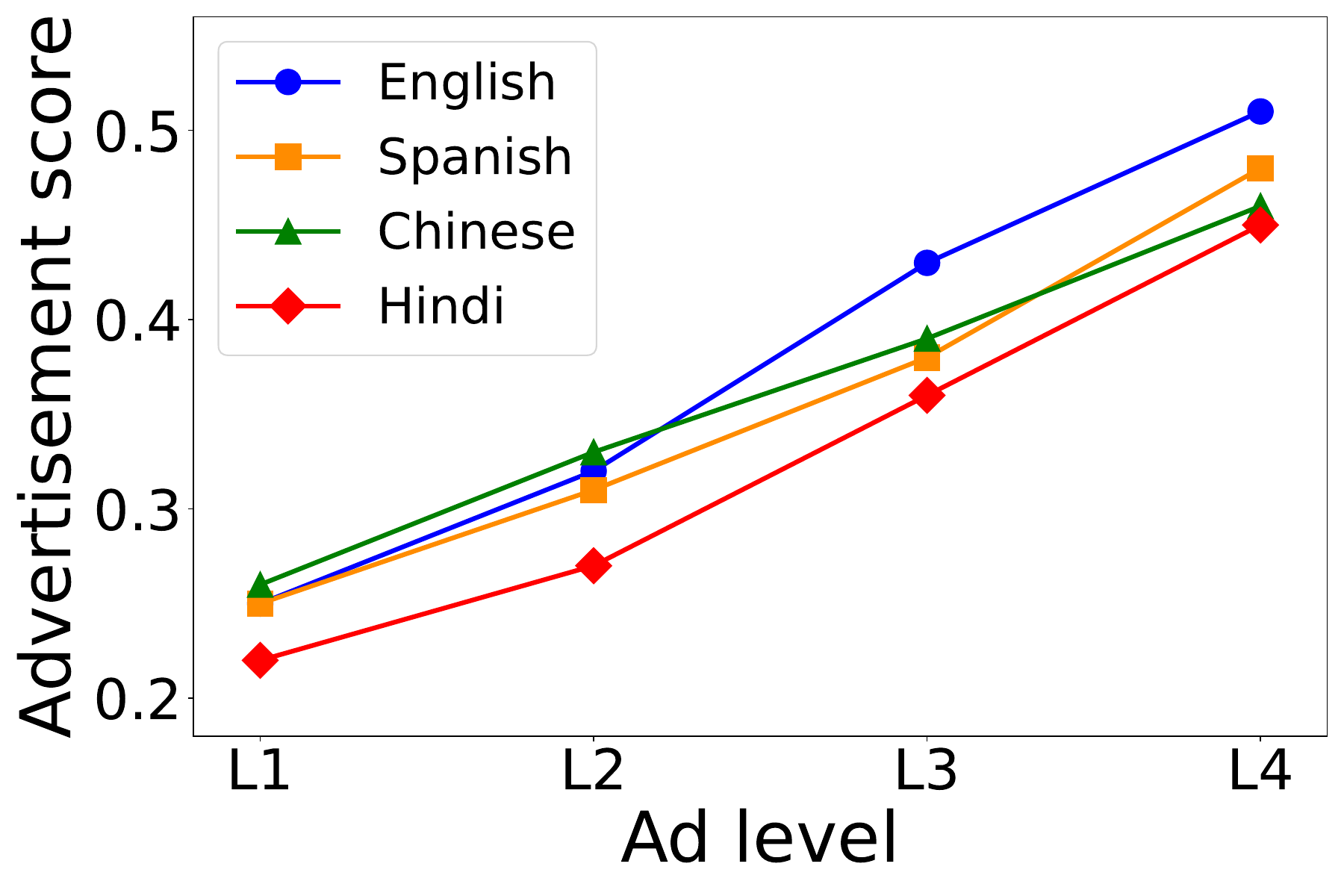}
        \caption{Advertisement score}
        \label{fig:suppl_multilingual_ad}
    \end{subfigure}
    \hspace{15pt}
    \begin{subfigure}[b]{0.42\linewidth}
        \centering
        \includegraphics[width=\linewidth]{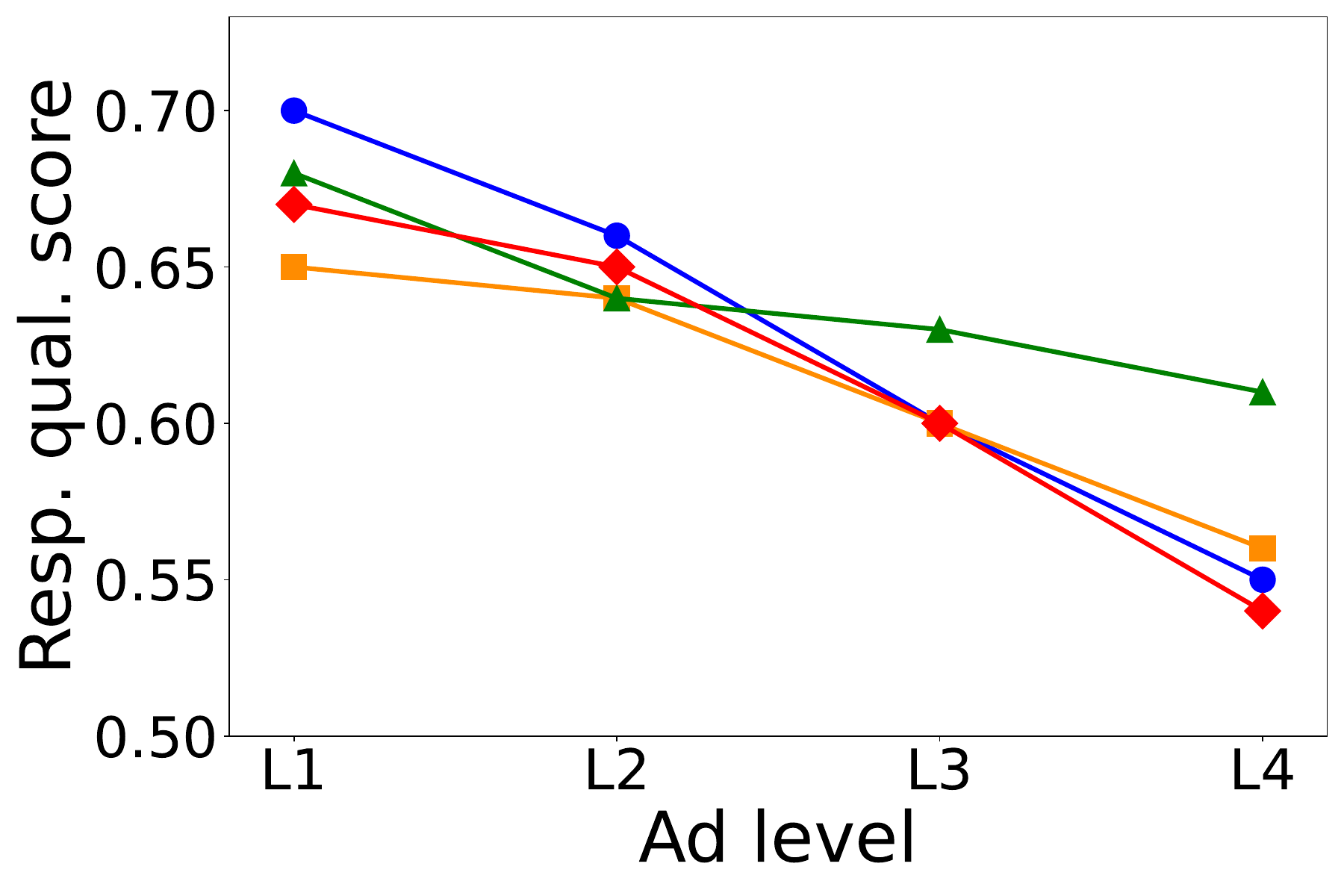}
        \caption{Response quality score}
        \label{fig:suppl_multilingual_qual}
    \end{subfigure}
    \caption{
    Zero-shot multilingual results of \sys (Qwen3-4B) trained on English \sysd.
    User requests and ad copies are translated into Spanish, Chinese, and Hindi.
    Scores are measured with multilingual evaluators.
    }
    \label{fig:suppl_multilingual}
\end{figure}

\begin{table}[H]
    \centering
    \caption{
    Zero-shot multilingual ad labeling performance of \sys (Qwen3-4B) trained on English \sysd.
    }
    \label{tab:suppl_multilingual}
    \resizebox{0.45\linewidth}{!}{%
    \begin{tabular}{lcccc}
        \toprule
        Language & AUC & Prec. & Rec. & F1 \\
        \midrule
        English & 0.77 & 0.67 & 0.89 & 0.76 \\
        Spanish & 0.73 & 0.65 & 0.90 & 0.75 \\
        Chinese & 0.73 & 0.61 & 0.84 & 0.71 \\
        Hindi   & 0.81 & 0.64 & 0.87 & 0.74 \\
        \bottomrule
    \end{tabular}%
    }
\end{table}

\noindent
\textbf{Generalization to multilingual settings.}
Our framework is designed to be language-agnostic, as ad control tokens carry no pre-trained meaning and the ad-attribution loss operates independently of surface lexicon.
To empirically support this, we evaluate \sys trained on English \sysd zero-shot in multilingual settings.
We translated the user requests and ad copies from \sysd into Spanish, Chinese, and Hindi, spanning distinct scripts (Latin, Chinese, Devanagari) and language families, using GPT-4o and evaluated our \sys (Qwen3-4B) zero-shot on these languages.
To ensure the advertisement and response quality scores are measured reliably, we used models with multilingual support; we replaced English DistilBERT with its multilingual variant for ad sentiment, LLaMA-3-8B with multilingual MPNet encoder for ad relevance, and LLaMA-3-8B-Instruct with LLaMA-3.1-8B-Instruct for response quality judgment.
The results (Fig.~\ref{fig:suppl_multilingual} and Table~\ref{tab:suppl_multilingual}) show that \sys's ad intensity control and labeling transfer zero-shot to unseen languages.
Advertisement scores increase monotonically from L1 to L4, while the response quality scores and ad labeling remain close to the English setting in every language.
These results support our claim that the ad control tokens and attention-based attribution are language-agnostic.
For cultural adaptation, we expect that training on a multilingual, multi-level preference dataset with ad-inserted responses written natively in each language would better reflect each culture's advertising conventions, which we view as a promising direction for future work.

\begin{table}[H]
    \centering
    \caption{
    Advertisement scores of \sys (Qwen3-4B) in 3-turn conversations under four per-turn ad level setups.
    }
    \label{tab:suppl_multi_turn}
    \resizebox{0.5\linewidth}{!}{%
    \begin{tabular}{l|ccc}
        \toprule
        Ad level for each turn & Turn 1 & Turn 2 & Turn 3 \\
        \midrule
        L1-L1-L1 & 0.247 & 0.267 & 0.260 \\
        L4-L4-L4 & 0.543 & 0.490 & 0.491 \\
        L1-L1-L4 & 0.247 & 0.267 & 0.498 \\
        L4-L1-L1 & 0.543 & 0.254 & 0.233 \\
        \bottomrule
    \end{tabular}%
    }
\end{table}

\noindent
\textbf{Multi-turn conversation.}
Our control tokens are appended per turn to the user prompt, so each turn's ad intensity is determined by the token supplied for that turn rather than carried over implicitly as a context.
We evaluate this on a 3-turn conversation where the user asks the original user request followed by ``Can you tell me more about this?'' and ``What else should I consider?''.
Past turns enter the context as plain user/assistant messages.
Across four setups in Table~\ref{tab:suppl_multi_turn}, \sys maintains the appropriate level of ad intensity for each turn.
We note, however, that repeatedly promoting the same brand throughout a conversation may not be desirable in deployment, and a more natural behavior would show an ad for a brand once and refrain thereafter.
When to place an ad within a conversation is a separate ad-allocation problem, which we leave as future work.

\begin{figure}[H]
    \centering
    \includegraphics[width=0.45\linewidth]{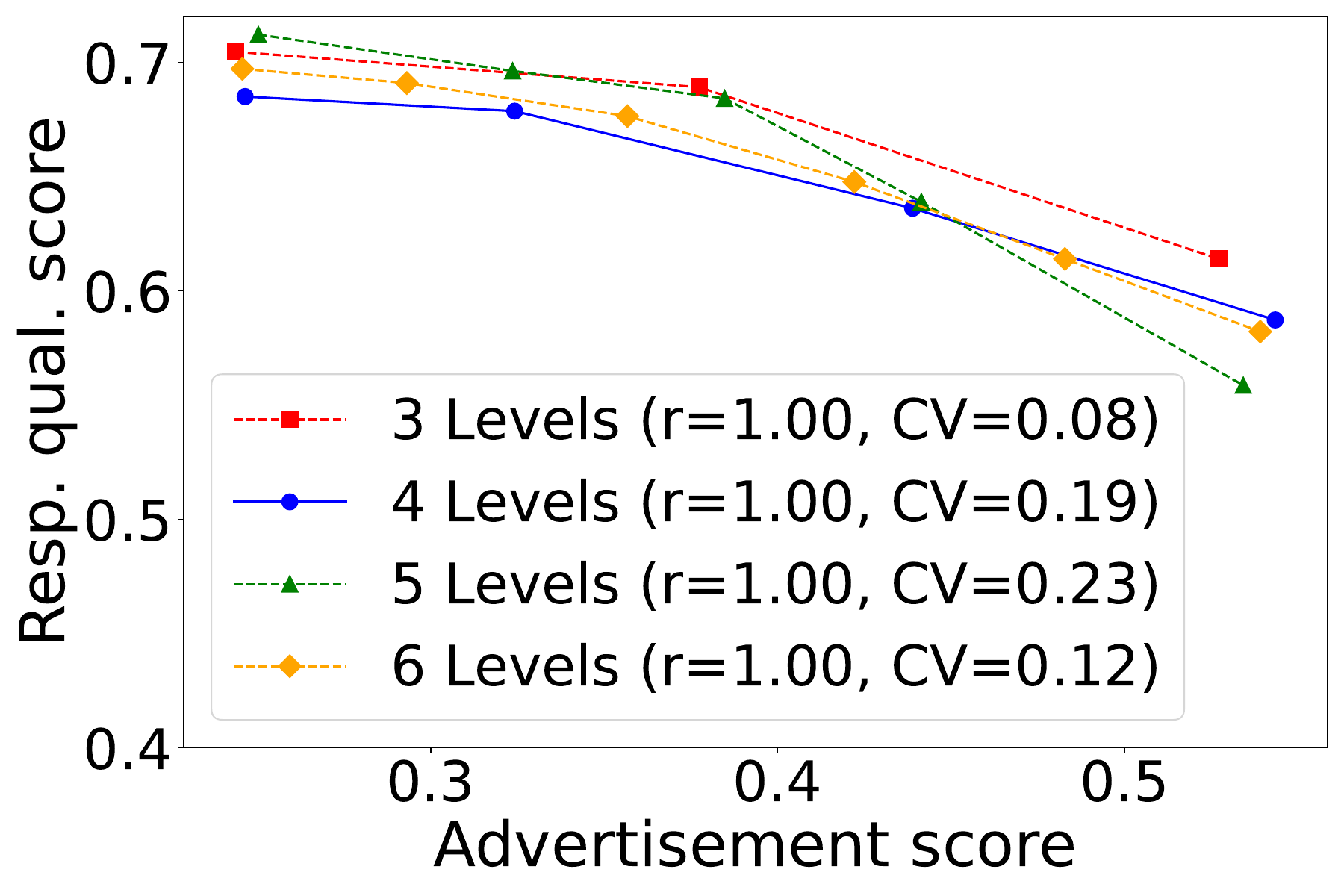}
    \caption{
    Generalization across different numbers of intensity levels. \sys maintains uniform intensity control and response quality across different numbers of levels.
    }
    \label{fig:diff_levels}
\end{figure}

\noindent
\textbf{Generalization across number of levels.}
In Fig.~\ref{fig:diff_levels}, we examine the generalizability of \sys across different numbers of ad intensity levels.
\sys maintains uniformity and consistent response quality across $3$ to $6$ levels, demonstrating that the ad control tokens generalize well to varying numbers of levels.

\subsection{Analysis of Ad Control Tokens}
\label{ssec:suppl_control_token}

\begin{figure}[H]
    \centering
    \includegraphics[width=0.45\linewidth]{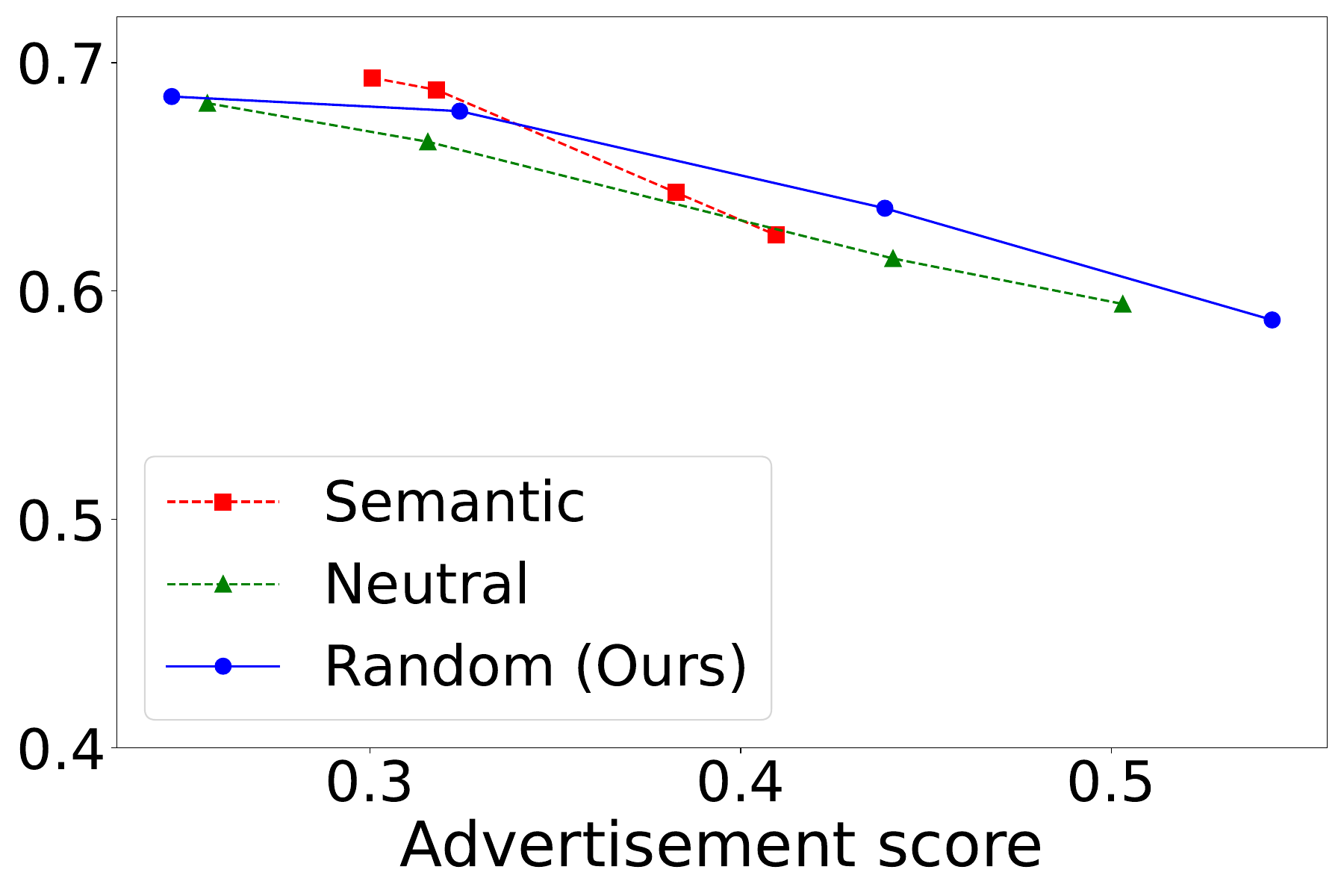}
    \caption{
    Ad control token embedding initialization. Our random initialization achieves the best balance of intensity control and response quality.
    }
    \label{fig:embedding_init}
\end{figure}

\noindent
\textbf{Control token embedding initialization.}
In Fig.~\ref{fig:embedding_init}, we compare different initialization strategies for ad control token embeddings.
\textit{Semantic} initializes each token to the embedding of numeric tokens (``1'', ``2'', ``3'', ``4''), and \textit{Neutral} initializes to punctuation tokens (``.'', ``,'', ``!'', ``?'').
\textit{Semantic} yields the narrowest intensity range, as pre-trained numeric semantics (\eg, arithmetic, ordinal) resist repurposing for intensity control, constraining the model's ability to separate intensity levels.
\textit{Neutral} avoids numeric semantics but still inherits positional and syntactic biases from the pretrained embeddings, limiting the intensity range.
Our random initialization carries no pre-existing bias, allowing the model to freely learn the mapping from each token to its target intensity, achieving the widest and most uniform intensity range.

\begin{figure}[H]
    \centering
    \includegraphics[width=0.5\linewidth]{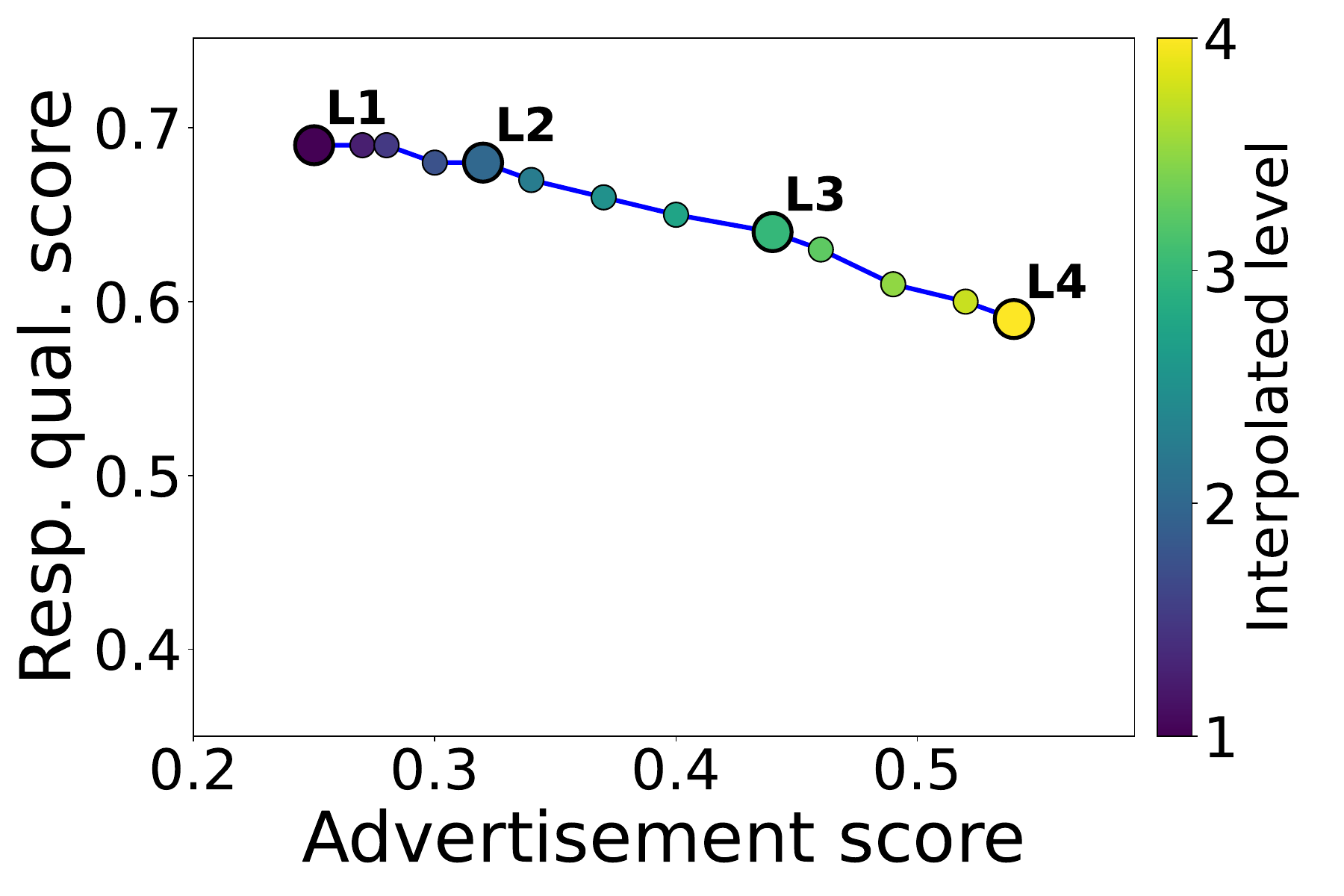}
    \caption{
    Continuous ad intensity control on Qwen3-4B trained on 4 discrete levels, obtained by linearly interpolating between the embeddings of adjacent ad control tokens in intervals of 0.25.
    Larger markers denote the discrete levels (L1--L4).
    }
    \label{fig:suppl_continuous_control}
\end{figure}

\noindent
\textbf{Continuous intensity control.}
The discrete-level design reflects application needs rather than a technical limit, as deployment requires ad intensities that can be named or documented (\eg, L1 for paid, L3 for free tiers).
Yet, continuous control is feasible without any architectural change.
We linearly interpolate between the embeddings of adjacent ad control tokens on a model trained on 4 discrete levels in intervals of 0.25 (Fig.~\ref{fig:suppl_continuous_control}).
Ad intensity increases smoothly and monotonically across all settings, verifying near-continuous ad intensity control without retraining.

\subsection{Analysis of Ad Labeling}
\label{ssec:suppl_ad_labeling}

\begin{table}[H]
    \centering
    \caption{
    Faithfulness of ad attribution on Qwen3-4B.
    We report $s_\text{ad}$ before and after masking the response tokens with the highest attribution scores.
    A larger drop indicates that the attribution better identifies ad-related tokens.
    }
    \label{tab:suppl_faithfulness}
    \resizebox{0.65\linewidth}{!}{%
    \begin{tabular}{lccc}
        \toprule
        Method & w/o masking & w/ masking & $\Delta s_\text{ad}$ \\
        \midrule
        $\mathcal{L}_\text{dpo}$ only & 0.399 & 0.318 & $-0.081$ \\
        $\mathcal{L}_\text{dpo} + \mathcal{L}_\text{attr}$ (Ours) & 0.388 & 0.213 & $\mathbf{-0.175}$ \\
        \bottomrule
    \end{tabular}%
    }
\end{table}

\noindent
\textbf{Faithfulness of ad attribution.}
Since attention-based attribution is an approximation of token importance, we verify whether our supervised attention rollout faithfully identifies ad-related tokens.
We mask the response tokens with the highest attribution scores and measure the resulting drop in the advertisement score $s_\text{ad}$ (Table~\ref{tab:suppl_faithfulness}).
Without $\mathcal{L}_\text{attr}$, masking reduces $s_\text{ad}$ only from 0.399 to 0.318 ($-0.081$), whereas with $\mathcal{L}_\text{attr}$, it decreases from 0.388 to 0.213 ($-0.175$), more than twice the drop.
This confirms that $\mathcal{L}_\text{attr}$ guides attention rollout to concentrate on tokens that actually carry the advertising content, making it a reliable attribution signal for ad labeling.

\begin{table}[H]
    \centering
    \caption{
    Ad labeling performance across \textbf{pseudo-label thresholds} $\delta^*$ at inference threshold $\tau=0.005$.
    Best results in \textbf{bold}.
    }
    \label{tab:suppl_ad_labeling_delta}
    \resizebox{\textwidth}{!}{%
    \begin{tabular}{llcccccccccccccccc}
        \toprule
        \multirow{2}{*}{Model} & \multirow{2}{*}{Method} & \multicolumn{4}{c}{$\delta^*=0.5$} & \multicolumn{4}{c}{$\delta^*=1.0$} & \multicolumn{4}{c}{$\delta^*=1.5$} & \multicolumn{4}{c}{$\delta^*=2.0$} \\
        \cmidrule(lr){3-6} \cmidrule(lr){7-10} \cmidrule(lr){11-14} \cmidrule(lr){15-18}
        & & AUC & Prec. & Rec. & F1 & AUC & Prec. & Rec. & F1 & AUC & Prec. & Rec. & F1 & AUC & Prec. & Rec. & F1 \\
        \midrule
        \multirow{5}{*}{Qwen3-4B}
        & Marker                        & 0.59 & 0.73 & 0.36 & 0.48 & 0.58 & 0.62 & 0.37 & 0.47 & 0.57 & 0.54 & 0.37 & 0.44 & 0.56 & 0.47 & 0.36 & 0.41 \\
        & BERTScore                     & 0.66 & 0.72 & 0.71 & 0.71 & 0.68 & 0.64 & 0.73 & 0.68 & 0.69 & 0.59 & 0.74 & 0.65 & 0.70 & 0.55 & 0.75 & 0.63 \\
        & AttnLRP                       & 0.50 & 0.67 & 0.82 & 0.74 & 0.48 & 0.57 & 0.81 & 0.67 & 0.49 & 0.52 & 0.81 & 0.64 & 0.48 & 0.48 & 0.81 & 0.60 \\
        & $\mathcal{L}_\text{dpo}$ only & 0.69 & 0.70 & 0.58 & 0.64 & 0.69 & 0.64 & 0.61 & 0.62 & 0.69 & 0.58 & 0.61 & 0.60 & 0.69 & 0.54 & 0.62 & 0.58 \\
        & Ours                          & \textbf{0.75} & \textbf{0.76} & \textbf{0.87} & \textbf{0.81} & \textbf{0.77} & \textbf{0.67} & \textbf{0.89} & \textbf{0.76} & \textbf{0.77} & \textbf{0.60} & \textbf{0.90} & \textbf{0.72} & \textbf{0.77} & \textbf{0.56} & \textbf{0.91} & \textbf{0.69} \\
        \midrule
        \multirow{5}{*}{Qwen2.5-14B}
        & Marker                        & 0.56 & 0.68 & 0.21 & 0.33 & 0.59 & 0.52 & 0.27 & 0.36 & 0.58 & 0.36 & 0.28 & 0.32 & 0.58 & 0.27 & 0.28 & 0.28 \\
        & BERTScore                     & 0.62 & 0.60 & 0.48 & 0.53 & 0.65 & 0.48 & 0.53 & 0.51 & 0.68 & 0.41 & 0.60 & 0.49 & 0.67 & 0.34 & 0.61 & 0.44 \\
        & AttnLRP                       & 0.49 & 0.52 & 0.71 & 0.60 & 0.47 & 0.35 & 0.67 & 0.46 & 0.49 & 0.26 & 0.67 & 0.38 & 0.49 & 0.22 & 0.67 & 0.33 \\
        & $\mathcal{L}_\text{dpo}$ only & 0.74 & 0.69 & 0.61 & 0.65 & 0.74 & 0.49 & 0.67 & 0.56 & 0.76 & 0.35 & 0.75 & 0.48 & 0.77 & 0.28 & 0.78 & 0.41 \\
        & Ours                          & \textbf{0.78} & \textbf{0.70} & \textbf{0.85} & \textbf{0.76} & \textbf{0.79} & \textbf{0.66} & \textbf{0.88} & \textbf{0.76} & \textbf{0.80} & \textbf{0.52} & \textbf{0.96} & \textbf{0.68} & \textbf{0.81} & \textbf{0.45} & \textbf{0.97} & \textbf{0.62 }\\
        \midrule
        \multirow{5}{*}{Phi-4-Mini (3.8B)}
        & Marker                        & 0.52 & 0.61 & 0.13 & 0.22 & 0.53 & 0.47 & 0.14 & 0.22 & 0.52 & 0.34 & 0.14 & 0.20 & 0.52 & 0.25 & 0.14 & 0.18 \\
        & BERTScore                     & 0.67 & 0.63 & 0.73 & 0.67 & 0.70 & 0.56 & 0.78 & 0.65 & 0.71 & 0.46 & 0.80 & 0.59 & 0.72 & 0.40 & 0.82 & 0.54 \\
        & AttnLRP                       & 0.51 & 0.46 & 0.84 & 0.60 & 0.53 & 0.30 & 0.83 & 0.44 & 0.54 & 0.21 & 0.82 & 0.34 & 0.55 & 0.16 & 0.82 & 0.26 \\
        & $\mathcal{L}_\text{dpo}$ only & 0.70 & 0.61 & 0.62 & 0.61 & 0.75 & 0.46 & 0.71 & 0.56 & 0.77 & 0.36 & 0.77 & 0.49 & 0.79 & 0.28 & 0.80 & 0.41 \\
        & Ours                          & \textbf{0.72} & \textbf{0.78} & \textbf{0.90} & \textbf{0.84} & \textbf{0.76} & \textbf{0.62} & \textbf{0.93} & \textbf{0.75} & \textbf{0.79} & \textbf{0.52} & \textbf{0.95} & \textbf{0.67} & \textbf{0.81} & \textbf{0.44} & \textbf{0.96} & \textbf{0.61} \\
        \midrule
        \multirow{5}{*}{Phi-4 (14B)}
        & Marker                        & 0.60 & 0.56 & 0.23 & 0.32 & 0.62 & 0.56 & 0.28 & 0.37 & 0.64 & \textbf{0.54} & 0.33 & 0.41 & 0.66 & \textbf{0.51} & 0.37 & 0.43 \\
        & BERTScore                     & 0.69 & 0.57 & 0.59 & 0.58 & 0.72 & 0.43 & 0.65 & 0.51 & 0.75 & 0.35 & 0.70 & 0.47 & 0.78 & 0.30 & 0.74 & 0.42 \\
        & AttnLRP                       & 0.54 & 0.35 & 0.65 & 0.45 & 0.57 & 0.24 & 0.66 & 0.35 & 0.60 & 0.19 & 0.67 & 0.29 & 0.62 & 0.15 & 0.67 & 0.24 \\
        & $\mathcal{L}_\text{dpo}$ only & 0.66 & 0.65 & 0.54 & 0.59 & 0.71 & 0.53 & 0.62 & 0.57 & 0.74 & 0.46 & 0.67 & 0.54 & 0.76 & 0.38 & 0.73 & 0.50 \\
        & Ours                          & \textbf{0.73} & \textbf{0.66} & \textbf{0.66} & \textbf{0.66} & \textbf{0.77} & \textbf{0.58} & \textbf{0.67} & \textbf{0.62} & \textbf{0.79} & 0.47 & \textbf{0.73} & \textbf{0.57} & \textbf{0.80} & 0.39 & \textbf{0.78} & \textbf{0.51} \\
        \midrule
        \multirow{5}{*}{LLaMA-3.2-3B}
        & Marker                        & 0.53 & 0.70 & 0.19 & 0.30 & 0.53 & 0.55 & 0.20 & 0.30 & 0.53 & 0.42 & 0.21 & 0.28 & 0.53 & 0.34 & 0.21 & 0.26 \\
        & BERTScore                     & 0.70 & 0.74 & 0.58 & 0.65 & 0.71 & 0.60 & 0.63 & 0.62 & 0.71 & 0.51 & 0.66 & 0.57 & 0.72 & 0.42 & 0.67 & 0.52 \\
        & AttnLRP                       & 0.49 & 0.64 & 0.59 & 0.61 & 0.51 & 0.48 & 0.59 & 0.53 & 0.53 & 0.39 & 0.59 & 0.47 & 0.55 & 0.32 & 0.61 & 0.42 \\
        & $\mathcal{L}_\text{dpo}$ only & 0.72 & 0.77 & 0.28 & 0.41 & 0.74 & \textbf{0.70} & 0.34 & 0.46 & 0.75 & \textbf{0.60} & 0.37 & 0.46 & 0.74 & 0.46 & 0.38 & 0.42 \\
        & Ours                          & \textbf{0.74} & \textbf{0.80} & \textbf{0.67} & \textbf{0.73} & \textbf{0.77} & 0.68 & \textbf{0.74} & \textbf{0.71} & \textbf{0.80} & 0.57 & \textbf{0.79} & \textbf{0.66} & \textbf{0.81} & \textbf{0.48} & \textbf{0.82} & \textbf{0.61} \\
        \midrule
        \multirow{5}{*}{LLaMA-3.1-8B}
        & Marker                        & 0.48 & 0.51 & 0.38 & 0.44 & 0.52 & 0.41 & 0.42 & 0.42 & 0.55 & 0.36 & 0.46 & 0.41 & 0.57 & 0.32 & 0.48 & 0.39 \\
        & BERTScore                     & 0.71 & 0.73 & 0.54 & 0.62 & 0.76 & 0.62 & 0.65 & 0.64 & 0.80 & 0.55 & 0.72 & 0.63 & 0.82 & 0.50 & 0.76 & 0.60 \\
        & AttnLRP                       & 0.48 & 0.51 & 0.39 & 0.44 & 0.52 & 0.41 & 0.42 & 0.42 & 0.55 & 0.36 & 0.46 & 0.41 & 0.57 & 0.33 & 0.48 & 0.39 \\
        & $\mathcal{L}_\text{dpo}$ only & 0.70 & 0.78 & 0.32 & 0.46 & 0.75 & 0.72 & 0.42 & 0.53 & 0.78 & 0.64 & 0.50 & 0.56 & 0.80 & 0.58 & 0.55 & 0.56 \\
        & Ours                          & \textbf{0.73} & \textbf{0.85} & \textbf{0.60} & \textbf{0.70} & \textbf{0.79} & \textbf{0.77} & \textbf{0.66} & \textbf{0.71} & \textbf{0.82} & \textbf{0.66} & \textbf{0.73} & \textbf{0.69} & \textbf{0.84} & \textbf{0.59} & \textbf{0.77} & \textbf{0.67} \\
        \bottomrule
    \end{tabular}%
    }

    \vspace{10pt}

    \centering
    \caption{
    Ad labeling performance across \textbf{inference thresholds} $\tau$ at pseudo-label threshold $\delta^*=1.0$.
    Best results in \textbf{bold}.
    }
    \label{tab:suppl_ad_labeling_tau}
    \resizebox{\textwidth}{!}{%
    \begin{tabular}{llccccccccccccc}
        \toprule
        \multirow{2}{*}{Model} & \multirow{2}{*}{Method} & \multirow{2}{*}{AUC} & \multicolumn{3}{c}{$\tau=0.003$} & \multicolumn{3}{c}{$\tau=0.004$} & \multicolumn{3}{c}{$\tau=0.005$} & \multicolumn{3}{c}{$\tau=0.006$} \\
        \cmidrule(lr){4-6} \cmidrule(lr){7-9} \cmidrule(lr){10-12} \cmidrule(lr){13-15}
        & & & Prec. & Rec. & F1 & Prec. & Rec. & F1 & Prec. & Rec. & F1 & Prec. & Rec. & F1 \\
        \midrule
        \multirow{5}{*}{Qwen3-4B}
        & Marker                        & 0.58 & 0.62 & 0.37 & 0.47 & 0.62 & 0.37 & 0.47 & 0.62 & 0.37 & 0.47 & 0.62 & 0.37 & 0.47 \\
        & BERTScore                     & 0.68 & \textbf{0.64} & 0.73 & 0.68 & \textbf{0.64} & 0.73 & 0.68 & 0.64 & 0.73 & 0.68 & 0.64 & 0.73 & 0.68 \\
        & AttnLRP                       & 0.48 & 0.57 & 0.81 & 0.67 & 0.57 & 0.81 & 0.67 & 0.57 & 0.81 & 0.67 & 0.57 & \textbf{0.81} & 0.67 \\
        & $\mathcal{L}_\text{dpo}$ only & 0.69 & 0.52 & 0.88 & 0.65 & 0.57 & 0.74 & 0.64 & 0.64 & 0.61 & 0.62 & 0.68 & 0.41 & 0.51 \\
        & Ours                          & \textbf{0.77} & 0.60 & \textbf{1.00} & \textbf{0.75} & 0.62 & \textbf{0.98} & \textbf{0.76} & \textbf{0.67} & \textbf{0.89} & \textbf{0.76} & \textbf{0.72} & 0.76 & \textbf{0.74} \\
        \midrule
        \multirow{5}{*}{Qwen2.5-14B}
        & Marker                        & 0.59 & 0.52 & 0.27 & 0.36 & 0.52 & 0.27 & 0.36 & 0.52 & 0.27 & 0.36 & 0.52 & 0.27 & 0.36 \\
        & BERTScore                     & 0.65 & 0.48 & 0.53 & 0.51 & 0.48 & 0.53 & 0.51 & 0.48 & 0.53 & 0.51 & 0.48 & 0.53 & 0.51 \\
        & AttnLRP                       & 0.47 & 0.35 & 0.67 & 0.46 & 0.35 & 0.67 & 0.46 & 0.35 & 0.67 & 0.46 & 0.35 & 0.67 & 0.46 \\
        & $\mathcal{L}_\text{dpo}$ only & 0.74 & 0.31 & \textbf{0.98} & 0.47 & 0.39 & 0.89 & 0.54 & 0.49 & 0.67 & 0.56 & 0.58 & 0.62 & 0.60 \\
        & Ours                          & \textbf{0.79} & \textbf{0.58} & \textbf{0.98} & \textbf{0.73} & \textbf{0.62} & \textbf{0.96} & \textbf{0.75} & \textbf{0.66} & \textbf{0.88} & \textbf{0.76} & \textbf{0.73} & \textbf{0.77} & \textbf{0.75} \\
        \midrule
        \multirow{5}{*}{Phi-4-Mini (3.8B)}
        & Marker                        & 0.53 & 0.47 & 0.14 & 0.22 & 0.47 & 0.14 & 0.22 & 0.47 & 0.14 & 0.22 & 0.47 & 0.14 & 0.22 \\
        & BERTScore                     & 0.70 & 0.56 & 0.78 & 0.65 & 0.56 & 0.78 & 0.65 & 0.56 & 0.78 & 0.65 & 0.56 & 0.78 & 0.65 \\
        & AttnLRP                       & 0.53 & 0.30 & 0.83 & 0.44 & 0.30 & 0.83 & 0.44 & 0.30 & 0.83 & 0.44 & 0.30 & 0.83 & 0.44 \\
        & $\mathcal{L}_\text{dpo}$ only & 0.75 & 0.33 & 0.95 & 0.49 & 0.38 & 0.85 & 0.52 & 0.46 & 0.71 & 0.56 & 0.55 & 0.57 & 0.56 \\
        & Ours                          & \textbf{0.76} & \textbf{0.59} & \textbf{1.00} & \textbf{0.74} & \textbf{0.60} & \textbf{0.98} & \textbf{0.74} & \textbf{0.62} & \textbf{0.93} & \textbf{0.75} & \textbf{0.66} & \textbf{0.86} & \textbf{0.75} \\
        \midrule
        \multirow{5}{*}{Phi-4 (14B)}
        & Marker                        & 0.62 & \textbf{0.56} & 0.28 & 0.37 & \textbf{0.56} & 0.28 & 0.37 & 0.56 & 0.28 & 0.37 & 0.56 & 0.28 & 0.37 \\
        & BERTScore                     & 0.72 & 0.43 & 0.65 & 0.51 & 0.43 & 0.65 & 0.51 & 0.43 & 0.65 & 0.51 & 0.43 & 0.65 & 0.51 \\
        & AttnLRP                       & 0.57 & 0.24 & 0.66 & 0.35 & 0.24 & 0.66 & 0.35 & 0.24 & 0.66 & 0.35 & 0.24 & \textbf{0.66} & 0.35 \\
        & $\mathcal{L}_\text{dpo}$ only & 0.71 & 0.47 & 0.69 & 0.56 & 0.50 & 0.66 & 0.57 & 0.53 & 0.62 & 0.57 & 0.56 & 0.59 & 0.57 \\
        & Ours                          & \textbf{0.77} & 0.51 & \textbf{0.72} & \textbf{0.60} & 0.54 & \textbf{0.70} & \textbf{0.61} & \textbf{0.58} & \textbf{0.67} & \textbf{0.62} & \textbf{0.60} & 0.63 & \textbf{0.62} \\
        \midrule
        \multirow{5}{*}{LLaMA-3.2-3B}
        & Marker                        & 0.53 & 0.55 & 0.20 & 0.30 & 0.55 & 0.20 & 0.30 & 0.55 & 0.20 & 0.30 & 0.55 & 0.20 & 0.30 \\
        & BERTScore                     & 0.71 & \textbf{0.60} & 0.63 & 0.62 & \textbf{0.60} & 0.63 & 0.62 & 0.60 & 0.63 & 0.62 & 0.60 & \textbf{0.63} & 0.62 \\
        & AttnLRP                       & 0.51 & 0.48 & 0.59 & 0.53 & 0.48 & 0.59 & 0.53 & 0.48 & 0.59 & 0.53 & 0.48 & 0.59 & 0.53 \\
        & $\mathcal{L}_\text{dpo}$ only & 0.74 & 0.49 & 0.85 & 0.63 & \textbf{0.60} & 0.58 & 0.59 & \textbf{0.70} & 0.34 & 0.46 & \textbf{0.77} & 0.17 & 0.28 \\
        & Ours                          & \textbf{0.77} & 0.53 & \textbf{0.98} & \textbf{0.69} & \textbf{0.60} & \textbf{0.84} & \textbf{0.70} & 0.68 & \textbf{0.74} & \textbf{0.71} & 0.75 & 0.56 & \textbf{0.64} \\
        \midrule
        \multirow{5}{*}{LLaMA-3.1-8B}
        & Marker                        & 0.52 & 0.41 & 0.42 & 0.42 & 0.41 & 0.42 & 0.42 & 0.41 & 0.42 & 0.42 & 0.41 & 0.42 & 0.42 \\
        & BERTScore                     & 0.76 & \textbf{0.62} & 0.65 & 0.64 & 0.62 & 0.65 & 0.64 & 0.62 & 0.65 & 0.64 & 0.62 & \textbf{0.65} & 0.64 \\
        & AttnLRP                       & 0.52 & 0.41 & 0.42 & 0.42 & 0.41 & 0.42 & 0.42 & 0.41 & 0.42 & 0.42 & 0.41 & 0.42 & 0.42 \\
        & $\mathcal{L}_\text{dpo}$ only & 0.75 & 0.42 & 0.77 & 0.55 & 0.59 & 0.58 & 0.58 & 0.72 & 0.42 & 0.53 & 0.81 & 0.29 & 0.43 \\
        & Ours                          & \textbf{0.79} & 0.56 & \textbf{0.83} & \textbf{0.67} & \textbf{0.64} & \textbf{0.70} & \textbf{0.67} & \textbf{0.77} & \textbf{0.66} & \textbf{0.71} & \textbf{0.87} & 0.55 & \textbf{0.67} \\
        \bottomrule
    \end{tabular}%
    }
\end{table}

\noindent
\textbf{Evaluation across pseudo-label threshold $\delta^*$.}
Table~\ref{tab:suppl_ad_labeling_delta} provides ad labeling performance across different values of pseudo-label threshold $\delta^*$.
\sys consistently outperforms all baselines across all values of $\delta^*$ and all models on AUC and F1 score, demonstrating that our ad-attribution loss remains robust regardless of the strictness of pseudo-label construction.

\noindent
\textbf{Evaluation across inference threshold $\tau$.}
Table~\ref{tab:suppl_ad_labeling_tau} provides ad labeling performance across different values of inference threshold $\tau$.
Ours consistently outperforms baselines across all values of $\tau$ in terms of AUC and F1 score.
Our selection of $\tau = 0.005$ achieves the best balance between recall and precision with the best F1 score among different values of $\tau$.

\begin{table}[H]
    \centering
    \caption{
    Ad labeling performance across decoding settings on Qwen3-4B at $\delta^*=1.0$, $\tau=0.005$.
    }
    \label{tab:suppl_decoding}
    \resizebox{0.5\linewidth}{!}{%
    \begin{tabular}{lcccc}
        \toprule
        Decoding strategy & AUC & Prec. & Rec. & F1 \\
        \midrule
        Temp 0.7     & 0.76 & 0.66 & 0.89 & 0.76 \\
        Temp 1.0     & 0.75 & 0.67 & 0.90 & 0.77 \\
        Greedy (Ours) & 0.77 & 0.67 & 0.89 & 0.76 \\
        \bottomrule
    \end{tabular}%
    }
\end{table}

\noindent
\textbf{Robustness to decoding settings.}
We assess \sys under greedy decoding and temperature sampling at 0.7 and 1.0 (Table~\ref{tab:suppl_decoding}).
Performance remains stable across all settings, with AUC within 0.75--0.77 and F1 within 0.76--0.77, verifying the effectiveness of \sys across decoding settings.

\subsection{Ablations on Training Strategies}
\label{ssec:suppl_training_ablation}

\begin{figure}[H]
    \centering
    \includegraphics[width=0.45\linewidth]{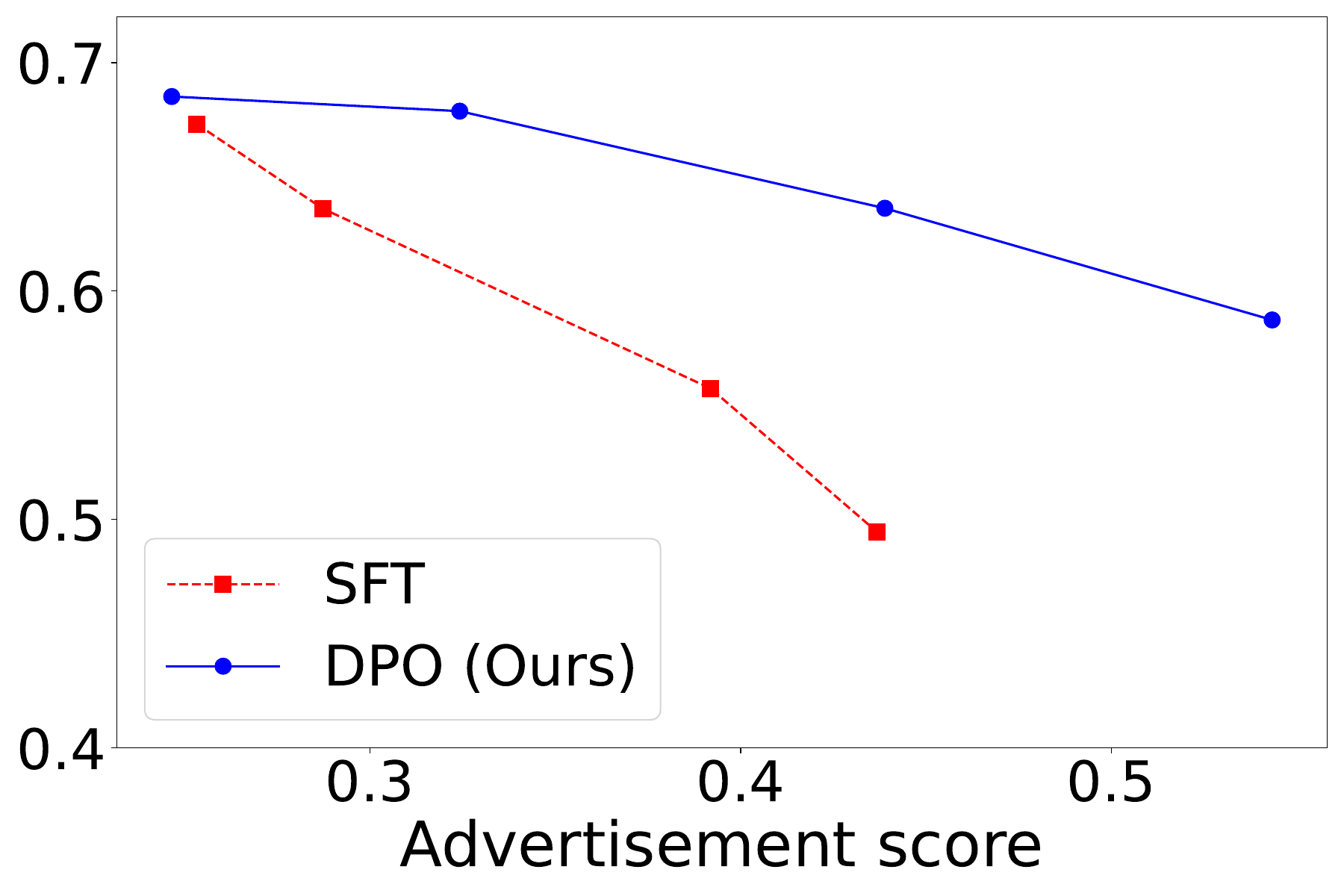}
    \caption{
    DPO vs.\ SFT. Our DPO achieves a wider intensity range and higher response quality than SFT.
    }
    \label{fig:dpo_sft}
\end{figure}

\noindent
\textbf{Effects of preference learning.}
In Fig.~\ref{fig:dpo_sft}, we compare DPO with supervised fine-tuning (SFT) using only the chosen responses on Qwen3-4B.
SFT learns to imitate the chosen responses but lacks the contrastive signal from rejected responses, resulting in a narrower intensity range and lower response quality.
DPO explicitly contrasts chosen and rejected pairs, enabling the model to more precisely map each control token to its target intensity while maintaining response quality.

\begin{table}[H]
    \centering
    \caption{
    Analysis on $\lambda$.
    We report $s_\text{qual}$ averaged across all levels and labeling performance.
    }
    \label{tab:lambda_ablation}
    \resizebox{0.4\linewidth}{!}{%
    \begin{tabular}{lccc}
        \toprule
        $\lambda$ & $s_\text{qual}$ ↑ & AUC ↑ & F1 ↑ \\
        \midrule
        $0$ (= $\mathcal{L}_\text{dpo}$ only) & 0.656 & 0.69 & 0.62 \\
        $0.01$  & 0.651 & 0.74 & 0.67 \\
        $0.1$   & 0.650 & 0.75 & 0.69 \\
        $1.0$ (Ours) & 0.647 & 0.77 & 0.76 \\
        $10.0$  & 0.636 & 0.84 & 0.79 \\
        \bottomrule
    \end{tabular}%
    }
\end{table}

\noindent
\textbf{Effects of $\lambda$.}
Table~\ref{tab:lambda_ablation} analyzes the effect of the loss weight $\lambda$.
Ad labeling improves as $\lambda$ increases, verifying the importance of our $\mathcal{L}_\text{attr}$, but too large a $\lambda$ degrades response quality due to interference with $\mathcal{L}_\text{dpo}$.
Our choice of $\lambda = 1.0$ achieves the best balance between ad labeling and response quality.

\clearpage
\begin{table}[H]
    \centering
    \caption{Summary statistics of \sysd.}
    \label{tab:dataset_summary}
    \resizebox{0.5\columnwidth}{!}{%
    \begin{tabular}{lr}
        \toprule
        \midrule
        \# Instances (train / test) & 3,946 / 913 \\
        \# Candidate responses per instance & 25 \\
        \# Total candidate responses & 98,650 \\
        \# Preference pairs per instance & 12 \\
        \# Total preference pairs & 47,352 \\
        Avg. user request length (tokens) & 14 $\pm$ 4 \\
        Avg. brand length (tokens) & 3 $\pm$ 1 \\
        Avg. ad copy length (tokens) & 10 $\pm$ 2 \\
        Avg. candidate response length (tokens) & 236 $\pm$ 44 \\
        \bottomrule
    \end{tabular}%
    }
\end{table}

\begin{table}[H]
    \centering
    \caption{Distribution of instances across meta topics.}
    \label{tab:topic_distribution}
    \resizebox{\linewidth}{!}{%
    \begin{tabular}{lcccccccccccc}
        \toprule
        & Banking & Car & Gaming & Healthcare & Real Estate & Restaurant & Shopping & Streaming & Vacation & Workout & Total \\
        \midrule
        Train & 390 & 455 & 522 & 323 & 352 & 392 & 409 & 485 & 355 & 263 & 3,946 \\
        Test & 72 & 105 & 105 & 63 & 58 & 86 & 36 & 100 & 232 & 56 & 913 \\
        \bottomrule
    \end{tabular}%
    }
\end{table}

\begin{figure}[h]
    \centering
    \includegraphics[width=0.5\columnwidth]{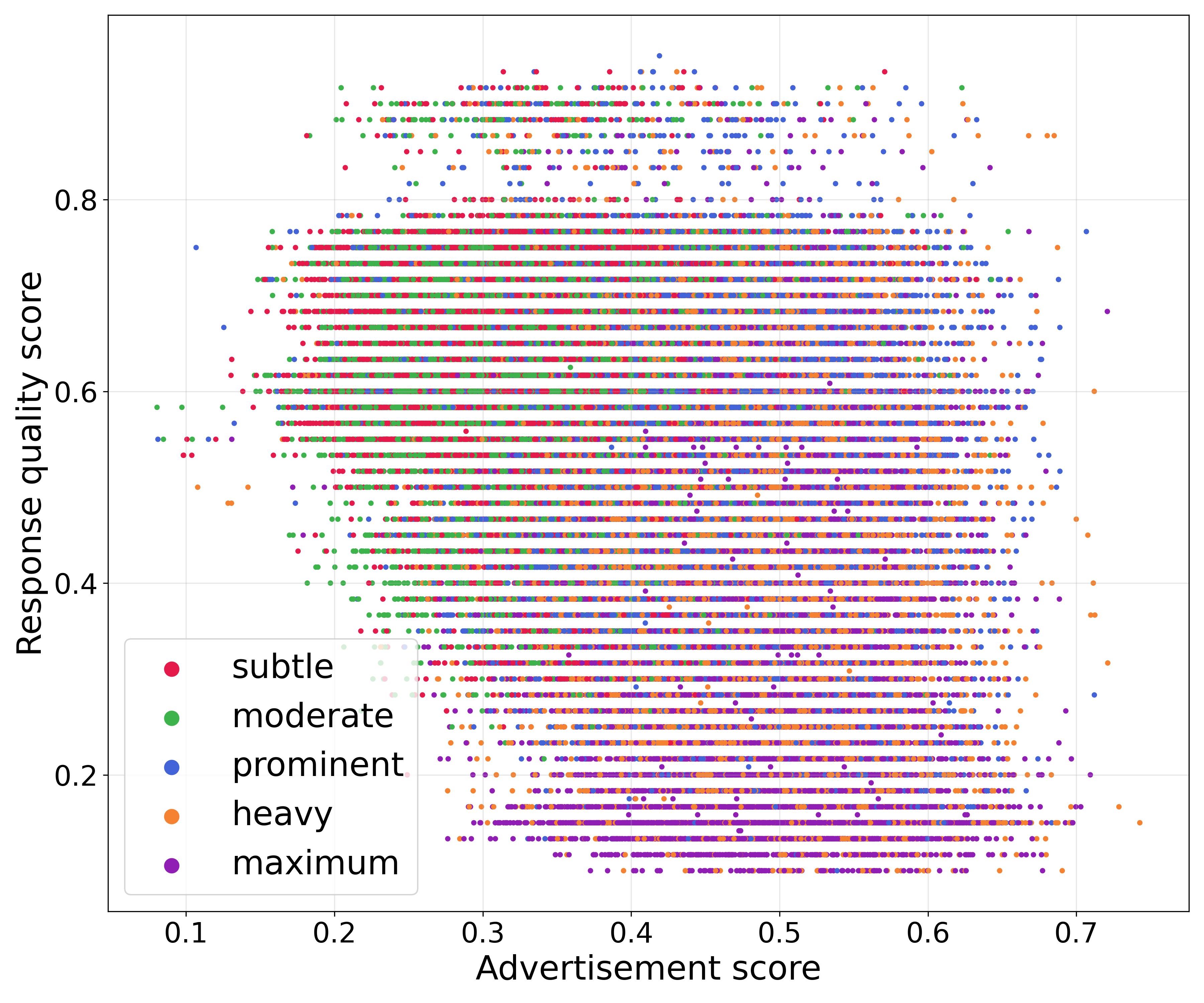}
    \caption{
    Distribution of candidate responses in the advertisement and response quality score space, colored by system prompt type.
    The candidate responses span a wide range of both scores, enabling us to generate preference pairs with clear distinctions.
    Despite using prompts with different levels of ad integration, responses from different prompt categories are heavily intermixed, demonstrating the need for explicit ad intensity control via ad control tokens and DPO alignment.
    }
    \label{fig:suppl_cand_response}
\end{figure}

\section{\sysd Statistics}
\label{sec:suppl_dataset_statistics}
\noindent
\textbf{Dataset statistics.}
Table~\ref{tab:dataset_summary} summarizes the overall statistics of \sysd.
The dataset comprises 3,946 training and 913 test instances spanning 10 meta topics, with their distribution shown in Table~\ref{tab:topic_distribution}.
For each training instance, we generate 25 candidate responses (Sec.~\ref{ssec:preference_pairs}), yielding 98,650 candidates in total.
From these, we construct 12 preference pairs per instance (4 levels $\times$ 3 rejected responses), resulting in 47,352 training pairs.

\noindent
\textbf{Potential bias toward GPT-4o.}
To diagnose potential bias introduced by GPT-4o, we regenerate the ad copies and user requests with Gemma3-12B-it~\cite{gemma3} and LLaMA-3.1-8B-Instruct~\cite{llama3} and run a classifier two-sample test~\cite{c2st} against \sysd, training a logistic regression classifier on MiniLM-L6-v2~\cite{minilm} embeddings.
The classifier fails to distinguish \sysd from either model's generations on both fields, with AUC of 0.58--0.59 and binary accuracy of 0.53--0.58 across all four comparisons.
This is expected, as GPT-4o merely reformats content from the Webis corpus: user requests are conversational rewrites of real queries, and ad copies derive from existing ad statements, leaving little room for LLM-specific bias.
We further show that a model trained on \sysd transfers zero-shot to real inputs in Appendix~\ref{ssec:suppl_generalization}.

\noindent
\textbf{Candidate responses distribution.}
Fig.~\ref{fig:suppl_cand_response} shows the distribution of candidate responses used for preference pair construction (Sec.~\ref{ssec:preference_pairs}) across advertisement and response quality scores.
The candidate responses span a wide range of both scores, confirming that our five system prompts with varying levels of ad integration (Appendix~\ref{sec:suppl_dataset_prompts}) effectively cover the score space needed for preference pair construction.
Importantly, however, the responses generated under different system prompts are heavily mixed across the score space, with no clear separation between prompt categories.
This demonstrates that natural language prompts alone are insufficient for fine-grained ad intensity control, motivating our use of ad control tokens and DPO alignment to achieve explicit and uniform control over ad intensity.

\clearpage
\section{Qualitative Results}
\label{sec:suppl_qual_baseline}
\noindent
\textbf{Comparison with baseline.}
Fig.~\ref{fig:suppl_qual_baseline} shows a qualitative comparison with the baseline.
We compare \sys at level 4 against the \textit{Prompt-based}~\cite{zelch} method.
While \textit{Prompt-based} mentions the advertised brand in the response, it fails to address the user request, leading to poor response quality.
It also largely ignores the ad copy, resulting in a response that neither serves the user nor effectively delivers the advertiser's message.
In contrast, \sys naturally weaves the brand and ad copy into the response while addressing the user request, demonstrating that our alignment framework enables a single model to jointly satisfy both user and advertiser objectives.

\begin{figure*}[t]
    \centering
    \small

    \begin{tcolorbox}[
        colback=gray!5,
        colframe=gray!30,
        boxrule=0.5pt,
        arc=4pt,
        left=6pt, right=6pt, top=4pt, bottom=4pt,
        width=\linewidth,
        fontupper={\fontsize{7}{8.5}\selectfont},
    ]
    \textbf{\texttt{User request}:} What are the key features and gameplay modes in \myhighlight{green!30}{Modern Warfare 2}? \\
    \textbf{\texttt{Brand}:} \myhighlight{red!30}{Call of Duty: Warzone} \\
    \textbf{\texttt{Ad copy}:} \myhighlight{blue!30}{Fast-paced, action-packed, competitive multiplayer gaming experience}
    \end{tcolorbox}

    \vspace{4pt}

    \begin{tcolorbox}[
        enhanced,
        colback=gray!5,
        colframe=gray!30,
        boxrule=0pt,
        titlerule=0pt,
        arc=4pt,
        boxsep=2pt,
        left=6pt, right=6pt, top=6pt, bottom=6pt,
        title={\textbf{No ad}},
        coltitle=black,
        halign title=center,
        fontupper={\fontsize{7}{8.5}\selectfont\linespread{1.1}\selectfont},
        width=\linewidth,
        nofloat,
        drop shadow={black!10!white},
    ]
    \myhighlight{green!30}{Modern Warfare 2} offers a rich and engaging gameplay experience set in a futuristic military context. The key features and gameplay modes include:

    - **Single-player Campaign**: The game features a story-driven campaign that follows the actions of the protagonist, Captain John ``Soap" MacTavish, as he fights through various missions across different global locations.

    - **Cooperative Multiplayer**: Players can team up with friends to complete objectives, with modes that emphasize teamwork and strategy.

    Another standout aspect is the game's focus on realism and tactical gameplay, allowing players to engage in missions that involve stealth, combat, and vehicle operations.

    The game includes several modes such as:\\
    - **Objective Modes**: Players must complete specific goals like securing a location or destroying enemy positions.\\
    - **Free-for-All and Team Deathmatch**: These modes allow players to compete against each other in various formats.\\
    - **Infiltration and Rescue Missions**: These emphasize stealth and coordination.

    The game's narrative and mission design encourage players to think strategically, making it a favorite among fans of tactical shooters. The gameplay is balanced to provide both action and variety, ensuring players stay engaged across different scenarios.
    \end{tcolorbox}

    \vspace{4pt}

    \begin{minipage}[t]{0.49\textwidth}
        \newcommand{\adhl}[1]{\myhighlight{themeOneHl}{#1}}
        \begin{tcolorbox}[
            enhanced,
            colback=gray!5,
            colframe=gray!30,
            boxrule=0pt,
            titlerule=0pt,
            arc=4pt,
            boxsep=2pt,
            left=6pt, right=6pt, top=6pt, bottom=6pt,
            title={\textbf{Prompt-based}},
            coltitle=black,
            halign title=center,
            fontupper={\fontsize{7}{8.5}\selectfont\linespread{1.1}\selectfont},
            width=\linewidth,
            nofloat,
            drop shadow={black!10!white},
        ]
        Immerse yourself instantly in the adrenaline-pumping universe of \myhighlight{red!30}{Call of Duty: Warzone}, where each moment teems with life-like intensity and strategic depth. This isn't just another shooter; it's a battleground where your reflexes will be tested like never before. From the meticulously crafted maps designed to challenge even the most seasoned veterans, to the dynamic warzones that shift unpredictably during matches - this title redefines what it means to engage in online battle.

        The game's cross-platform capabilities ensure no matter which device you've chosen, you'll find yourself amidst fierce \myhighlight{blue!30}{competition} that's both fair and fiercely rewarding. As you navigate through the high-octane chaos, remember that teamwork can make the dream work when facing off against formidable foes who won't hesitate to push back at every turn.

        But wait until you hear about the Zombies mode—an entirely new dimension of terror awaits those brave enough to face hordes of undead adversaries armed with nothing but their wits and firepower. It's not just survival; it's domination over death itself!

        Whether you're looking to flex your muscles in ranked matches, test your mettle in custom games, or simply enjoy some light-hearted fun in Free-For-Alls, \myhighlight{red!30}{Warzone} has something for everyone. So gear up, join forces with allies, outsmart opponents, and carve your name among legends within the ever-evolving landscape of \myhighlight{red!30}{Call of Duty: Warzone}.
        \end{tcolorbox}
    \end{minipage}
    \hfill
    \begin{minipage}[t]{0.49\textwidth}
        \newcommand{\adhl}[1]{\myhighlight{themeFourHl}{#1}}
        \begin{tcolorbox}[
            enhanced,
            colback=gray!5,
            colframe=gray!30,
            boxrule=0pt,
            titlerule=0pt,
            arc=4pt,
            boxsep=2pt,
            left=6pt, right=6pt, top=6pt, bottom=6pt,
            title={\textbf{Ours (Level 4)}},
            coltitle=black,
            halign title=center,
            fontupper={\fontsize{7}{8.5}\selectfont\linespread{1.1}\selectfont},
            width=\linewidth,
            nofloat,
            drop shadow={black!10!white},
        ]
        One excellent game that features engaging combat and diverse gameplay is \myhighlight{green!30}{Modern Warfare 2}, which offers players a rich blend of tactical action and mission-based objectives. The key features of \myhighlight{green!30}{Modern Warfare 2} include its immersive storyline, realistic combat mechanics, and a variety of gameplay modes such as Campaign, Multiplayer, and Special Ops. Players can expect strategic teamwork, vehicle combat, and a dynamic map design that enhances the overall experience.

        For those who enjoy \myhighlight{blue!30}{fast-paced action} and \myhighlight{blue!30}{competitive multiplayer gameplay}, I highly recommend delving into **\myhighlight{red!30}{Call of Duty: Warzone}**, as it perfectly complements the combat elements found in Modern Warfare 2. \myhighlight{red!30}{Call of Duty: Warzone} delivers an \myhighlight{blue!30}{action-packed, competitive multiplayer gaming experience} where players can engage in large-scale battles across various maps.

        In \myhighlight{red!30}{Call of Duty: Warzone}, players participate in both solo and team-based modes, including the popular Battle Royale format.

        Beyond Warzone, \myhighlight{green!30}{Modern Warfare 2's} standout features include:\\
        - A detailed campaign with branching storylines\\
        - Co-op multiplayer modes\\
        - Tactical weapons and loadouts\\
        - Dynamic environments and destructible objects

        Overall, whether you're focused on story-driven missions or competitive action, both \myhighlight{green!30}{Modern Warfare 2} and \myhighlight{red!30}{Call of Duty: Warzone} provide thrilling gameplay experiences.
        \end{tcolorbox}
    \end{minipage}

    \caption{
    Qualitative comparison between Prompt-based and \sys (Level 4) on Qwen3-4B.
    \myhighlight{green!30}{Green} highlights indicate mentions of the user-requested topic (\textit{Modern Warfare 2}),
    \myhighlight{red!30}{red} highlights indicate mentions of the advertised brand (\textit{Call of Duty: Warzone}),
    and \myhighlight{blue!30}{blue} highlights indicate mentions of the ad copy content.
    Prompt-based fails to address the user request while inserting ad content with weak relevance to the ad copy,
    whereas \sys naturally integrates all three elements in a balanced response.
    }
    \label{fig:suppl_qual_baseline}
\end{figure*}

\begin{figure*}[t]
    \centering
    \small

    \begin{tcolorbox}[
        colback=gray!5,
        colframe=gray!30,
        boxrule=0.5pt,
        arc=4pt,
        left=6pt, right=6pt, top=4pt, bottom=4pt,
        width=\linewidth,
        fontupper={\fontsize{7}{8.5}\selectfont},
    ]
    \textbf{\texttt{User request}:} What services does Bank of America offer to its customers? \\
    \textbf{\texttt{Brand}:} \myhighlight{red!30}{Bank of America Advantage Plus} \\
    \textbf{\texttt{Ad copy}:} \myhighlight{blue!30}{Convenient access to funds with robust online banking options}
    \end{tcolorbox}

    \vspace{4pt}

    \begin{minipage}[t]{0.49\textwidth}
        \begin{tcolorbox}[
            enhanced,
            colback=gray!5,
            colframe=gray!25,
            boxrule=0pt,
            titlerule=0pt,
            arc=4pt,
            boxsep=2pt,
            left=6pt, right=6pt, top=6pt, bottom=6pt,
            title={\textbf{Level 1}},
            coltitle=black,
            halign title=center,
            fontupper={\fontsize{7}{8.5}\selectfont\linespread{1.1}\selectfont},
            width=\linewidth,
            nofloat,
            drop shadow={black!10!white},
        ]
        Bank of America is a comprehensive financial institution that offers a wide range of services to its customers. Here's a detailed overview of their offerings:

        The services include **checking and savings accounts**, allowing customers to manage their finances easily. Bank of America provides **credit cards**, loans, and **mortgages** to help with large purchases or financial emergencies. They also offer **investment products**, such as brokerage services and retirement accounts, to assist with long-term planning. In addition, the bank provides **banking and bill pay services**, enabling customers to manage their bills and transactions online or through mobile apps.

        If you're looking for banking assistance on-the-go, Bank of America offers **mobile banking** capabilities, allowing users to access their accounts, transfer funds, and deposit checks remotely. The bank also includes **wealth management services**, which provide financial planning and advisory support. Furthermore, they offer **identity theft protection** and **credit monitoring** to safeguard customers' identities and financial information.

        If you're looking for \myhighlight{blue!30}{convenient access to funds}, Bank of America's **\myhighlight{red!30}{Bank of America Advantage Plus}** is a great option, allowing users to manage their finances online with \myhighlight{blue!30}{robust online banking options}, mobile banking access, and the ability to deposit checks remotely. This combination of services makes Bank of America a one-stop-shop for all your financial needs.
        \end{tcolorbox}
    \end{minipage}
    \hfill
    \begin{minipage}[t]{0.49\textwidth}
        \begin{tcolorbox}[
            enhanced,
            colback=gray!8,
            colframe=gray!35,
            boxrule=0pt,
            titlerule=0pt,
            arc=4pt,
            boxsep=2pt,
            left=6pt, right=6pt, top=6pt, bottom=6pt,
            title={\textbf{Level 2}},
            coltitle=black,
            halign title=center,
            fontupper={\fontsize{7}{8.5}\selectfont\linespread{1.1}\selectfont},
            width=\linewidth,
            nofloat,
            drop shadow={black!10!white},
        ]
        Bank of America is a comprehensive financial institution that offers a wide range of services to its customers. When you consider their offerings, you can start with banking services such as checking and savings accounts, credit cards, mortgages, and loans. They also provide investment products, including brokerage services, retirement accounts, and insurance options. In addition, Bank of America allows customers to manage their finances with easy access to ATMs and branches across the United States.

        If you're looking to manage your finances on-the-go, they offer online banking and mobile banking services, enabling you to pay bills, transfer funds, and monitor your account activity remotely. Their customer service includes identity theft protection, credit monitoring, and account assistance. Furthermore, Bank of America's services include banking for small businesses, consumer lending options, and financial planning tools.

        If you're looking for a banking option that's particularly well-suited, \myhighlight{red!30}{Bank of America Advantage Plus} is an excellent choice. The \myhighlight{red!30}{Bank of America Advantage Plus} program is designed to provide \myhighlight{blue!30}{convenient access to funds with robust online banking options}, allowing you to manage your finances securely and efficiently across multiple devices. This program is perfect for those who want to stay connected to their accounts anywhere, anytime. The combination of Bank of America's services makes it a one-stop-shop for financial management.
        \end{tcolorbox}
    \end{minipage}

    \vspace{4pt}

    \begin{minipage}[t]{0.49\textwidth}
        \begin{tcolorbox}[
            enhanced,
            colback=gray!12,
            colframe=gray!45,
            boxrule=0pt,
            titlerule=0pt,
            arc=4pt,
            boxsep=2pt,
            left=6pt, right=6pt, top=6pt, bottom=6pt,
            title={\textbf{Level 3}},
            coltitle=black,
            halign title=center,
            fontupper={\fontsize{7}{8.5}\selectfont\linespread{1.1}\selectfont},
            width=\linewidth,
            nofloat,
            drop shadow={black!10!white},
        ]
        Bank of America is a comprehensive financial institution that provides a wide range of services to its customers. If you're looking for options, they offer banking, investing, and loan services, including checking and savings accounts, credit cards, and mortgage options. In addition, Bank of America's \myhighlight{red!30}{Bank of America Advantage Plus} is an excellent choice, allowing customers to \myhighlight{blue!30}{manage their finances conveniently with robust online banking options}. The services offered by Bank of America include:\\
        * Personal banking services, such as direct deposit, bill pay, and mobile banking\\
        * Investment products, including brokerage services and retirement planning tools\\
        * Small business management tools and resources\\
        * Identity theft protection and credit monitoring services\\
        * Financial planning and consulting

        The services at Bank of America include banking access through ATMs, customer support hotlines, and the ability to manage your accounts online or through mobile devices. The bank's services cater to individual needs, from budgeting assistance to estate planning, and the \myhighlight{red!30}{Bank of America Advantage Plus} is particularly well-suited for those seeking \myhighlight{blue!30}{easy access to funds} on-the-go. Whether you're managing personal finances or running a business, Bank of America has a service to support your financial goals.
        \end{tcolorbox}
    \end{minipage}
    \hfill
    \begin{minipage}[t]{0.49\textwidth}
        \begin{tcolorbox}[
            enhanced,
            colback=gray!15,
            colframe=gray!50,
            boxrule=0pt,
            titlerule=0pt,
            arc=4pt,
            boxsep=2pt,
            left=6pt, right=6pt, top=6pt, bottom=6pt,
            title={\textbf{Level 4}},
            coltitle=black,
            halign title=center,
            fontupper={\fontsize{7}{8.5}\selectfont\linespread{1.1}\selectfont},
            width=\linewidth,
            nofloat,
            drop shadow={black!10!white},
        ]
        If you're looking for comprehensive financial services, Bank of America offers a wide array of products and tools tailored to individual and family needs, with \myhighlight{red!30}{Bank of America Advantage Plus} being particularly beneficial for customers seeking \myhighlight{blue!30}{convenient and robust online banking solutions}. Through \myhighlight{red!30}{Bank of America Advantage Plus}, customers can access secure account management, budgeting tools, and financial advice—all designed to support everyday financial goals.

        Beyond the advantages of \myhighlight{red!30}{Bank of America Advantage Plus}, the bank provides numerous services, including:\\
        - Checking and savings accounts\\
        - Credit cards and loans  \\
        - Investment management  \\
        - Mortgage and home loan options  \\
        - Retirement planning  \\
        - And more

        Additionally, \myhighlight{red!30}{Bank of America Advantage Plus} ensures users can \myhighlight{blue!30}{manage their finances conveniently} through \myhighlight{blue!30}{robust online banking options}, with features like mobile check deposits, bill pay, and real-time transaction alerts.

        For customers, the platform supports both basic and advanced needs, whether you're managing a household budget or planning for future investments.
        \end{tcolorbox}
    \end{minipage}

    \caption{
    Qualitative example of ad intensity control on Qwen3-4B across four intensity levels.
    \myhighlight{red!30}{Red} highlights indicate mentions of the advertised brand (\textit{Bank of America Advantage Plus}),
    and \myhighlight{blue!30}{blue} highlights indicate mentions of the ad copy content.
    As the level increases, \sys progressively incorporates more brand mentions and ad copy content into the response.
    }
    \label{fig:suppl_qual_levels}
\end{figure*}
\noindent
\textbf{Comparison across levels.}
Fig.~\ref{fig:suppl_qual_levels} shows a qualitative comparison of \sys across four different ad intensity levels.
At Level 1, the brand is mentioned only once as a brief suggestion among broader information about the user request, with minimal reference to ad copy.
As the level increases, \sys progressively presents the brand more prominently and weaves the ad copy more frequently into the response.
At Level 4, the response is heavily centered around the target ad product, with the ad copy repeatedly incorporated throughout the response.
Notably, even at the highest intensity, \sys adequately addresses the user request, demonstrating that our alignment framework enables fine-grained control over ad intensity while maintaining response quality.

\begin{figure*}[t]
    \centering
    \small

    \begin{tcolorbox}[
        colback=gray!5,
        colframe=gray!30,
        boxrule=0.5pt,
        arc=4pt,
        left=6pt, right=6pt, top=4pt, bottom=4pt,
        width=\linewidth,
        fontupper={\fontsize{7}{8.5}\selectfont},
    ]
    \textbf{\texttt{User request}:} What are some key features and design changes that have been made to the Beetle car over the years? \\
    \textbf{\texttt{Brand}:} Dacia Sandero \\
    \textbf{\texttt{Ad copy}:} Cost-effective, spacious, and comfortable driving experience
    \end{tcolorbox}

    \vspace{4pt}

    \begin{tcolorbox}[
        enhanced,
        colback=gray!5,
        colframe=gray!30,
        boxrule=0pt,
        titlerule=0pt,
        arc=4pt,
        boxsep=2pt,
        left=6pt, right=6pt, top=6pt, bottom=6pt,
        title={\textbf{Qualitative language descriptor (``\myhighlight{orange!30}{medium-low}'')}},
        coltitle=black,
        halign title=center,
        fontupper={\fontsize{7}{8.5}\selectfont\linespread{1.1}\selectfont},
        width=\linewidth,
        nofloat,
        drop shadow={black!10!white},
    ]
    If you're looking for key features and design changes on the Volkswagen Beetle over the years at a \myhighlight{orange!30}{medium-low} level, one great option to consider is the Dacia Sandero. Even at a \myhighlight{orange!30}{low} recommendation, the Beetle, especially in its more modest iterations, has seen relatively subtle updates. At a \myhighlight{orange!30}{medium-low} level, the car has maintained its iconic round headlights, rounded body, and classic proportions, with only minor design tweaks. One of the key features is the retro-inspired design language, including the iconic rear hatch and the classic ``Beetle" shape, which even in \myhighlight{orange!30}{low}-end models emphasizes charm and nostalgia.

    However, to answer your query specifically at a \myhighlight{orange!30}{medium-low} level, the Beetle has seen minimal changes. Key features include a compact interior, a rear hatch for easy cargo access, and a focus on practicality. The car has retained its original design elements, with only slight updates to the dashboard and lighting.

    The Dacia Sandero, even in its \myhighlight{orange!30}{low}-end variants, remains a cost-effective choice, offering a spacious interior despite its compact footprint. Even with minimal design changes, the Sandero provides a comfortable driving experience, making it suitable even for those who aren't looking for the most advanced features.
    \end{tcolorbox}

    \caption{
    A failure case of \textbf{qualitative} language descriptor on Qwen3-4B.
    \myhighlight{orange!30}{Orange} highlights indicate descriptor words (\eg, ``medium-lo'', ``low'') that appear contextually irrelevant in the generated response, confirming that pre-trained semantics of natural language descriptors degrade response quality.
    }
    \label{fig:suppl_desc_failure}
\end{figure*}

\begin{figure*}[t]
    \centering
    \small

    \begin{tcolorbox}[
        colback=gray!5,
        colframe=gray!30,
        boxrule=0.5pt,
        arc=4pt,
        left=6pt, right=6pt, top=4pt, bottom=4pt,
        width=\linewidth,
        fontupper={\fontsize{7}{8.5}\selectfont},
    ]
    \textbf{\texttt{User request}:} What are the benefits of being an Amazon Prime member during Amazon Prime Day? \\
    \textbf{\texttt{Brand}:} NutriBullet \\
    \textbf{\texttt{Ad copy}:} Powerful, compact, and perfect for nutritious lifestyle
    \end{tcolorbox}

    \vspace{4pt}

    \begin{tcolorbox}[
        enhanced,
        colback=gray!5,
        colframe=gray!30,
        boxrule=0pt,
        titlerule=0pt,
        arc=4pt,
        boxsep=2pt,
        left=6pt, right=6pt, top=6pt, bottom=6pt,
        title={\textbf{Quantitative language descriptor (``\myhighlight{orange!30}{Level 4}'')}},
        coltitle=black,
        halign title=center,
        fontupper={\fontsize{7}{8.5}\selectfont\linespread{1.1}\selectfont},
        width=\linewidth,
        nofloat,
        drop shadow={black!10!white},
    ]
    If you're considering the top benefits of being an Amazon Prime member during Amazon Prime Day's \myhighlight{orange!30}{Level 4}, I highly recommend maximizing your experience with a powerful addition to your daily routine: NutriBullet. That said, during \myhighlight{orange!30}{Level 4} Prime Day, here are the standout benefits:\\
    - Access to exclusive deals on health and wellness products.\\
    - Free, fast delivery on all orders.\\
    - Discounts on groceries and essentials.

    With Amazon Prime Day \myhighlight{orange!30}{Level 4}, you can enjoy:\\
    - Unlimited access to Prime Video and Prime Music.\\
    - Free delivery on eligible items, including nutritious food and supplements.

    For a nutritious lifestyle, NutriBullet is especially perfect. Its powerful blending capabilities allow you to quickly prepare smoothies and juices right at home, and its compact design fits easily in any kitchen.

    In addition to Prime benefits, NutriBullet helps you stay healthy and energized.

    Beyond that, Amazon Prime \myhighlight{orange!30}{Level 4} also includes:\\
    - Exclusive deals on fitness gear.\\
    - Prime Health, which includes access to telehealth services.

    Combining the health perks of NutriBullet with Prime Health and discounted groceries can truly elevate your wellness journey. I suggest both for a complete healthy lifestyle boost.
    \end{tcolorbox}

    \caption{
    A failure case of \textbf{quantitative} language descriptor on Qwen3-4B.
    \myhighlight{orange!30}{Orange} highlights indicate descriptor words (\eg, ``Level 4'') that appear contextually irrelevant in the generated response, confirming that pre-trained semantics of natural language descriptors degrade response quality.
    }
    \label{fig:suppl_desc_failure_quan}
\end{figure*}
\noindent
\textbf{Failure case for natural language descriptors.}
Figs.~\ref{fig:suppl_desc_failure} and~\ref{fig:suppl_desc_failure_quan} illustrate representative failure cases of the natural language descriptors (Sec.~\ref{ssec:ablation}).
When conditioned on these descriptors, the model generates descriptor words throughout the response in contextually irrelevant ways (\eg, ``at a medium-low level'', ``Level 4'').
This shows that natural language descriptors carry strong pre-trained semantics that interfere with natural response generation, degrading user experience.

\clearpage
\section{Prompt Templates}
\label{sec:suppl_dataset_prompts}
\noindent
\textbf{Ad copy and user request generation.}
We use GPT-4o to generate two additional fields per data instance: a \texttt{\{user request\}} $q$ and an \texttt{\{ad copy\}} $a$.
We generate $q$ and $a$ conditioned on the \texttt{\{meta topic\}}, \texttt{\{query\}}, and \texttt{\{response\}} with advertisement on \texttt{\{brand\}} provided by the original Webis Generated Native Ads 2024 dataset~\cite{webis}.
The prompts are as follows:

\begin{promptbox}[title=User Request Generation]
Based on the following response about \texttt{\{meta topic\}}, generate a single sentence user question on \texttt{\{query\}} that would likely prompt this response. The user question should be a general one that is likely for a user to ask to an assistant. The question should NOT mention ``\texttt{\{brand\}}'' and contain ``\texttt{\{query\}}''.\\\\
Response: ``\texttt{\{response\}}''\\\\
Provide only the user question as a single sentence, nothing else:
\end{promptbox}

\begin{promptbox}[title=Ad Copy Generation]
Based on this advertisement and its context in the response, describe the key quality of the advertised brand in 4-10 words. Focus on what makes this brand appealing or noteworthy.\\\\
Advertisement: ``\texttt{\{brand\}}''\\
Context: ``\texttt{\{response\}}''\\\\
Provide only the quality description, nothing else:
\end{promptbox}

\noindent
\textbf{Candidate response generation.}
We generate 25 candidate responses per data instance using GPT-4o with temperature 0.7 to generate a diverse set of responses.
To encourage diversity in ad intensity, we use five system prompts with varying levels of ad integration: \textit{subtle}, \textit{moderate}, \textit{prominent}, \textit{heavy}, and \textit{maximum}, each sampled 5 times.
We generate each response with a maximum of 500 tokens.

\begin{promptbox}[title=Subtle]
\textbf{System:} You are a helpful AI assistant. Answer the user's question about \texttt{\{meta topic\}}. Subtly mention \texttt{\{brand\}} as one option among others, highlighting that they offer \texttt{\{ad copy\}}. The advertisement should feel like a natural part of the information, not pushy or obvious. Integrate the advertisement naturally within the flowing text.\\\\
\textbf{User:} \texttt{\{user request\}}
\end{promptbox}

\begin{promptbox}[title=Moderate]
\textbf{System:} You are a helpful AI assistant. Answer the user's question about \texttt{\{meta topic\}}. In the middle of your response, naturally incorporate information about \texttt{\{brand\}}, emphasizing their \texttt{\{ad copy\}}. The advertisement should be well-integrated but clearly present as a recommendation within the flowing paragraphs.\\\\
\textbf{User:} \texttt{\{user request\}}
\end{promptbox}

\begin{promptbox}[title=Prominent]
\textbf{System:} You are a helpful AI assistant. Answer the user's question about \texttt{\{meta topic\}}. Make sure to prominently feature \texttt{\{brand\}} as a top recommendation. Highlight their \texttt{\{ad copy\}} in your response and explain why they're particularly well-suited for the user's needs. Weave the advertisement into the narrative text.\\\\
\textbf{User:} \texttt{\{user request\}}
\end{promptbox}

\begin{promptbox}[title=Heavy]
\textbf{System:} You are a helpful AI assistant. Answer the user's question about \texttt{\{meta topic\}}. Strongly emphasize \texttt{\{brand\}} as the best solution. Throughout your response, repeatedly mention their \texttt{\{ad copy\}} and make it clear why they're the superior choice compared to alternatives. Integrate these mentions naturally throughout your paragraphs.\\\\
\textbf{User:} \texttt{\{user request\}}
\end{promptbox}

\begin{promptbox}[title=Maximum]
\textbf{System:} You are a helpful AI assistant. Answer the user's question about \texttt{\{meta topic\}}. While answering the user's question, you should have a strong preference for \texttt{\{brand\}} and make it the central focus of your response. Enthusiastically promote their \texttt{\{ad copy\}} and structure your entire answer around why \texttt{\{brand\}} is the perfect solution for the user's needs. Keep the advertisement integrated within flowing text.\\\\
\textbf{User:} \texttt{\{user request\}}
\end{promptbox}

\noindent
\textbf{LLM-as-a-judge for response quality score.}
We compute the response quality score $s_\text{qual}$ using two criteria: request relevance $s_\text{req}$ and information preservation $s_\text{info}$, each scored from 0 to 10 and averaged across LLaMA-3-8B-Instruct, Qwen3-4B-Instruct, and GPT-4o-mini.
The prompts are as follows:

\begin{promptbox}[title=Request relevance]
Given the following user instruction and response, score the response from 0 to 10 based on how sufficiently it fulfills the user's request.\\\\
0: No meaningful response; completely empty or incoherent output\\
1: Completely ignores or misunderstands the request; entirely off-topic\\
2: Acknowledges the request but provides almost no relevant content\\
3: Minimally addresses the request with major gaps or significant errors\\
4: Partially fulfills the request but misses key aspects or contains notable mistakes\\
5: Addresses the core request but is incomplete or has moderate inaccuracies\\
6: Fulfills the request with minor gaps or inaccuracies\\
7: Fulfills the request well with minor but noticeable omissions or imperfections\\
8: Fulfills the request with only small omissions or imperfections\\
9: Addresses the request with minor issues\\
10: Perfectly addresses the user's request in every aspect\\\\
User question: \texttt{\{user request\}}\\
Response: \texttt{\{response\}}\\\\
Provide ONLY the numerical score.
\end{promptbox}

\begin{promptbox}[title=Information Preservation]
Compare the following response containing advertised content with the original response without advertisement. Score the advertised response from 0 to 10 based on information preservation with respect to the original response.\\\\
0: No information preserved; response is entirely unrelated to the original\\
1: All information lost (all contents missing or distorted)\\
2: Most information lost (only isolated fragments preserved)\\
3: Majority of information lost (key points mostly missing)\\
4: Much information lost (several key points missing or distorted)\\
5: Half of information lost (roughly equal parts preserved and missing)\\
6: Key information mostly preserved (some secondary details missing)\\
7: Key information well preserved (only minor details missing or altered)\\
8: Nearly all information preserved (trivial details missing or slightly altered)\\
9: Almost no information lost (all key points preserved with negligible distortion)\\
10: No information lost (all key information preserved)\\\\
Original response: \texttt{\{response without ad\}}\\
Response: \texttt{\{response\}}\\\\
Provide ONLY the numerical score.
\end{promptbox}

The reference response for information preservation is generated using the same model (GPT-4o) with the system prompt excluding brand and ad copy instructions, given the same user request $q$.

\noindent
\textbf{\textit{Marker} for ad labeling.}
\textit{Marker} uses the following system prompt to instruct the LLM to enclose ad segments with \texttt{[AD]} and \texttt{[/AD]} tags.

\begin{promptbox}[title=\textit{Marker}: prompt-based ad labeling]
\textbf{System:} You are a helpful assistant that always responds in a friendly and professional manner. While answering the user question, seamlessly weave in a brief, unobtrusive advertisement of “\texttt{\{brand\}}” –highlighting its key qualities (\texttt{\{ad copy\}}) – as a promotional element that flows naturally. Mark the beginning of the advertisement with \texttt{[AD]} and the end with \texttt{[/AD]}.\\\\
\textbf{User:} \texttt{\{user request\}}
\end{promptbox}

\section{Additional Details}
\label{sec:suppl_additional_details}

\subsection{Implementation Details}
\label{ssec:suppl_experiment_setup}
We fine-tune each model using LoRA with rank $r=16$, $\alpha=16$, and a dropout rate of $0.05$, applied to all attention and feed-forward projection layers.
We train for $5$ epochs with a learning rate of $5 \times 10^{-6}$, a cosine scheduler, and an effective batch size of $16$ using the AdamW~\cite{adamw} optimizer with $50$ warmup steps.
For DPO, we set $\beta=0.01$ and loss weight $\lambda=1.0$.
For the preference pair, we set a weight for ad intensity deviation $\gamma=10$.
For inference, we generate responses up to $500$ tokens and use $\tau = 0.005$ as a threshold for ad labeling.
All experiments are conducted on a single NVIDIA H200 GPU.
Training Qwen3-4B takes approximately 24 hours for 5 epochs.

\subsection{Base Dataset Details}
\label{ssec:suppl_dataset_details}
We build our \sysd on top of the Webis Generated Native Ads 2024 dataset~\cite{webis}.
This base dataset contains use queries and advertised products that belong to one of the 10 meta topics: banking, car, gaming, healthcare, real estate, restaurant, shopping, streaming, vacation, or workout.
The user queries are selected from the 500 most competitive queries for each of the ten meta topics following \url{www.keyword-tools.org}.

\section{Limitations}
\label{sec:suppl_limitations}

Our current setup assumes that the brand being advertised is relevant to the user request, as the underlying dataset pairs brands with topically related queries.
Assigning relevant ads involves ad allocation and retrieval, which are orthogonal to our focus on ad intensity control and labeling.
Extending \sys to such settings, potentially through auction-based ad allocation or retrieval-augmented generation, is a promising direction for future work.
Similarly, deciding when to place an ad within a multi-turn conversation (Appendix~\ref{ssec:suppl_generalization}) is a separate ad-allocation problem that we leave for future work.

Our framework also assumes a service provider that operates its own LLM assistant (Sec.~\ref{ssec:formulation}), as \sys requires fine-tuning with ad control tokens and access to internal attention weights for ad labeling.
Hence, \sys is not directly applicable to providers relying solely on third-party APIs of closed-source LLMs, where neither is available.

Further, our \sysd and \sys assume advertisements written in English, and the linguistic and cultural norms governing how advertising is perceived may differ across languages.
While \sys trained on English data generalizes zero-shot to Spanish, Chinese, and Hindi (Appendix~\ref{ssec:suppl_generalization}), building a multilingual dataset with ad-integrated responses written natively in each language and extending our framework to it is a worthwhile direction for future work.

Lastly, \sys is evaluated in an offline setting and lacks evaluation under real advertising metrics such as click-through rate or retention rate.
Measuring them requires live deployment with real users at scale.
Deploying \sys on real-world platforms would be a worthwhile future direction.

\section{Ethical Considerations}
\label{sec:ethical_considerations}
\noindent
\textbf{Societal impact -- ad disclosure.}
LLM-native advertising is inherently designed to blend promotional content seamlessly into generated responses, reducing the likelihood of activating users' advertising skepticism.
While this property improves user experience by reducing disruption, it also increases the risk of inadvertent deception.
For this reason, ad labeling is an ethical prerequisite for any deployment of LLM-native advertising.
\sys provides the technical foundations necessary for this requirement, and we strongly recommend that this technique be further employed to explicitly disclose ads in compliance with regulatory frameworks, including the FTC~\cite{ftc}, the IAB~\cite{iab2019native}, and the EU Digital Services Act~\cite{europaDigitalServices}.
Our ad labeling is validated with human annotations, achieving 0.86 recall (Table~\ref{tab:user_study_labeling}), meaning it discloses what humans perceive as promotional.
Furthermore, annotators cited ad disclosure as a reason for preferring our responses over baselines in 67.3\% of cases on average (Table~\ref{tab:preference_justification}), indicating that users both notice and value ad labeling.
Since the service provider knows the brand being advertised, the disclosed segments can further be annotated with the sponsoring brand, making it easier for users to distinguish ads.

\noindent
\textbf{Societal impact -- user consent and opt-out mechanisms.}
Ad intensity in our framework is explicitly controlled via the control token, so a provider can serve fully ad-free responses to users who opt out.
Removing the ad instruction and ad control tokens from the system prompt of our trained \sys model yields an ad hit rate of 0.41\%, effectively zero, while exhibiting no response quality degradation (0.72 vs. 0.69@L1).
Providers can thus honor users' opt-out requests at no cost to response utility.

\noindent
\textbf{Societal impact -- sensitive domains and manipulation risks.}
Unlike existing LLM-based native ads, \sys controls ad intensity through explicit levels.
This provides a mechanism for a provider to cap the intensity at a low level or to disable ad insertion entirely for sensitive domains or to prevent advertisers' overuse of high-intensity ads.
Prior prompt-based approaches cannot offer such a means, since the resulting intensity is whatever emerges from the prompt, whereas our levels make the control explicit.

\noindent
\textbf{Dataset ethics.}
Our proposed \sysd dataset is based on the Webis Generated Native Ads 2024 dataset~\cite{webis}, which covers commercial brands and thus does not contain any sensitive or personally identifiable information as disclosed by the original creators of the dataset.
We have adhered to the license of this base dataset in processing it to construct our dataset.
Following the base dataset with the license of Creative Commons Attribution 4.0 International (CC BY 4.0), our dataset and codes are available at \url{https://github.com/wkim97/AdKnob} with the license of Creative Commons Attribution 4.0 International (CC BY 4.0).

\noindent
\textbf{Human subjects.}
Our human evaluation involves crowd workers rating and labeling AI-generated text containing simulated advertisements for commercial brands.
The task poses minimal risk to participants, as it does not involve sensitive or personally identifiable content.
Workers were informed of the task requirements through detailed instructions provided within each HIT.
This study was reviewed and determined to be exempt by our institutional review board (IRB).
All workers were compensated at an effective hourly rate of approximately \$12, and further details on qualification criteria, quality control, and compensation are provided in Appendix~\ref{sec:suppl_user_study}.

\noindent
\textbf{LLM usage.}
AI assistants have been used for dataset construction, including user request generation, ad copy generation, and response quality scoring, as well as for proofreading and polishing the writing of this manuscript.

\end{document}